\documentclass{article}

\usepackage[final,nonatbib]{neurips_2024}

\usepackage{hyperref}       % hyperlinks
\usepackage{url}            % simple URL typesetting
\usepackage{booktabs}       % professional-quality tables
\usepackage{amsfonts}       % blackboard math symbols
\usepackage{nicefrac}       % compact symbols for 1/2, etc.
\usepackage{microtype}      % microtypography
\usepackage{xcolor}         % colors

\usepackage{floatrow}

\usepackage{epsfig}
\usepackage{graphicx}
\usepackage{amsmath}
\usepackage{amssymb}
\usepackage{multirow}
\usepackage{booktabs}
\usepackage{subcaption}
\usepackage{algorithm}
\usepackage{algorithmic}
\usepackage{colortbl}
\usepackage{tcolorbox}

\definecolor{LightCyan}{rgb}{0.88,1,1}
\definecolor{Gray}{gray}{0.85}
\definecolor{LightGray}{gray}{0.5}
\definecolor{LightBlue}{rgb}{0.678,0.847,0.902} % Light blue color

\usepackage[font=footnotesize,skip=1pt]{caption}

\title{Adapting Diffusion Models for Improved Prompt Compliance and Controllable Image Synthesis}
\author{%
  Deepak Sridhar\\ 
  \And
  Abhishek Peri \\
  \And
  Rohith Rachala \\
  \And
  Nuno Vasconcelos \\
  \and
  Department of Electrical and Computer Engineering\\
  University of California, San Diego\\
  \texttt{\{desridha, aperi, rrachala, nvasconcelos\}@ucsd.edu}
}

\begin{document}

\maketitle

\begin{abstract}
  Recent advances in generative modeling with diffusion processes (DPs) enabled breakthroughs in image synthesis. Despite impressive image quality, these models have various prompt compliance problems, including low recall in generating multiple objects, difficulty in generating text in images, and meeting constraints like object locations and pose. For fine-grained editing and manipulation, they also require fine-grained semantic or instance maps that are tedious to produce manually. While prompt compliance can be enhanced by addition of loss functions at inference, this is time consuming and does not scale to complex scenes. 
  To overcome these limitations, this work introduces a new family of  \textit{Factor Graph Diffusion Models} (FG-DMs) that models the joint distribution of images and conditioning variables, such as semantic, sketch, depth or normal maps via a factor graph decomposition. This joint structure has several advantages, including support for efficient sampling based prompt compliance schemes, which produce images of high object recall, semi-automated fine-grained editing, text-based editing of conditions with noise inversion, explainability at intermediate levels, ability to produce labeled datasets for the training of downstream models such as segmentation or depth, training with missing data, and continual learning where new conditioning variables can be added with minimal or no modifications to the existing structure. We propose an implementation of FG-DMs by adapting a pre-trained Stable Diffusion (SD) model to implement all FG-DM factors, using only COCO dataset, and show that it is effective in generating images with 15\% higher recall than SD while retaining its generalization ability. We introduce an attention distillation loss that encourages consistency among the attention maps of all factors, improving the fidelity of the generated conditions and image. We also show that training FG-DMs from scratch on MM-CelebA-HQ, Cityscapes, ADE20K, and COCO produce images of high quality (FID) and diversity (LPIPS). \textbf{Project Page:} \href{https://deepaksridhar.github.io/factorgraphdiffusion.github.io/}{FG-DM}
  % \keywords{Diffusion \and Controllable Synthesis \and Consistency} domain-specific datasets such as 
\end{abstract}

\section{Introduction}
\label{sec:intro}

\begin{figure*}[t]\RawFloats
\centering
\includegraphics[keepaspectratio, width=0.72\columnwidth,trim=30 40 34 40, clip]{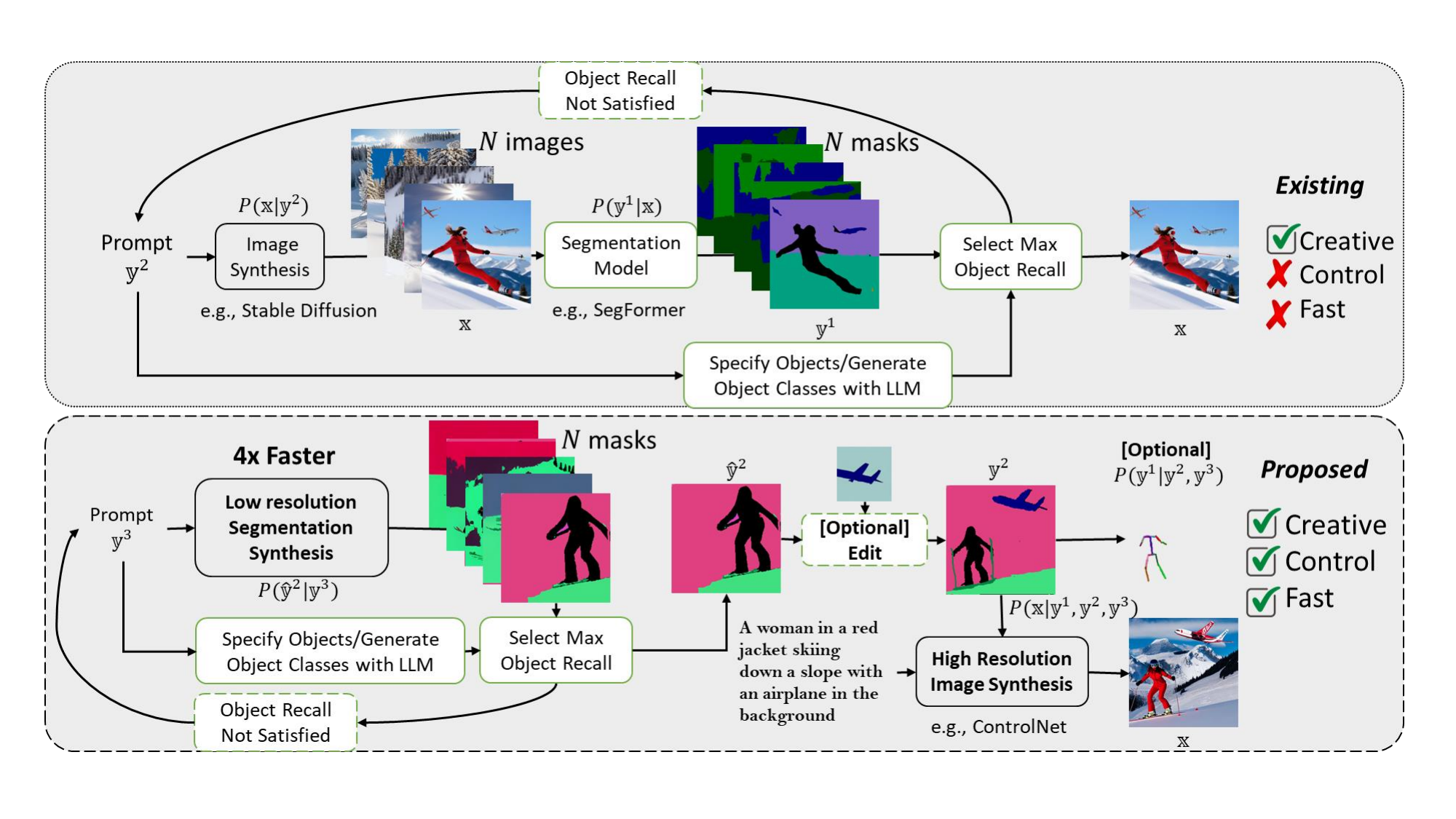}
\caption{Comparison of FG-DM (bottom) against Stable Diffusion (top) for sampling images with high object recall by modeling the joint distribution of images and conditioning variables. FG-DM supports creative, controllable, interpretable and faster (4x) image synthesis than Stable Diffusion to achieve the desired object recall. Note that the conditions $\mathbf{y^1}$ or $\mathbf{y^2}$ can be null due to classifier-free guidance training.
}
\label{fig:teaser}
\end{figure*}

Diffusion models (DMs) \cite{nethermo-sohl-dickstein15,ddpm_neurips_20, dhariwal2021diffusion, rombach2022high} have recently shown great promise for image synthesis and popularized text-to-image (T2I) synthesis, where an image is generated in response to a text prompt.
However, T2I synthesis offers limited control over image details. Even models trained at scale, such as Stable Diffusion (SD)~\cite{rombach2022high} or DALL·E 2~\cite{Ramesh2022HierarchicalTI}, have significant prompt compliance problems, such as difficulty in generating multiple objects \cite{chefer2023attendandexcite,phung2023grounded-af}, difficulty in generating text in images \cite{writingdifficult, writingdifficult2}, or to consistently produce images under certain spatial constraints, like object locations and poses \cite{yu2022scaling, casanova2023controllable, Avrahami_2023_CVPR}. These limitations have been addressed through two main lines of research. One possibility is to use {\it inference-based prompt-compliance} (IBPC) methods~\cite{chefer2023attendandexcite, phung2023grounded-af, rassin2024linguistic}, which use loss functions that operate on the cross-attention maps between image and prompt tokens to improve prompt compliance at inference. While these methods are effective for prompts involving a small number of objects, they tend to underperform for prompts involving complex scenes. Furthermore, because their complexity grows linearly with the number of scene objects, they tend to be prohibitively time-consuming for such scenes. A second possibility is to rely on DMs that support {\it visual conditioning,\/} in the form of sketches~\cite{peng2023difffacesketch}, bounding boxes~\cite{cheng2023layoutdiffuse}, scene graphs~\cite{yang2022diffusion}, reference images~\cite{yang2022paint}, etc. {\it Visually conditioned DMs\/} (VC-DMs) are usually extensions of T2I-DMs trained at scale. For example,
ControlNet \cite{zhang2023adding}, T2I-Adapter \cite{mou2023t2i} and Uni-ControlNet \cite{zhao2023uni} use a learnable branch to modulate the features of a pre-trained SD model according to a visual condition. Despite their success, VC-DMs have important limitations, inherent to {\it models of the conditional distribution} $P(\mathbf{x}|\{\mathbf{y}^i\})$ of image $\mathbf{x}$ given conditions $\mathbf{y}^i$: the need for user supplied conditions. The manual specification of visual conditions, like segmentation masks or normal maps, requires users with considerable time and skill. While, as illustrated in Figure \ref{fig:teaser} (top), conditions $\mathbf{y}^i$ can be extracted from existing images, this requires additional vision models (e.g. segmentation or edge detection), which is time-consuming.

In this work, we consider an alternative framework that attempts to mitigate all these problems using a simple but unexplored prompt compliance scheme that we denote as {\it sampling-based prompt compliance\/} (SBPC). The idea, illustrated in the top of Figure~\ref{fig:teaser} is to sample a batch of $N$ images using different DM seeds, relying on an external model (e.g. segmentation) to measure prompt compliance (e.g. by measuring object recall) and choosing the image that best complies with the prompt. While this strategy is frequently successful, generating multiple high resolution images significantly increases the inference time, rendering the approach impractical even for small values of $N$ as we will show in section \ref{object-recall-discussion}. Furthermore, it does not address the need for specification of the  conditions $\mathbf{y}^i$ required by the VC-DM. We address these problems by introducing a new family of  {\it Factor Graph-DMs\/} (FG-DMs). As illustrated in the bottom of Figure \ref{fig:teaser}, a FG-DM is a modular implementation of the joint distribution $P(\mathbf{x}, \{\mathbf{y}^i\})$ by decomposing the image synthesis into two or more factors, that are implemented by jointly trained VC-DMs. The figure shows an example decomposition of the distribution $P({\bf x}, \{{\bf y}^i\}_{i=1}^2|{\bf y}^3)$ of image $\bf x$, pose ${\bf y}^1$, and segmentation ${\bf y}^2$, given prompt ${\bf y}^3$, into three factors: $P({\bf y}^2|{\bf y}^3)$ for the synthesis of segmentation given prompt, $P({\bf y}^1|\{{\bf y}^i\}_{i=2}^3)$ for pose ${\bf y}^1$ synthesis conditioned on both, and $P(\mathbf{x}| \{\mathbf{y}^i\}_{i=1}^3)$  for image  synthesis given all conditions. 

\begin{figure*}[t]
\centering
\includegraphics[width=0.8\columnwidth, trim=0 132 0 132, clip]{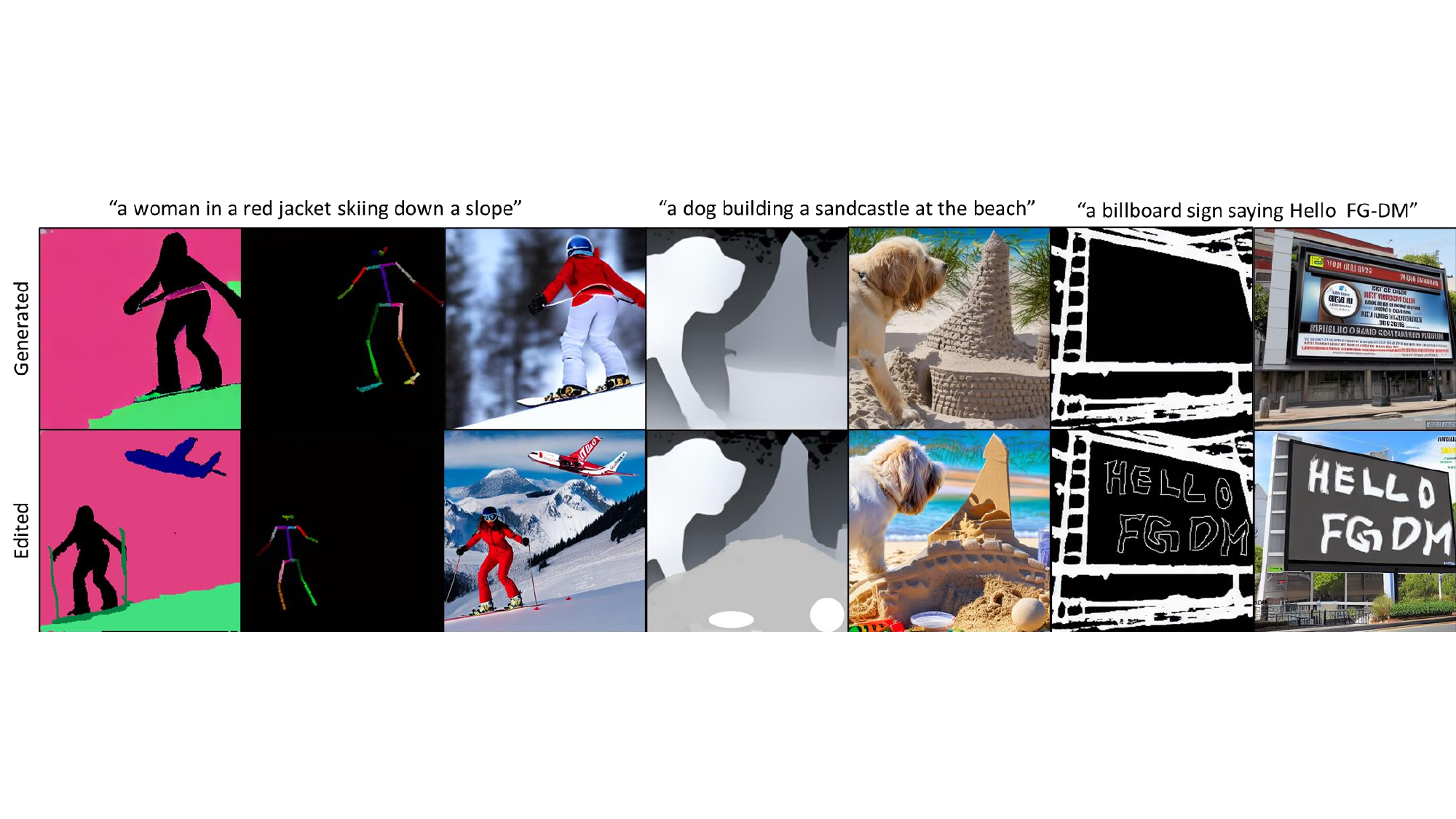}
\caption{\label{fig:edit}\label{edit_sketch}\label{edit_depth}\label{edit_map} FG-DM-based controllable image generation via editing segmentation, depth and sketch maps. Top: generated conditions and images. Bottom: edited ones. Note that only the segmentation map is edited, pose and images are conditionally generated given edited map.}
\end{figure*}

The FG-DM framework has several advantages. First, prompt compliance can usually be measured (e.g. by computing object recall) by inspecting the conditions ${\bf y}_i$ (e.g. segmentation map). The gain is that these can be generated with less diffusion steps and resolution than the final image $\bf x$. For example, we have observed no loss of image quality by sampling segmentation maps of quarter resolution. This increases the speed of SBPC \textbf{by 4x}, making it a practical prompt compliance scheme. We show that sampling with $N=10$ different seeds and choosing the image of maximum recall increases prompt compliance (object recall) by 15\% as compared to sampling with one seed. For complex scenes, it is also much faster and more effective than using IBPC methods (see Table~\ref{tab:compare_rec_hyperparam}). Second, as illustrated on the bottom of Figure \ref{fig:teaser}, the modular nature of the FG-DM offers image editing  capabilities.
New objects can be added by \textit{synthesizing} them separately while existing objects can be resized and/or moved to the desired spatial location. In Figure \ref{fig:teaser} (also in Figure \ref{fig:edit} with more detail), an airplane is added to the background while the person is resized, moved to the left of the image, and pose flipped. We introduce a simple image editing tool for performing these edits. Figure~\ref{fig:edit} shows other examples of fine-grained image editing with FG-DM  for semantic, depth, and sketch maps factors. In the center, the dog is placed behind the sandcastle (and some objects are added to the foreground) by manipulation of a depth map, and in the right the desired text ``Hello FG-DM" is scribbled on the sketch map produced by the model. 
Third, the FG-DM can reuse factors in the literature. For example, the ControlNet is used to implement the image synthesis factor $P({\bf x}|\{{\bf y}^i\}_{i=1}^3)$ in Figures~\ref{fig:teaser} and \ref{fig:edit}. Fourth, the FG-DM can produce labeled datasets for the training of downstream systems (e.g. image segmentation),  
and naturally supports continual learning schemes, where image synthesis and manipulation are gradually enhanced by the progressive addition of new VC-DMs to the factor graph, with limited or no retraining of existing ones. 

\begin{figure*}[t]\RawFloats
\centering
\includegraphics[width=0.8\columnwidth, trim=0 62 0 62, clip]{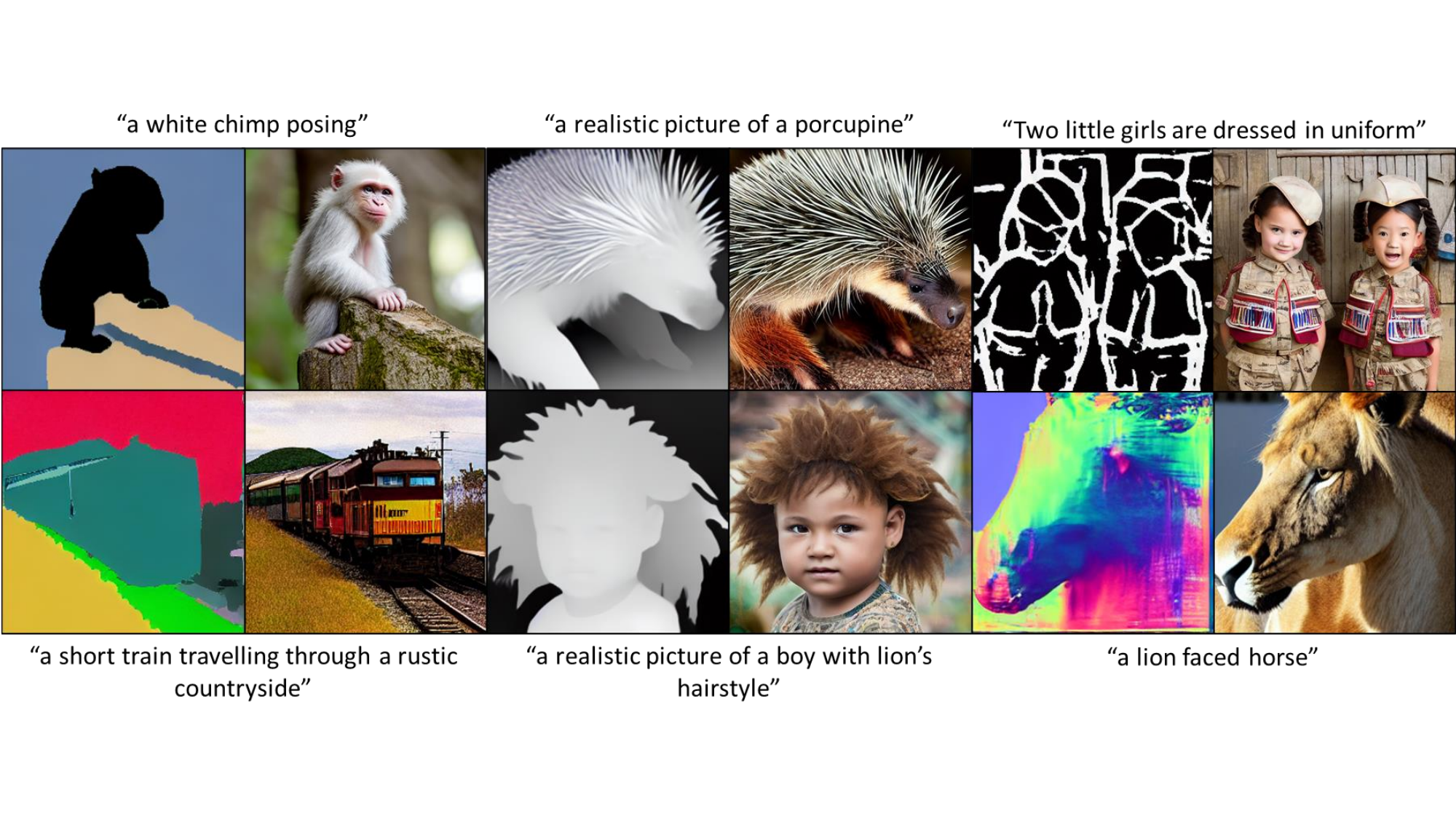}
\caption{\label{fig:JDM} Synthesized segmentation/depth/sketch/normal maps and corresponding images by an FG-DM adapted from SD using COCO. The FG-DM generalizes to prompts beyond this dataset such as porcupine, chimp and other creative prompts shown.}
\end{figure*}

Since FG-DM models the joint distribution $P({\bf x}, \{{\bf y}^i\}),$ it is a variant of {\it joint DMs\/} (JDMs) and training a FG-DM from scratch requires large scale datasets of  (condition, image) pairs, which are expensive to obtain. However, we show that this difficulty can be overcome by adapting existing foundation VC-DMs, such as SD,  to implement each factor of the FG-DM. We propose a joint prompting scheme to implement this adaptation and introduce an attention distillation loss that distills the attention maps from a pre-trained SD model to implement the condition synthesis factors $P({\bf y}^k | \{{\bf y}^i\}_{i=1}^{k-1})$, by minimizing the KL divergence between the two and show that this improves the fidelity of the generated conditions.
This greatly reduces the training costs and enables much greater generalization ability than would be possible by training on existing (condition,image) datasets alone. Figure~\ref{fig:JDM} shows that FG-DM exhibits robust generalization by synthesizing depth, normals and their corresponding images for novel objects not present in the training data. This approach also facilitates cross-model information transfer, enhancing explainability and showcasing the FG-DM's versatility in complex synthesis tasks. In summary, this paper makes the following contributions

\begin{itemize}
    \vspace{-.1in}
    \item We propose a new framework for T2I, the FG-DM, which supports the modeling of the joint distribution of images and conditioning variables, while maintaining access to all their conditional relationships
    \vspace{-.05in}
    \item We show that FG-DMs enable practical SBPC, by leveraging fast condition synthesis and filtering by object recall, and allow both fine-grained image editing with minimal effort and data augmentation.
    \vspace{-.05in}
    \item We show that FG-DMs can be implemented by adapting pre-trained T2I models (e.g., SD) using efficient prompt-based adaptation on relatively small datasets (e.g. COCO) while exhibiting interesting generalization, e.g. generalizing to prompts involving concepts not covered by these datasets. 
    \vspace{-.05in}
    \item We introduce an attention distillation loss that improves the fidelity of the synthesized conditions and enables transfer of information 
    from SD to all factors of the FG-DM.
    \vspace{-.05in}
    % \vspace{-.1in}
    \item We show that FG-DMs trained from scratch on domain-specific datasets such as MM-CelebA-HQ, ADE20K, Cityscapes and COCO consistently obtains high quality images (lower FID scores) with higher image diversity, i.e. higher LPIPS scores, than standard JDMs.
\end{itemize}

\section{Related Work}
\textbf{Text-to-Image (T2I) Diffusion models}\cite{ho2020denoising,nichol2021improved, Ramesh2022HierarchicalTI,ramesh2021zero} learn to synthesize images from noise conditioned by a text encoder, usually CLIP~\cite{radford2021learning}.
Latent DMs (LDMs) \cite{rombach2022high} implement DMs in the latent space learned by an autoencoder trained on a very large image dataset, to  reduce inference and training costs. T2I models typically employ classifier-free guidance \cite{ho2022classifierfree} to balance prompt compliance with sample diversity. However, they often struggle with complex prompts, which led to the development of
\text{conditional DMs}. They model the  distribution $P(\mathbf{x}| \{\mathbf{y}^k\})$ of image $\mathbf{x}$ given a set of $K$ conditions $\mathbf{y}^k$.  
%LDMs perform the forward diffusion and reverse denoising process in a compressed latent space without compromising on the synthesis quality. 
VC-DMs, such as ControlNet \cite{zhang2023adding}, T2I-Adapter \cite{mou2023t2i}, and HumanSD \cite{ju2023humansd}, use adapters to condition image generation on visual variables $\mathbf{y}^i$, such as depth, semantic, pose, or normal maps.
However, these methods are limited to conditions  $\mathbf{y}^k$ generated from existing images or by manual sketching, which can be hard to obtain, especially for $K>1$ (e.g. simultaneous segmentation and surface normals), and may be inconsistent. Instead, the proposed FG-DM enables the automated joint generation of all conditions while still allowing users the ability to edit them.

\text{\bf Joint Models} model the distribution $P(\mathbf{x}, \{\mathbf{y}^k\})$, frequently by concatenating all variables during image generation. For example, SemanticGAN \cite{azadi2019semantic} and GCDP \cite{park2023learning} use a single model to generate pairs of images $\mathbf{x}$ and semantic maps $\mathbf{y}$,
while Hyper Human \cite{liu2023hyperhuman} uses a single model to generate depth maps, normal maps, and images from human pose skeletons. 
These models lack access to the conditional distribution  $P(\mathbf{x}| \mathbf{y})$, which is critical for fine-grained image editing. 4M \cite{mizrahi2023m} trains an autoregressive generalist model using all conditioning variables jointly. However, it is not modular, requires large paired datasets to train and does not support continual learning of new classes.
FG-DMs are more closely related to methods like Semantic Bottleneck GAN \cite{azadi2019semantic}, Make-a-Scene \cite{makeascene} and Semantic
Palette \cite{lemoing2021semanticpalette}, which model the joint distribution as the composition of a semantic map distribution $P(\mathbf{y})$ and a conditional 
model $P(\mathbf{x} | \mathbf{y})$ for generating image $\mathbf{x}$ given synthesized layout $\mathbf{y}$. "Make a Scene" \cite{makeascene} learns a VAE from segmentation maps and samples segmentation tokens from its latent space. A transformer then combines those with tokens derived from text, to synthesize an image.  
 All joint models above have important limitations: they are trained from scratch, only consider semantic (discrete) conditioning variables and do not scale to the generation of high-fidelity images of complex natural scenes, involving a large number of semantic classes \cite{park2023learning}, such as those in COCO~\cite{cocodataset} without access to large-scale datasets. 

 \noindent\textbf{Inference-Based Prompt Compliance (IBPC)} methods\cite{chefer2023attendandexcite, phung2023grounded-af, rassin2024linguistic} attempt to improve prompt compliance by optimizing the noise latent at each diffusion iteration with a loss that maximizes attention to each noun or the binding between prompt attributes and nouns. However, these methods are time-consuming, require careful hyperparameter fine-tuning and do not work well for multiple object scenes. The proposed FG-DMs build on the power of VC-DMs and rely on SBPC, which samples various images and selects the one most compliant with the prompt. Since this only requires the synthesis of conditions, it can be done efficiently even for complex scenes. 
 %adaptation of existing foundation models, such as SD, which can be done with small datasets (e.g. COCO) without sacrificing generalization. It  for high compliance (object recall) and controllable T2I synthesis with diverse visual conditioning variables. 

\section{The Factor-Graph Diffusion Model}

\subsection{Diffusion Models}

{\bf DMs:} DMs~\cite{nethermo-sohl-dickstein15,ddpm_neurips_20} are probabilistic models based on two Markov chains. In the forward direction,  white Gaussian noise is recursively added to image $\mathbf{x}$, according to 
\begin{eqnarray}
\mathbf{z}_t = \sqrt{\alpha_t} \mathbf{z}_0 + \sqrt{1-\alpha_t} \epsilon_t,  \quad \epsilon_t \sim {\cal N}( \mathbf{0}, \mathbf{I}),
\label{eq:zt}
\end{eqnarray}
where $\mathbf{z}_0=\mathbf{x}$, ${\cal N}$ is the normal distribution, $\mathbf{I}$ the identity matrix,  $\alpha_t = \prod_{k=1}^t (1-\beta_k)$, and $\beta_t$ a pre-specified variance.
In the reverse process, a neural network $\epsilon_\theta(\mathbf{z}_t, t)$ recurrently denoises $\mathbf{z}_t$ to recover $\mathbf{x}$. 
This network is trained to predict noise $\epsilon_t$, by minimizing the risk defined by the loss 
    $\mathcal{L} = ||\epsilon_t - \epsilon_\theta(\mathbf{z}_t, t)||^2.$
Samples are generated with $ \mathbf{z}_{t-1} = f( \mathbf{z}_{t}, \epsilon_\theta(\mathbf{z}_t, t))$ where
\begin{eqnarray}
     f( \mathbf{z}_{t}, \epsilon_\theta) = \frac{1}{\sqrt{\alpha_t}} \left(\mathbf{z}_t - 
    \frac{\beta_t}{\sqrt{1-\alpha_t}} \epsilon_\theta\right) + \sigma \xi,
    \label{eq:zt-1}
\end{eqnarray}
with $\xi \sim {\cal N}( \mathbf{0}, \mathbf{I}), \mathbf{z}_{T} \sim {\cal N}( \mathbf{0}, \mathbf{I})$. The network
$\epsilon_\theta(\mathbf{z}_t, t)$ is usually a U-Net~\cite{ronneberger2015u} with attention~\cite{vaswani2017attention}.

\begin{figure*}[t]\RawFloats
\centering
\includegraphics[keepaspectratio, width=0.8\columnwidth,trim=1 78 1 78, clip]{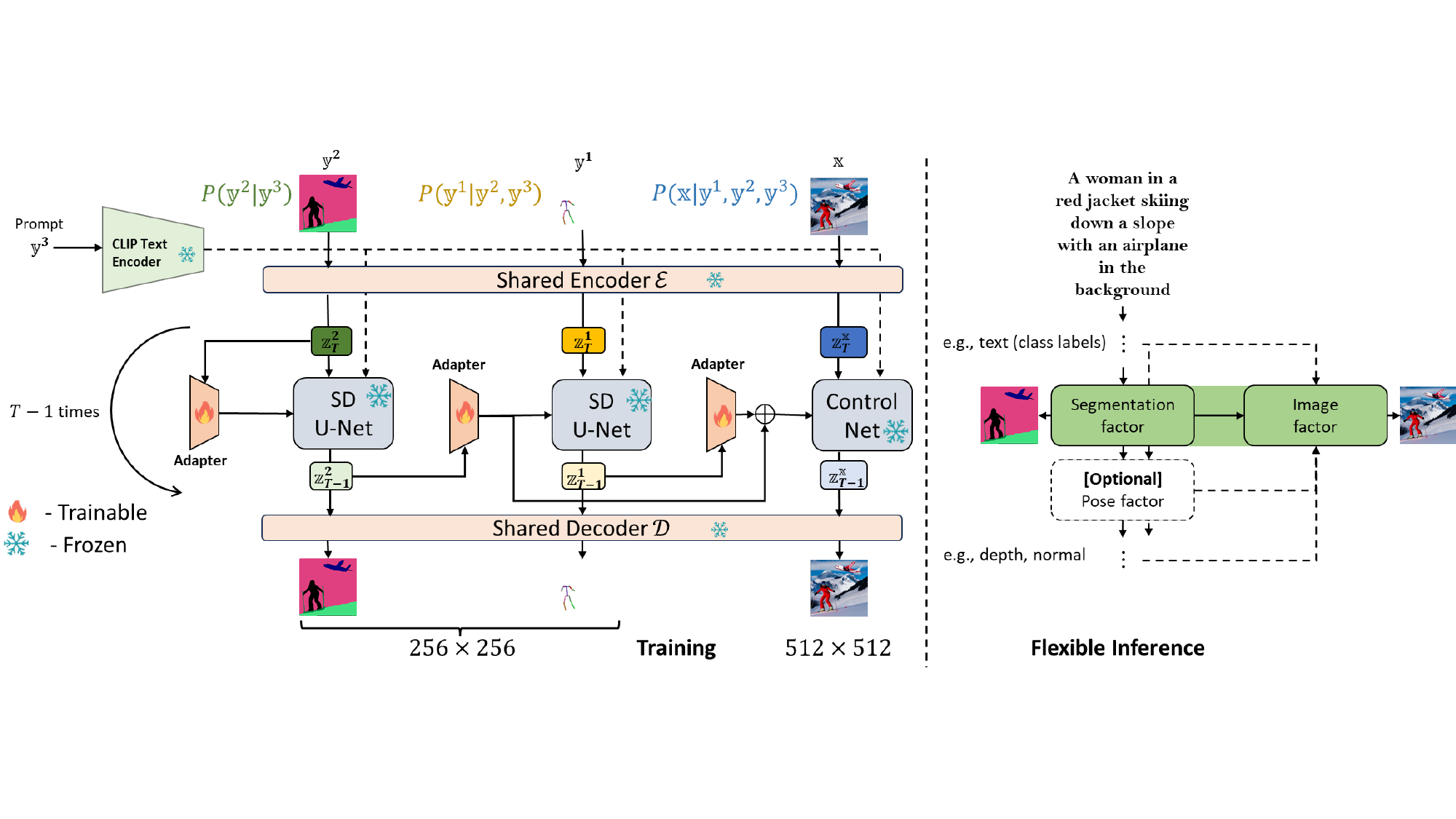}
\caption{\textbf{Left:} Training of FG-DM for distribution $P(\mathbf{x}, \mathbf{y}^1,  \mathbf{y}^2 | \mathbf{y}^3)$ of image $\mathbf{x}$, segmentation mask $\mathbf{y}^2$, and pose map $\mathbf{y}^1$, given text prompt $\mathbf{y}^3$. Each factor (conditional probability written at top of each figure) is implemented by adapting a pretrained SD model to generate a visual condition. The SD model is frozen and only a small adapter is learned per factor. The final (image generation) factor uses ControlNet without adaptation. The encoder-decoder pair and SD backbone are shared among all factors, reducing the total number of parameters. Conditional generation chains are trained at lower resolution for better inference throughput. \textbf{Right:} The FG-DM offers a flexible inference framework due to classifier-free guidance training, where only a desired subset of the factors are run, as shown in the highlighted green area. 
}
\label{fig:arch}
\end{figure*}

\subsection{The FG-DM model}

The FG-DM is a conceptually simple generalization of the DM to support $K$ conditioning variables $\mathbf{y}^i$. Rather than the conditional distribution $P(\mathbf{x}|\{\mathbf{y}^i\})$ it models the {\it joint\/} $P(\mathbf{x},\{\mathbf{y}^i\})$. Prior joint models \cite{park2023learning,liu2023hyperhuman} use a single joint denoising U-Net in the pixel or latent space to jointly synthesize $\bf x$ and $\{\mathbf{y}^i\}$. This limits  scalability to multiple conditions, increases the difficulty of editing the synthesized $\bf x$, and requires retraining the entire model to add new conditions $\bf y$. The FG-DM instead leverages the decomposition of the joint into a factor graph~\cite{forney2001codes} composed by a sequence of conditional distributions, or factors, according to
\begin{equation}
  P(\mathbf{x}, \{\mathbf{y}^i\}) = P(\mathbf{x}| \{\mathbf{y}^i\}_{i=1}^K)P(\mathbf{y}^1|  \{\mathbf{y}^i\}_{i=2}^K)\cdots P(\mathbf{y}^K),
 %       P(\mathbf{x}, \{\mathbf{y}^i\}) = P(\mathbf{x}| \mathbf{y}^1,\cdots, \mathbf{y}^K)P(\mathbf{y}^1| \mathbf{y}^2,\cdots, \mathbf{y}^K)\cdots P(\mathbf{y}^K),
        \label{eq:joint}
\end{equation}
where there are usually conditional independence relations that simplify the terms on the right hand side. 
In any case, (\ref{eq:joint}) enables the implementation of the joint DM as a modular composition of conditional DMs. This is illustrated in Figure~\ref{fig:arch}, which shows the FG-DM discussed in Figure~\ref{fig:teaser}.

\text{\bf Synthesis of conditioning variables.} We convert all conditioning variables to 3-channel inputs and use the pre-trained $\cal E$-$\cal D$ pair from the SD model to map them into the latent codes 
 for efficient training. See appendix section \ref{synthesis-conditions} for more details on this process.

\text{\bf Sampling:} The FG-DM samples $\mathbf{y}^i$ and $\mathbf{x}$ as follows. Let  $\epsilon_{\theta^x}$ be a DM for $P(\mathbf{x}| \{\mathbf{y}^i\})$, and $\epsilon_{\theta^i}$ a DM for $P(\mathbf{y}^{i} | \mathbf{y}^{i+1}, \cdots, \mathbf{y}^K)$.  In the forward direction, $\mathbf{z}^{x}_t$ and $\mathbf{z}^i_t$, the noisy versions of $\mathbf{x}$ and $\mathbf{y}^i$, respectively, are generated by using (\ref{eq:zt}) with  $\mathbf{z}^{x}_0 = \mathbf{x}$ and $\mathbf{z}^i_0 =\mathbf{y}^{i}$. In the reverse process, each denoising step is implemented with 
\begin{eqnarray}
     \mathbf{z}_{t-1}^K &=& f(\mathbf{z}^K_t, \epsilon_{\theta^K}(\mathbf{z}^K_t, t)), \label{eq:zkt-1}\\
    \mathbf{z}_{t-1}^i &=& f(\mathbf{z}^i_t, \epsilon_{\theta^i}(\mathbf{z}^i_t, \ldots, \mathbf{z}_{t-1}^K , t)), \forall i < K \label{eq:zit-1}\\
%     \quad \mbox{and} \quad  
     \mathbf{z}_{t-1}^x &= & f(\mathbf{z}^{x}_t, \epsilon_{\theta^x}(\mathbf{z}^{x}_t, \mathbf{z}^1_{t-1}, \ldots, \mathbf{z}^K_{t-1} , t)) \label{eq:zxt-1}
\end{eqnarray}
where $f(.)$ is the recursion of~(\ref{eq:zt-1}). All conditions are sampled at each denoising step.

\subsection{Adaptation of pretrained DM}

\text{\bf Architecture:} To adapt a pretrained SD model into a FG-DM factor, we modify the T2I-Adapter~\cite{mou2023t2i} to be conditioned on the current timestep $t$ and use encoded latent features of the condition(s) from previous factors as input to the adapter of the current factor. Figure~\ref{fig:arch} shows how the conditioning of (\ref{eq:zkt-1})-(\ref{eq:zxt-1}) is implemented: noisy latent feature $\mathbf{z}^K_t$ is fed to the first adapter, and the denoised latents $\mathbf{z}^i_{t-1}$ of each VC-DM are fed to the adapters of the subsequent VC-DMs (denoising of $\mathbf{z}^k_{t}, k < i$). The adapter consists of four feature extraction blocks with one convolution layer and two timestep-residual blocks for each scale. 
The encoder features $F_{i,t}^{\text{enc}}$ at the output of U-Net block $i$ are modulated as
% \begin{equation}
%F_c = F_{\text{AD}}(C), \hspace{2mm}
% \tag{3}
$\hat{F}_{i,t}^{\text{enc}} = F_{i,t}^{\text{enc}} + F_{i,t}^c, \quad i \in \{1, 2, 3, 4\}.$
% \tag{4}
% \end{equation}
where $F_{i,t}^c$ are the features produced by an adapter branch associated with $F_{i,t}^{\text{enc}}$ at timestep $t$. The SD model is kept frozen, only the adapter branches are learned per factor model. 

\text{\bf Training with Classifier-Free Guidance (CFG):} Given a training example with all the conditioning variables $(\mathbf{x}_j, \mathbf{y}^1_j, \ldots, \mathbf{y}^K_j)$, we randomly select $1,\cdots,K-1$ conditioning variables as null condition for CFG, 20\% of the training time. This facilitates unconditional training of each factor model which supports flexible inference. As a result, only a desired subset of the conditions of the FG-DM are run, as illustrated in the right of Figure~\ref{fig:arch} (highlighted in green). 

\text{\bf Attention Distillation.} The synthesis of conditions like semantic or normal maps requires learning to precisely associate spatial regions with object identities or geometry, which is a difficult task. To encourage the binding of these properties across conditions and image, we ground them on the SD attention maps, which are known to encode prompt semantics~\cite{hertz2022prompt}. The intuition is that the attention dot-products between word tokens and image regions should remain approximately constant for all factor models. Ideally, a given text prompt word should elicit a similar  region of attention in the VC-DM that synthesizes each condition. To encourage this, we introduce an attention distillation loss, based on the KL divergence between the self and cross attention maps of the pretrained SD model and those of the adapted models. This can be viewed as using the pretrained DM as a teacher that distills the knowledge contained in the attention maps to the adapted student. This distillation also helps the adapted model retain the generalization to unseen text prompts, such as the examples of Fig. \ref{fig:teaser}. 
Formally, the attention distillation loss is defined as
\begin{equation}
\resizebox{0.94\columnwidth}{!}{$
\begin{aligned}
    \mathcal{L}_{KL}(f^t, f^s) &= \sum_{j\in\{0,1\}} \text{KL}\left(\sum_{i=1}^{L_j} g_{ij}(F^t_{ij}(Q^t_{ij},K)) \, \Big\|\, \sum_{i=1}^{L_j} g_{ij}(F^s_{ij}(Q^s_{ij},K))\right) = \sum_{j\in\{0,1\}} \sum_{i=1}^{L_j} g_{ij}(F^t_{ij}(Q^t_{ij},K)) \log\left(\frac{\sum_{i=1}^{L_j} g_{ij}(F^t_{ij}(Q^t_{ij},K))}{\sum_{i=1}^{L_j} g_{ij}(F^s_{ij}(Q^s_{ij},K))}\right)
\end{aligned}
$}
\end{equation}
where superscript $t$ $(s)$ denotes  SD-teacher (SD-student), $g$ implements a bilinear interpolation needed to upscale all maps to a common size,
$F$ is the softmax of the products between query noise feature matrix ($Q$) and key CLIP text feature ($K$), $L_j$ number of attention layers, and index $j=0,1$ denotes self and cross attention layers respectively. The overall loss is the sum of distillation losses between teacher $t$ (pre-trained SD model) and all students $\mathcal{L}_{KL} = \sum_{i=1}^K \mathcal{L}_{KL}(f^t, f^{s^i}),$
where $s^i$  is the student model adapted to condition $\mathbf{y}^i$. For multi-condition FG-DMs, the distillation loss is only required for the synthesis of the first condition, which is conditioned by text alone. For the subsequent factors, which are already conditioned by a visual condition (e.g. pose conditioned by segmentation in Figure \ref{fig:arch}), the attention distillation loss is not needed.

\text{\bf Loss:} Since the noise introduced in the different factors is independent, the networks are optimized to minimize the risk defined by the loss 
\begin{equation}
% \resizebox{0.94\columnwidth}{!}{$
\begin{aligned}
      \mathcal{L}_{FG} &= ||\epsilon^x_t - \epsilon_{\theta^x}(\mathbf{z}^x_t, \mathbf{z}^1_{t-1}, \ldots, \mathbf{z}^K_{t-1}, t)||^2 
      +\sum_{i=1}^K ||\epsilon^i_t - \epsilon_\theta^i(\mathbf{z}^i_t, \ldots, \mathbf{z}_{t-1}^K, t)||^2 + \lambda_{KL}\mathcal{L}_{KL}
\end{aligned}
% $}
\end{equation}
% {\color{red} CHECK indices of $z$s.}
\text{\bf Training from scratch:}\label{fg-dm-cfg}
The FG-DM can also be trained from scratch by simply concatenating the latent representation of the previous condition(s) and noise to generate the next condition instead of using adapters. Please refer to the appendix section \ref{supp-ablation} for a detailed discussion.

\section{Experimental Results}

 \textbf{Datasets and models:} We consider four conditioning variables in this paper: segmentation, depth, normal and sketch maps. The pretrained SD v1.4 model is adapted using the COCO-WholeBody dataset\cite{cocodataset, jin2020whole}, with 256 input resolution, to train all condition factors. Groundtruth (GT) is as follows: COCO GT segmentations, HED soft edge \cite{xie15hed} for sketch maps,  and off-the-shelf MIDAS \cite{Ranftl2022, birkl2023midas} detector for depth and normal maps. 
 %We use MM-CelebAMaskHQ \cite{lee2020maskgan}, ADE20K\cite{ade20k}, Cityscapes \cite{Cordts2016Cityscapes} and COCO-WholeBody datasets to train FG-DMs from scratch. More implementation details are presented in appendix. 
 We also present results for an FG-DM trained from scratch on MM-CelebAMaskHQ \cite{lee2020maskgan}, and for other datasets in appendix, where implementation details are also given.

\text{\bf Evaluation:} Visual quality is evaluated with Frechet Inception Distance (FID) \cite{fid_gan_17}, image diversity with LPIPS \cite{lpips_metric_18}. We also report  the Precision/Recall Values of \cite{Kynkaanniemi2019-prec-rec} and evaluate prompt alignment with CLIP score (CLIP-s) \cite{radford2021learning}. All speeds are reported using a NVIDIA-A10 GPU.

\begin{figure*}[t]\RawFloats
\centering
\includegraphics[width=\columnwidth, trim= 0 120 0 120, clip]{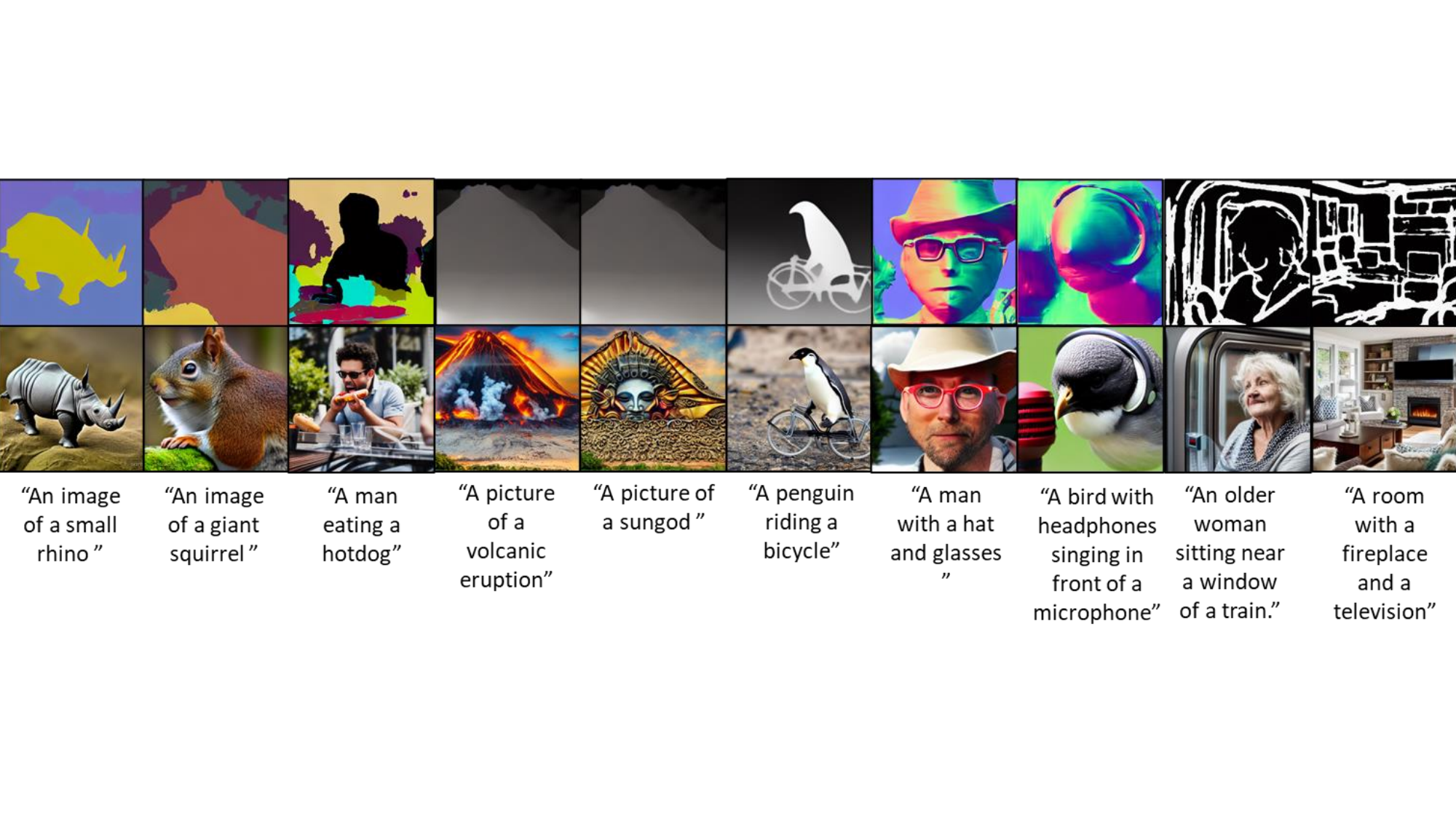}
\caption{\label{fig:seg}\label{fig:depth} \textbf{More qualitative results of FG-DM} to synthesize segmentation, depth, normal and sketch maps and their corresponding images. See appendix Figure. \ref{fig:zoom} for the higher resolution version.}
\end{figure*}

\begin{figure}[t!]\RawFloats
  \centering
\begin{minipage}{0.5\columnwidth}
  \begin{tabular}{cc}
  \includegraphics[width=.45\linewidth]{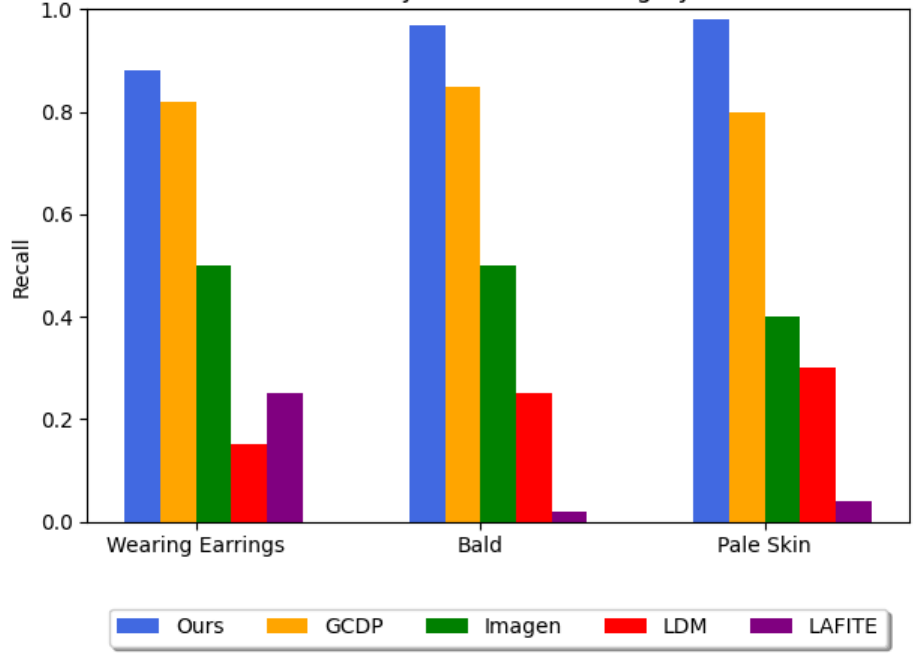} & \,\,
  \includegraphics[width=.48\linewidth]{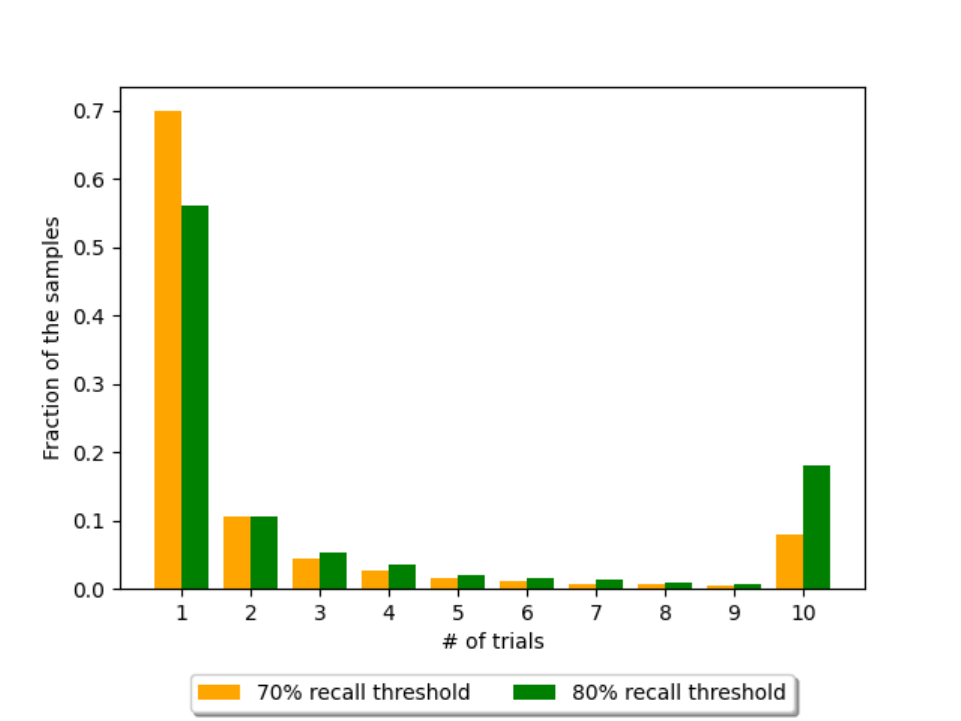} \\
  \end{tabular}
\caption{Attribute recall verification with FG-DM on MM-CelebA-HQ. \textbf{Left:} Semantic Attribute Recall. \textbf{Right:} Histogram of the number of  trials needed to reach the specified recall.}
\label{fig:sem-recall} 
\end{minipage}
\hfill
\begin{minipage}{0.48\columnwidth}
\centering
\scriptsize
\captionof{table}{Quantitative results of Text-to-Image synthesis on COCO for FG-DM with segmentation, depth, sketch and normal conditions.}
\label{tab:sdadaptfgdm}
\vspace{3mm}
\setlength{\tabcolsep}{1pt}
\begin{tabular}{ l| l| l| l| c}
\toprule
 Model& FID $\downarrow$& P $\uparrow$ & R $\uparrow$ & CLIP-s $\uparrow$\\
 % & & & Acc (\%)\\
\midrule
SD & 20.5& 0.50& \textbf{0.69}& 28.5\\
4M-XL Seg & 39.1& 0.49& 0.30& 28.8\\
\midrule
FG-DM Seg &  \textbf{19.6}& \textbf{0.56}& 0.52&\textbf{28.5}\\
FG-DM Depth &  27.0& 0.45& 0.52&27.6\\
% FG-DM Sketch &   24.6& 0.49& 0.48&28.2\\
FG-DM Sketch &   25.3& 0.39& 0.61& \textbf{30.3}\\
FG-DM Normal &   \textbf{19.3}  & \textbf{0.59}& 0.55&\textbf{28.7}\\
\bottomrule
\end{tabular}
\end{minipage}
\end{figure}

%\subsection{Experimental Results} 
\text{\bf Qualitative and Quantitative results: }
Figure \ref{fig:seg} shows additional qualitative results of synthesized segmentation (columns 1-3), depth (columns 4-6), normal (columns 7-8), sketch (columns 9-10) maps and their corresponding images for the prompts shown below  each image. The FG-DM  leverages the generalization of the pre-trained SD model to synthesize segmentation maps for object classes such as rhino and squirrel, beyond the training set (COCO). The semantic maps are colored with unique colors, allowing the easy extraction of both object masks and class labels. This shows the potential of the FG-DM for open-set segmentation, e.g the synthesis of training data for segmentation models. Note that these results demonstrate a "double-generalization" ability. While the FG-DM was never trained on squirrels or rhinos, SD was never trained to produce segmentation masks. However, FG-DM adapted from SD produces segmentations for squirrels and rhinos. 
The fourth column shows the depth map and image synthesized for the prompt \textit{"A picture of a volcanic eruption"}. In the fifth column the same caption is used to create the depth map, while the image is created with the prompt \textit{"A picture of a sungod"}. This shows that even when there is a mismatch between the prompts used for different factors, the FG-DM is able to produce meaningful images. This is a benefit of the FG-DM \emph{modularity}. Table \ref{tab:sdadaptfgdm} compares the text-to-image synthesis quality of SD, 4M-XL \cite{4m} and four FG-DM models. Despite using a bigger model and training from scratch, 4M-XL model generates lower quality images than SD and FG-DM as seen from the higher FID score. The FG-DM has higher image quality than SD for the segmentation and normal map conditions and higher clip score for sketch condition showing the effectiveness of adaptation.
% adapting a foundation model like SDs for generating visual conditions is better than training from scratch. 
% See appendix for more qualitative results and comparisons.

\text{\bf User Study:} 
We conducted a human evaluation to compare the qualitative performance of the FG-DM (adapted from SD) with $N=1$ to the conventional combination of SD+CEM, where CEM is an external condition extraction model (CEM), for both segmentation and depth conditions. We collected 51 unique prompts, composed by a random subset of COCO validation prompts and a subset of creative examples. We sampled 51 (image,condition) pairs - 35 pairs of (image,depth map), 16 pairs of (image,segmentation map) - using the FG-DM. For SD+CEM, images were sampled with SD for the same prompts, and fed to a CEM implemented with MIDAS \cite{birkl2023midas} for depth and  OpenSeed \cite{zhang2023simpleopenseed} for segmentation. The study was performed on Amazon Mechanical Turk, using 10 unique expert human evaluators per image. These were asked to compare the quality of the pairs produced by the two approaches and vote for the best result in terms of prompt alignment and visual appeal. Table \ref{tab:compare_amt_qual} shows that evaluators found FG-DM generated images (masks) to have higher quality  61.37\% (63.13\%) and better prompt alignment 57.68\% (60.98\%) of the time. These results show that the FG-DM produces images and masks that have higher quality and better prompt alignment.

\begin{figure*}[t]\RawFloats
\centering
\begin{minipage}{0.48\columnwidth}
    \centering
    \scriptsize
        \captionof{table}{User study on the qualitative preference of images/condition pairs generated by the FG-DM and SD+CEM, using 10 unique human evaluators. A. denotes (prompt) Adherence and Q. denotes Quality.}
    \label{tab:compare_amt_qual}
\setlength{\tabcolsep}{1pt}
    \begin{tabular}{l|c|c|c|c}
        \toprule
        Model & Img A.$\uparrow$ & Cond A.$\uparrow$ & Img Q.$\uparrow$ & Cond Q.$\uparrow$  \\
        \midrule
        No clear winner & 4.11& 4.70& 2.90 & 3.33\\
        \hline
        SD+CEM &  37.84&  34.31&  35.49& 33.52 \\
        FG-DM & \textbf{57.84} & \textbf{60.98} & \textbf{61.37} & \textbf{63.13} \\
        \bottomrule
    \end{tabular}
\end{minipage}
\hspace{1mm}
\begin{minipage}[h]{0.48\columnwidth}
\centering
\setlength{\tabcolsep}{1pt}
\scriptsize
\captionof{table}{Ablation of attention distillation loss for T2I synthesis on COCO validation for FG-DM. FID reported for Images/Conditions.}
\label{tab:ablationdistill}
%\resizebox{\textwidth}{!}{
\begin{tabular}{ l|c|c|c|c|c}
\toprule
 Model & Distill & FID $\downarrow$& P $\uparrow$ & R $\uparrow$ & CLIP-s $\uparrow$\\
 % & & & Acc (\%)\\
\midrule
Seg & &   20.9/164.8 & 0.54& 0.49&  28.5\\
Seg  & \checkmark &  \textbf{19.6}/\textbf{159.2} & \textbf{0.56}& \textbf{0.52}&  \textbf{28.5}\\
\hline
% sketch &  &  \textbf{24.50}  & 0.48& \textbf{0.49}& \textbf{28.17}\\
% sketch & \checkmark &  24.61 & \textbf{0.49}&  0.48  & 28.11 \\
Normal &  &  20.60/126.7  & 0.56& 0.49& 28.6\\
Normal & \checkmark &  \textbf{19.30}/\textbf{123.9} & \textbf{0.59}&  \textbf{0.55}  & \textbf{28.7} \\
\bottomrule
\end{tabular}
%}
\end{minipage}
\end{figure*}

\begin{figure*}[t]\RawFloats
\centering
\begin{minipage}{0.52\columnwidth}
\centering
\scriptsize
\captionof{table}{Object recall statistics for sampling FG-DM with different seeds and timesteps on the ADE20K validation set prompts.}
\label{tab:rec_seed_effect}
\setlength{\tabcolsep}{1pt}
\begin{tabular}{ l| c |c |c| c c c| c} 
\toprule
Samples/&Avg. Min.&Avg. Max.$\uparrow$&Avg. Med.$\uparrow$&\multicolumn{3}{c|}{Avg. \# Imgs}&Time$\downarrow$\\
Batch&Recall (\%)&Recall (\%)&Recall (\%)&0.5&0.75&0.9&(s)\\
\midrule
t=10,N=1&69&69&69&0.9&0.4&0.1&\textbf{0.45}\\
t=20,N=1&68&68&68&0.9&0.4&0.1&0.81\\
\hline
t=10,N=5&60&74&\textbf{70}&4.5&1.9&0.6&1.10\\
t=10,N=10&55&\textbf{75}&\textbf{70}&9.0&3.6&1.2&1.75\\
t=20,N=10&55&\textbf{75}&\textbf{70}&8.9&3.6&1.1&3.3\\
\bottomrule
\end{tabular}
\end{minipage}
\hspace{1mm}
\begin{minipage}{0.44\columnwidth}
    \centering
    \scriptsize
    \captionof{table}{Quantitative comparison of Object Recall for different models and configurations on the ADE20K validation set prompts.}
    \label{tab:compare_rec_hyperparam}
\setlength{\tabcolsep}{1pt}
    \begin{tabular}{l|c|c|c}
        \toprule
        Model & Clip-S$\uparrow$ & Avg. Recall$\uparrow$ & Time (s) $\downarrow$ \\
        \midrule
        SD (N=1,t=20) &  \textbf{0.301}& 59.8&\textbf{3.25}\\
        A-E (N=1,t=20)& 0.295& 63.6&36.0\\
        FG-DM (N=1,t=20) &  0.295& 
        \textbf{67.8}&4.81\\
        \hline
          SD (N=10,t=20)&  \textbf{0.301}& \textbf{75.1}&23.0\\
        FG-DM (N=10,t=10)&  0.294 & \textbf{75.1}&\textbf{5.75} \textbf{(4x$\downarrow$)}\\
        FG-DM (N=10,t=20)& 0.296 & \textbf{75.0}&\textbf{7.35} \textbf{(3x$\downarrow$)} \\
        \bottomrule
    \end{tabular}
\end{minipage}
\end{figure*}

\begin{figure*}[!t]\RawFloats
\centering
\includegraphics[width=0.9\columnwidth, trim=0 98 0 88, clip]{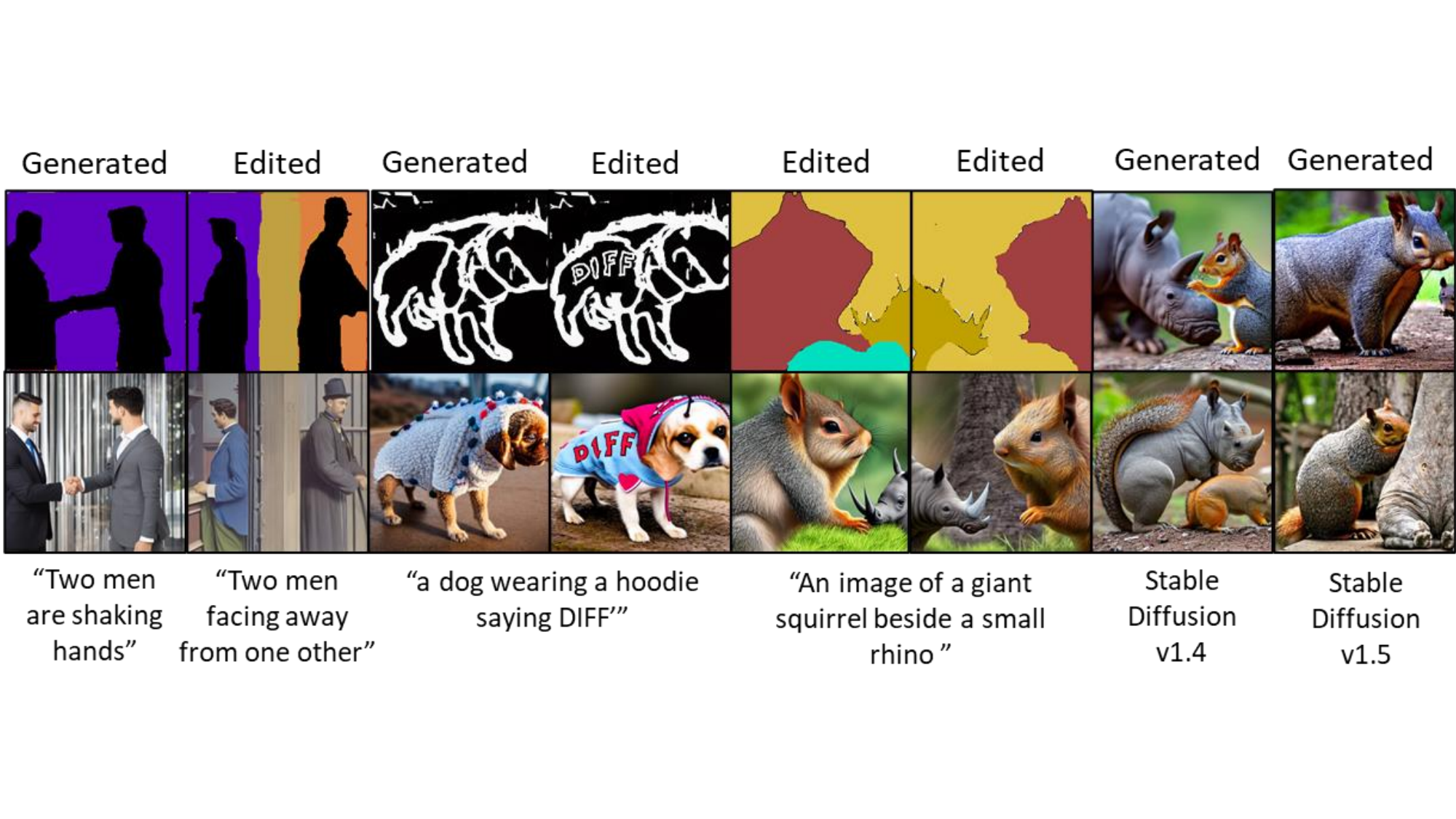}
\caption{\label{fig:edit-compare}\label{fig:edit-add} \textbf{Examples of images generated by FG-DM after editing and comparison with popular text-to-image models.} Editing is shown for flipping persons (columns 1-2), writing the desired text (columns 3-4) or realizing a difficult prompt (columns 5-6). Images generated by Stable Diffusion v1.4 and v1.5 for the same prompt are shown in the last two columns.}
\end{figure*}

\text{\bf Qualitative Image Editing Results:}
Figure \ref{fig:edit-add} shows additional examples of synthetic image editing using the FG-DM. Diffusion models have well known difficulties to perform operations such as switching people's locations~\cite{brooks2022instructpix2pix} or synthesizing images with text~\cite{writingdifficult,writingdifficult2}. The first four columns show that the FG-DM is a viable solution to these problems.  The first column shows the image generated by FG-DM for two men shaking hands while the second shows the edited version where the two men are flipped, so as to face away from each other, and combined with a different background. The 3rd and 4th columns show an example where the user scribbles "DIFF" in the synthesized sketch map, which is then propagated to the image. 
The last four columns show examples of a difficult prompt, \textit{“An image of a giant squirrel beside a small rhino”}, unlikely to be found on the web, for which existing T2I Models (SD v1.4/1.5) fail (columns 7-8). The FG-DM generates meaningful images (5th and 6th column) by simply editing the masks sampled by the segmentation factor, as discussed in Figure~\ref{fig:teaser}. In this example, the animal regions shown in Figure \ref{fig:seg} (columns 1-2) were resized, flipped and pasted onto a common segmentation. Note how FG-DM allows precise control over object positions and orientations, which differ from those of Figure \ref{fig:seg}.  
% Please see appendix for more details.

\noindent{\bf SBPC (Object Recall):}\label{object-recall-discussion} Table \ref{tab:rec_seed_effect}, ablates the influence of noise seeds (batch size $N$) and number of sampling timesteps $t$ on the prompt adherence (measured by object recall) of images synthesized by the FG-DM for ADE20K validation set prompts (8.5 objects per prompt on average). Avg. Max Recall is the average, over the 2000 prompts, of the maximum per image recall in each batch. Similar definitions hold for Avg.Min and Avg. Median. The top part of the table (columns 2-4) shows that for $N=1$ recall saturates at $69\%$ for $t=10$ timesteps. The bottom part shows that increasing $N$ maintains the Avg. Median recall at this value but produces images with a significant variation of recall values. The Avg. Max recall is $6$ points higher ($75\%$) and fairly stable across configurations of $N$ and $t$. The fifth column shows the number of images in a batch that satisfy the object recall thresholds of 0.5,0.75 and 0.9, averaged over 2000 prompts. While this number decreases for higher thresholds, a batch size of $N=10$ can produce at least one image with even higher prompt compliance, on average.

The FG-DM is particularly well suited to implement SBPC, because recall can be computed as soon as segmentations are sampled. Since the segmentation factor can be run with a smaller number of timesteps and at lower resolution, this is much faster than sampling the images themselves. The image synthesis factor is run only once, for the segmentation mask of highest recall. On the contrary, an SD-based implementation of SBPC requires synthesizing $N$ images and then feeding them to an external segmentation model (we used Segformer-B4 \cite{xie2021segformer} in our experients) to compute recall. Table \ref{tab:compare_rec_hyperparam} compares the object recall of different sampling configurations of SD and FG-DM for ADE20K validation prompts. The objects in the groundtruth masks are considered to compute recall. In appendix section \ref{chatgpt-3.5-extract-objects}, we show that the object classes can be automatically extracted from the prompt using an LLM. The top part of the table compares single run ($N=1$) implementations of SBPC by SD and the FG-DM to the popular IBPC Attend \& Excite (A-E) method~\cite{chefer2023attendandexcite}. The table shows that the FG-DM images have \textbf{8\%} and \textbf{4\%} higher recall than those of SD and A-E respectively, even though SD has higher clip score. This illustrates the deficiencies of clip score to evaluate prompt adherence. The IBPC method underperforms the FG-DM by $4$ points  while drastically increasing inference time to 36 seconds per prompt. This is because A-E computation scales linearly with the number of objects and it fails in multiple object scenarios.
While SBPC achieves good results for $N=1$, the bottom half of the table shows that its true potential is unlocked by larger batch sizes. Both the SD and FG-DM implementations of SBPC achieved the much higher average recall of $\bf 75\%$. However, for SD, the sampling of a batch of $N=10$ high resolution images requires an amount of computation (23 seconds per prompt) prohibitive for most applications. The FG-DM is \textbf{4x (3x) faster} when using 10 (20) DDIM steps for the segmentation mask and 20 steps for image synthesis. 
% Overall, it produces images with $15$ points higher recall than a single run of SD, and it is much faster than either the IBPC method or the SD-based implementation of SBPC.

Besides object recall, the FG-DM can also improve the recall of semantic attributes. We illustrate this with an experiment in face synthesis for FG-DM trained from scratch on MM-CelebA-HQ\cite{xia2021tedigan}. We compute the recall of semantic attributes such as bald, necklace, and pale skin etc. using the generated segmentation masks on validation prompts. Fig. \ref{fig:sem-recall} (left) compares the attribute recall of FG-DM to those of prior methods for $N=1$. The FG-DM has average recall of $\bf 75\%$ over all semantic attributes, outperforming competing methods~\cite{park2023learning}. Figure \ref{fig:sem-recall} (right) shows an histogram of the number of trials ($N$) required by the FG-DM to achieve 70\% and 80\% recall\footnote{10 runs includes samples that required 10 or more trials.}. The FG-DM generates $\approx 90\%$ of its samples with the specified 70\% recall in five trials. 

% \subsection{Ablation Studies} 

\text{\bf Attention distillation loss: } Table \ref{tab:ablationdistill} ablates the adaptation of SD with and without attention distillation loss. The FG-DM with attention distillation improves the image quality by \textbf{1.29/1.3} points for segmentation/normal map conditions respectively. The effect is more pronounced when comparing the fidelity of the conditions which improves the quality by \textbf{5.6}/\textbf{2.8} points for segmentation/normal synthesis showing the effectiveness of the loss in adapting SD to different conditions. Appendix Figure \ref{fig:qual_attn_loss} shows the qualitative comparison of ablating the attention distillation loss.

\begin{figure}[ht]\RawFloats
\centering
\includegraphics[keepaspectratio, width=0.88\columnwidth,trim=30 50 34 88, clip]{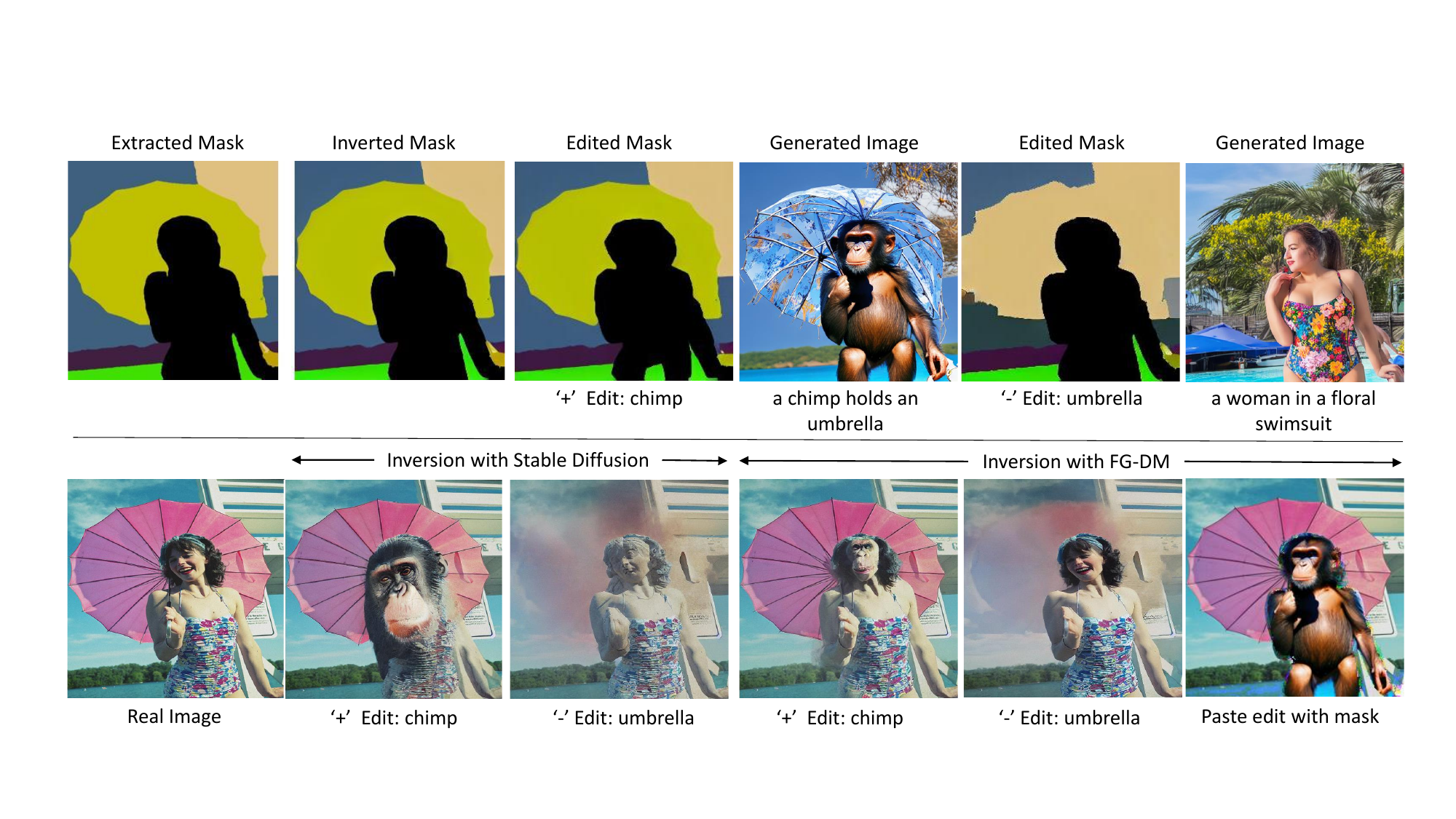}
\caption{\textbf{Top: }Inverting segmentation masks with FG-DM segmentation factor using the LEDITS++ method. Edits to replace the woman by a chimp or eliminate the umbrella. The FG-DM enbables text-based edits to modify or delete objects in a given mask. The image generated with the edited mask as condition is shown to the right of each edited masks. \textbf{Bottom: } Original image, LEDITS++ edited image for stable diffusion and for the image synthesis factor of the FG-DM. Please Zoom in for details.
}
\label{fig:inversion}
\end{figure}

\text{\bf Real Image Editing with FG-DM:}
We show the results of editing of both real images and their segmentation masks with FG-DM. The top of Figure \ref{fig:inversion} refers to inversion of the segmentation mask. We use an off-the-shelf OpenSEED \cite{zhang2023simpleopenseed} model to extract the segmentation map of a real image (shown on the bottom left of the figure) and apply the FG-DM segmentation factor model for inversion and editing using LEDITS++ \cite{brack2024ledits}, a recent method for text based image editing. We apply LEDITS++ to the segmentation factor to 1) replace the mask of the woman by that of a chimp (third image of the top row) and 2) to delete the mask of the umbrella (fifth image). New images (fourth and sixth) are then generated by the image synthesis factor conditioned on the edited segmentation masks. We have found that the inversion and editing of segmentation masks is quite robust. The synthesized masks usually reflect the desired edits. However, because the final image synthesis is only conditioned on these masks, the synthesized image does not maintain the background of the original image. The synthesized image is a replica of the original image at the semantic level (similar objects and layout) but not at the pixel level. From our experiments, this method has high robustness and quality for semantic-level editing. 

We next investigated pixel level inversion and editing, which is harder.  The bottom part of Figure \ref{fig:inversion} shows the comparison of LEDITS++ editing with inversion by SD and by the image synthesis factor of the FG-DM. For the latter, we apply inversion to the ControlNet image generation factor using the real image and the segmentation mask extracted from it. Then we perform the LEDITS++ edit using the edited mask from the top part of Figure \ref{fig:inversion} (inverted with the FG-DM segmentation factor) to produce the edited image as shown in columns 4 and 5. This pixel-level inversion and editing tends to maintain the background of the original image but is much less robust than mask-level editing in terms of editing quality. This can be seen from the images in columns 2 and 3, which show the inversion using SD, which fails to produce a realistic chimp and turns the woman into a stone sculpture. The FG-DM produces much more meaningful edits, as shown in columns 4 and 5. The last column of the bottom part of the Figure \ref{fig:inversion} shows an added advantage of FG-DM where the chimp generated in the top portion can be pasted to the original image due to availability of the segmentation mask. In this example the pasting is rough around the object edges since we have made no attempts to beautify it. It can be improved by denoising the generated image with one forward pass of SD at a higher timestep.
 % The pixel-level inversion and the copy and past technique can be improved further . 

% See appendix section \ref{supp-ablation} for ablation studies on data augmentation, order of conditions, joint modeling, sequential vs joint training/inference and image-mask alignment.

\section{Limitations, Future Work and Conclusion}
Although, the FG-DM uses low resolution synthesis for the conditions, runtime increases for chains with more than two factors. Since, the attention maps generated per factor must be consistent according to the joint model, sharing them across different factors and timesteps is a promising direction for further reducing the runtime. Furthermore, while the FG-DM allows easier control over generated images for operations like deleting, moving, or flipping objects, fine-grained manipulations (e.g. changing the branches of a tree) can still require considerable user effort. Automating the pipeline by using an LLM or methods like Instructpix2pix~\cite{brooks2022instructpix2pix} to instruct the edits of the synthesized conditions is another interesting research direction. See \ref{broader-impact} for a discussion on broader impact. 

In this work, we proposed the FG-DM framework for efficiently adapting SD for improved prompt compliance and controllable image synthesis. We showed that an FG-DM trained with relatively small datasets generalizes to prompts beyond these datasets, supports fine-grained image editing, enables improved prompt compliance by SBPC, allows adding new conditions without having to retrain all existing ones, and supports data augmentation for training downstream models. It was also shown that the FG-DM enables faster and creative image synthesis, which can be tedious or impossible with existing conditional image synthesis models. Due to this, we believe that the FG-DM is a highly flexible, modular and useful framework for various image synthesis applications.

\section*{Acknowledgements}
This work was partially funded by the NSF award IIS-2303153. We also acknowledge and thank the use of the Nautilus platform for some of the experiments discussed above.

{\small
\bibliographystyle{splncs04}
\bibliography{main}
}

%%%%%%%%%%%%%%%%%%%%%%%%%%%%%%%%%%%%%%%%%%%%%%%%%%%%%%%%%%%%
\newpage
\appendix

\section{Appendix}

% \subsection{Code Release}
% Code and trained models will be released upon acceptance of the paper with the necessary safety filters for images.

\begin{figure}[ht]\RawFloats
\centering
\includegraphics[keepaspectratio, width=\columnwidth,trim=10 20 34 20, clip]{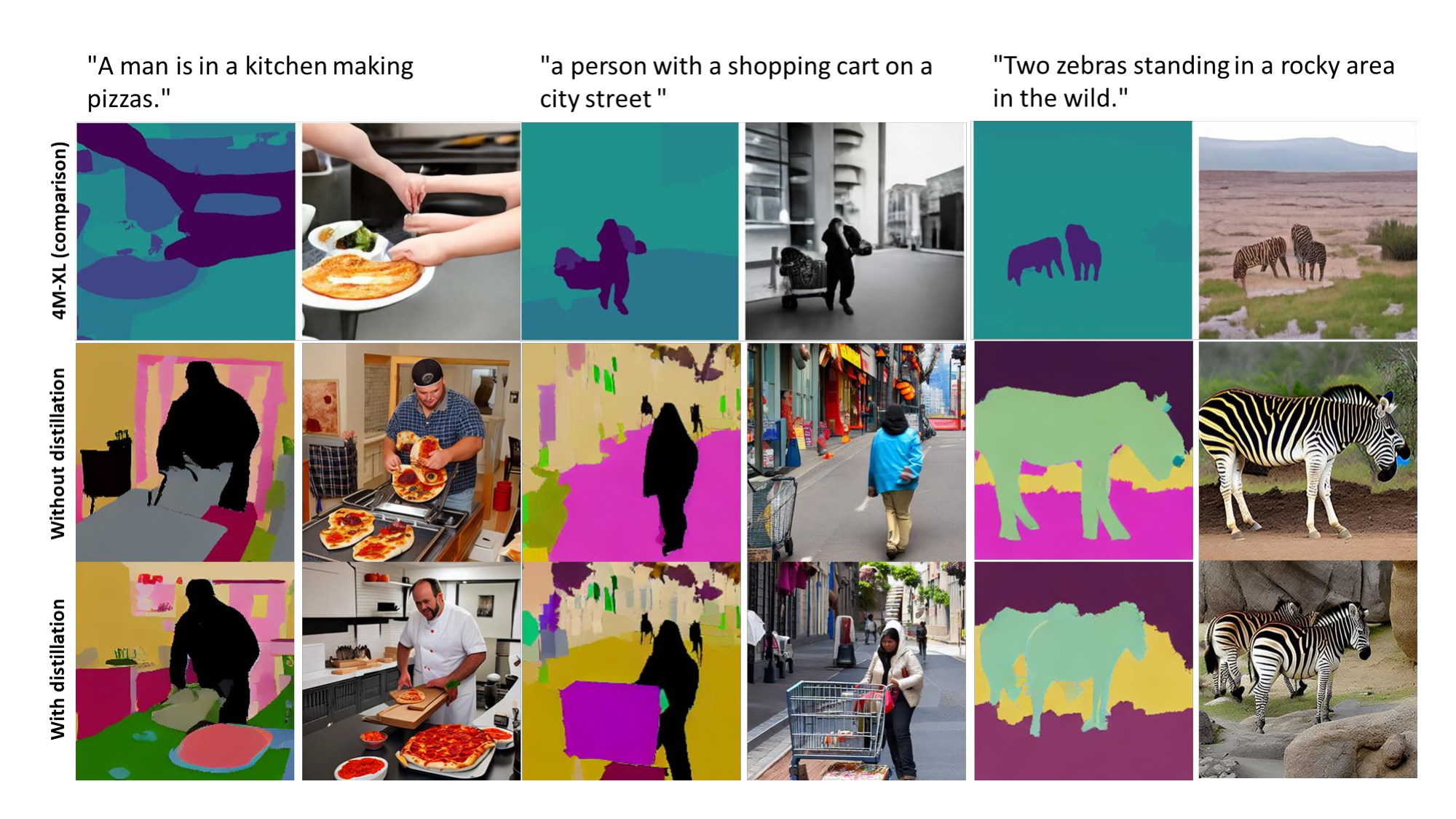}
\caption{Comparison of FG-DM with attention distill loss (bottom) against FG-DM without attention distill loss (center) and the recent 4M-XL model (top) for the prompts shown at the top of the figure. The two versions of the FG-DM use the same seed. Both versions of the FG-DM produce images of higher quality than 4M-XL. Attention distillation helps improve the quality of the generated segmentations. For example, the model without distillation has inaccurate masks/missing cart/less realistic zebras from left to right.
}
\label{fig:qual_attn_loss}
\end{figure}

\begin{figure*}[h]\RawFloats
\centering
\includegraphics[width=\columnwidth, trim=0 68 0 68, clip]{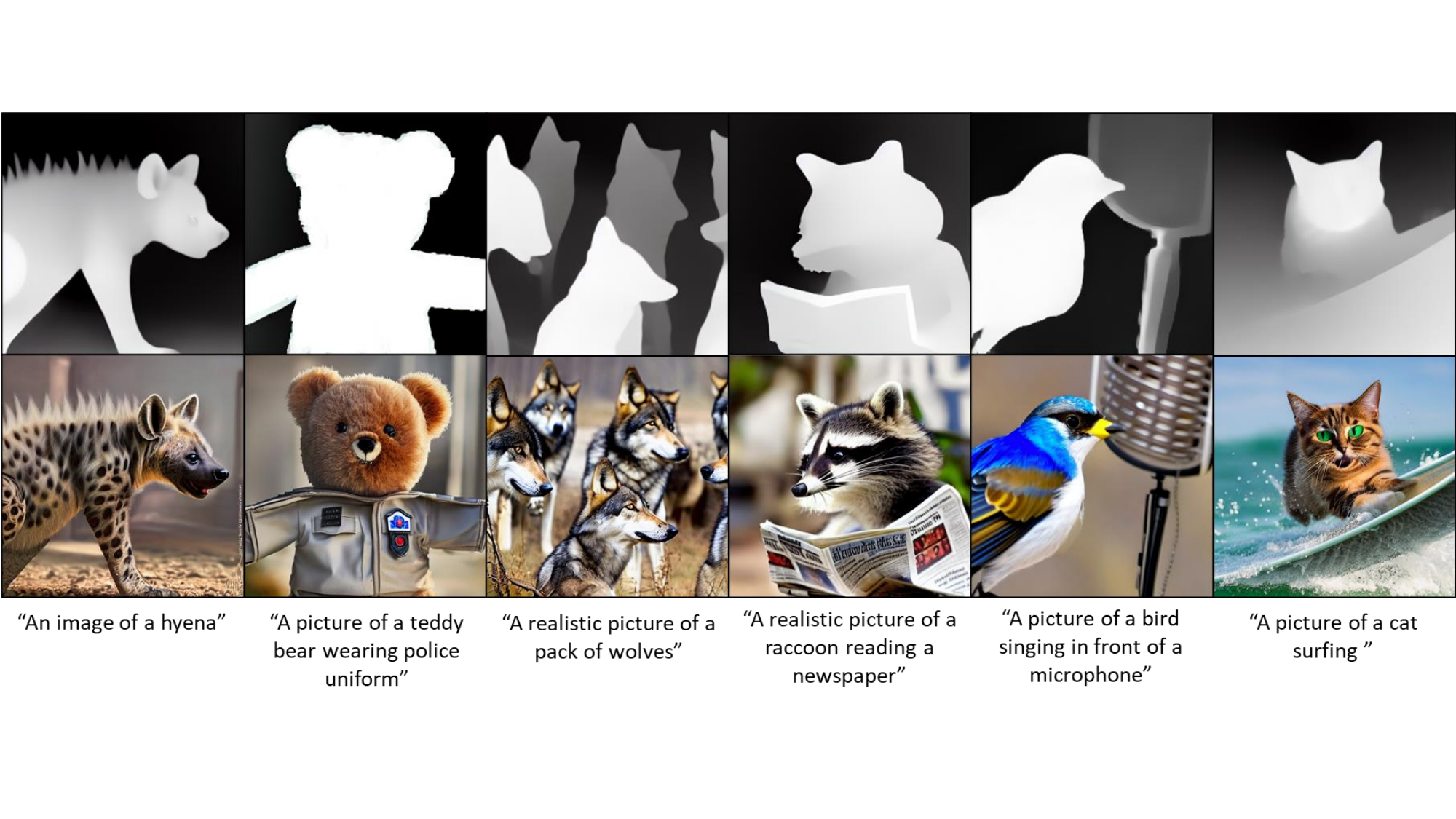}
\caption{\label{fig:depth-supp} \textbf{Qualitative results} of Depth map/Image pairs synthesized by FG-DM. }
\end{figure*}

\begin{figure*}[h]\RawFloats
\centering
\includegraphics[width=\columnwidth, trim=0 68 0 68, clip]{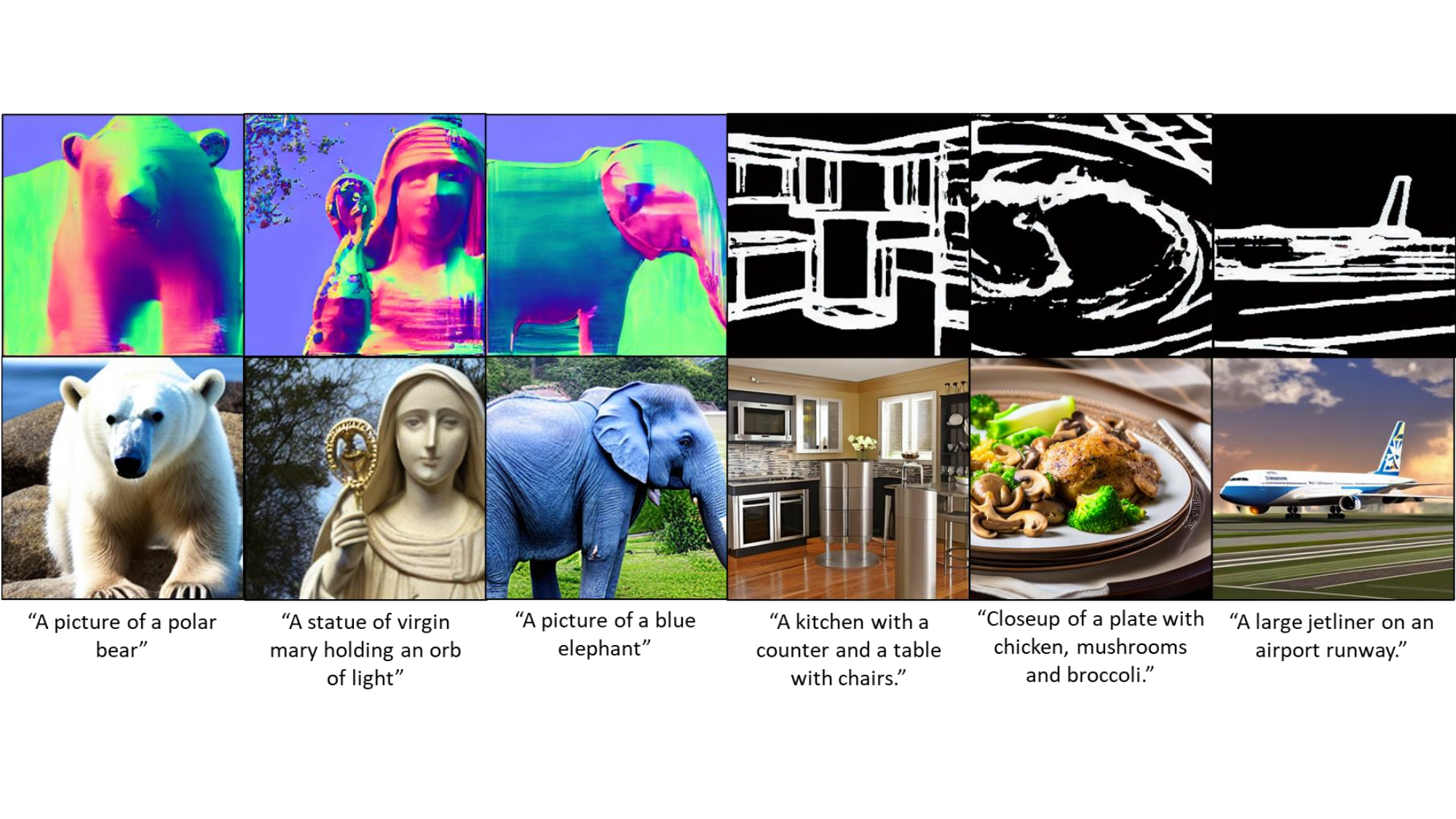}
\caption{\label{fig:normal} \textbf{Qualitative results} of Normal map/image and Sketch map/image pairs synthesized by FG-DM. FG-DM generalizes well across conditions and is able to generate condition-image pairs that are not seen during training.}
\end{figure*}

\begin{figure*}[h]\RawFloats
\centering
\includegraphics[width=\columnwidth, trim=0 68 0 68, clip]{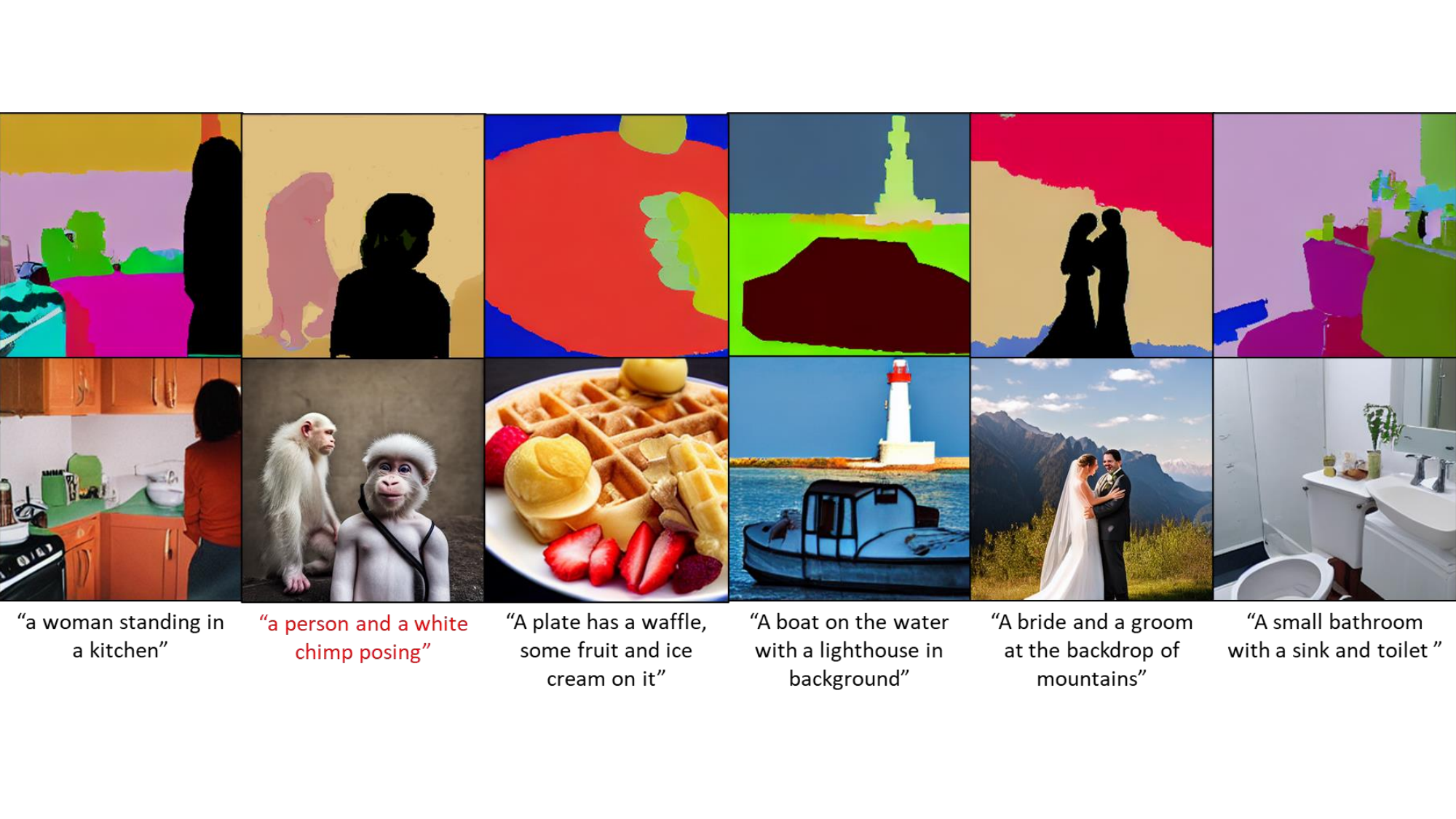}
\caption{\label{fig:seg-supp} \textbf{Qualitative results} of Segmentation map/image pairs synthesized by FG-DM. The second column shows the benefit of explainability with FG-DM which allows verifying intermediate conditions to understand
the hallucinations (mixup of a chimp and a person) of SD which are opaque otherwise. Here, FG-DM correctly generates the chimp and the person mask with different colors while ControlNet confuses between the two showing that the ControlNet needs to be corrected.}
\end{figure*}

\begin{figure*}[h]\RawFloats
\centering
\includegraphics[keepaspectratio, width=\columnwidth, trim=240 164 288 128, clip]{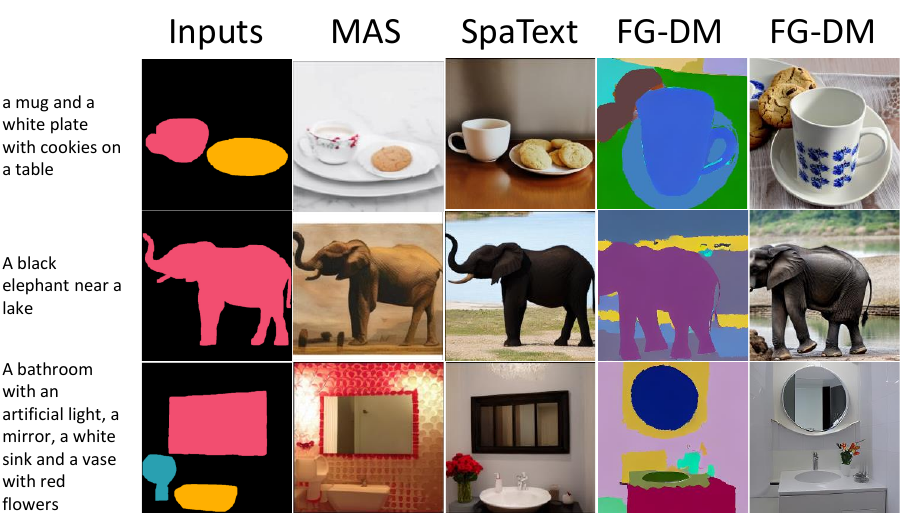}
\caption{\label{comparison_with_mas}\textbf{Qualitative comparison of FG-DM with prior works} such as Make-a-Scene and SpaText for the prompts shown on the left. Note that FG-DM generates both the map and the image while for MAS and SpaText, the condition was manually sketched and fed to the model.}
\label{fig:misc_comparison}
\end{figure*}

\begin{figure*}[h]\RawFloats
\centering
\includegraphics[width=\columnwidth, trim=0 28 0 28, clip]{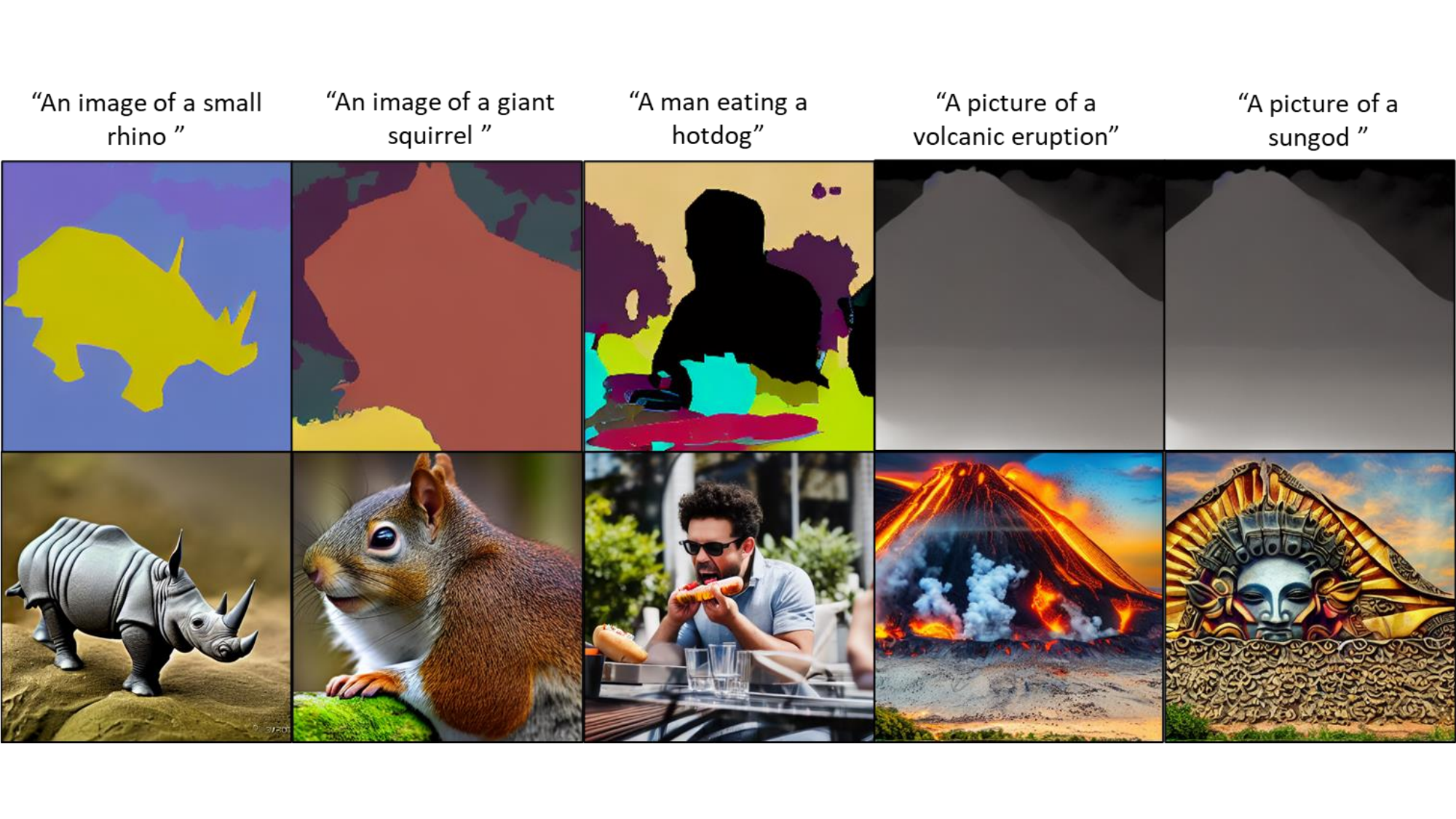}
\includegraphics[width=\columnwidth, trim=0 28 0 28, clip]{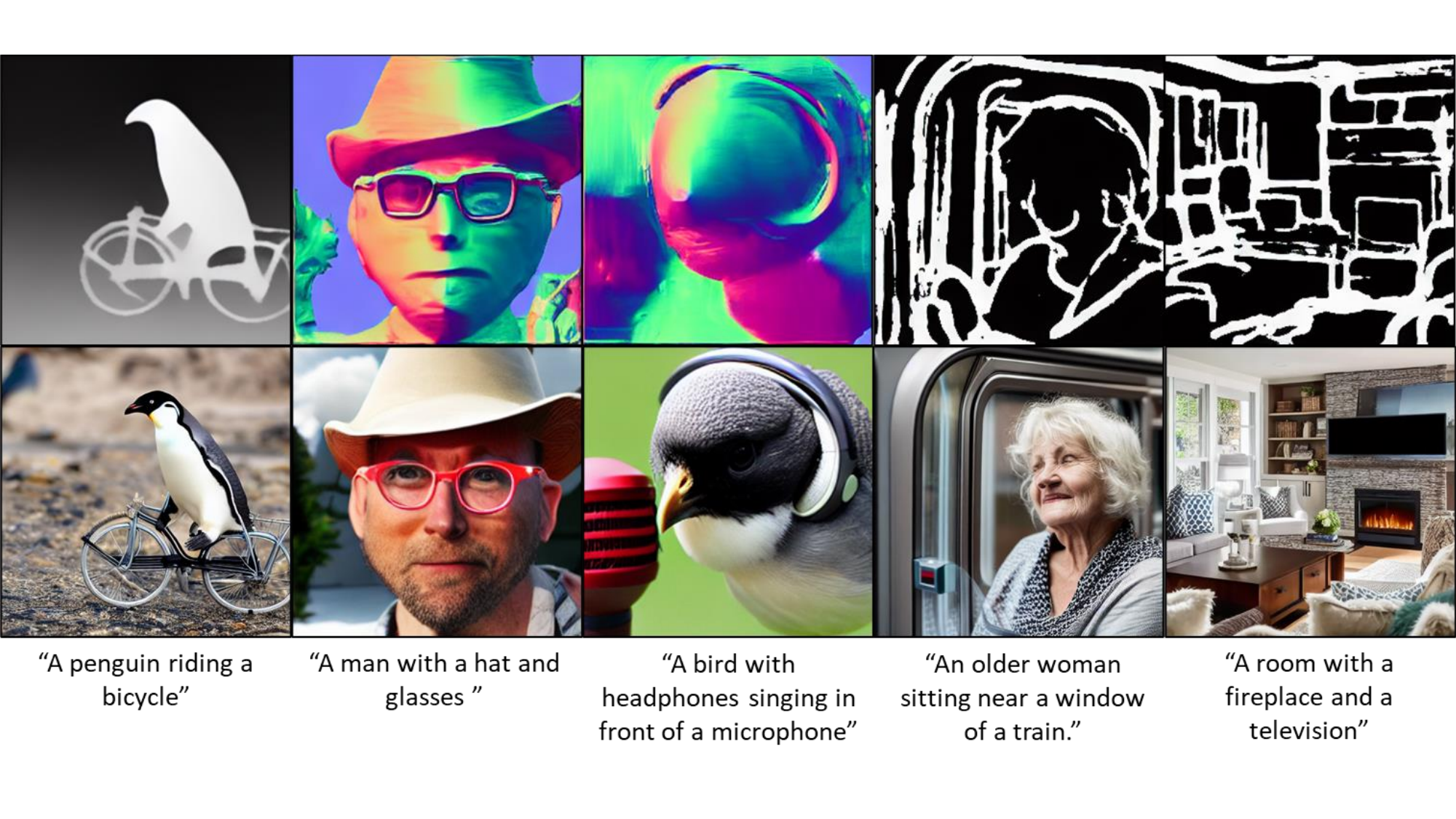}
\caption{\label{fig:zoom} \textbf{Zoomed version of Figure \ref{fig:seg}} Qualitative results of FG-DM to synthesize segmentation, depth, normal and sketch maps and their corresponding images.}
\end{figure*}

\subsection{Experimental Results}

\subsubsection{FG-DMs Adapted from Stable Diffusion (SD)}
We start by the discussion of additional results for FG-DMs obtained by adaptation of Stable Diffusion model, as shown in Figure \ref{fig:arch} of the paper.

\noindent\textbf{ Qualitative comparison of attention distillation loss}

Figure \ref{fig:qual_attn_loss} shows some qualitative results of the ablation for the impact of the attention distillation loss.  There is a clear qualitative benefit in introducing this loss. Without it, the model generates less accurate masks, leading to an unrealistic pizza making depiction/ cart-person relationship/ zebra pair from left to right. This confirms the qualitative ablation showing the benefits of the attention distillation loss in Table \ref{tab:ablationdistill} but provides a stronger illustration of the advantages of the loss, which tends to produce more "plausible" scenes. Such plausibility is difficult to measure with qualitative metrics. For example, the CLIP score is not sensitive to the fact that the cart and the person are not interacting in a normal way, or that the pizza making activity is unrealistic.

We also compared with the recent 4M model \cite{4m}, an autoregressive model trained from scratch on both discriminative and generative tasks. In this case, we use the largest model (4X-ML) released by the authors. Figure \ref{fig:qual_attn_loss} shows a qualitative comparison between FG-DM and 4M. It can be seen that 4M generates images of weaker quality  (distorted hands, missing person's head, deformed zebra bodies) as compared to FG-DM with/without the attention distillation loss.

\subsubsubsection{\textbf{Additional Qualitative Results for Segmentation, Depth, Normal and Sketch conditions}}

Figure \ref{fig:depth-supp} shows qualitative results of synthesized depth maps and images for the creative prompts shown on top/bottom of each image. The FG-DM framework is able to generate high quality images and normal maps for prompts that are not seen in the training setleveraging the generalization of SD. 

Figure \ref{fig:normal} (first three columns) shows qualitative results of synthesized normal maps and images for the creative prompts shown on top/bottom of each image. Figure \ref{fig:normal} (last three columns) shows qualitative results of synthesized sketch maps and images for the creative prompts shown on top/bottom of each image. 

Figure \ref{fig:seg-supp} shows qualitative results of synthesized segmentation maps and images for the creative prompts shown on top/bottom of each image. As shown in the main paper, the FG-DM is able to synthesize segmentation maps for object classes beyond the training set and the semantic maps are colored with unique colors, allowing the easy extraction of both object masks and class labels. This shows the potential of the FG-DM for open-set segmentation, e.g the synthesis of training sets to generate training data for segmentation models. Further, a number of interesting generalization properties emerge. Although the FG-DM is only trained to associate persons with black semantic maps segments, it also assigns the chimp of Figure~\ref{fig:teaser}, a class that it was not trained on, to that color. This shows that the FG-DM can integrate the prior knowledge by SD that ``chimps and persons are similar" into the segmentation task, which receives supervision from COCO alone. Conversely, the similarity between chimps and people might induce SD to synthesize a chimp in response to a prompt for people, or vice-versa. This is shown in the bottom left of Figure \ref{fig:seg-supp} where the FG-DM correctly synthesizes different colors for the chimp and the person, but the ControlNet fails. While the black box nature of SD makes these errors opaque, the FG-DM allows inspection of the intermediate conditions to understand these hallucinations and make corrections accordingly. For example, in the above example ControlNet has to be corrected by either finetuning or using inference optimization methods like A-E~\cite{chefer2023attendandexcite}. This illustrates its benefits in terms of explainability.

\subsubsubsection{\textbf{Comparison with prior works}}

Figure \ref{fig:misc_comparison} shows the qualitative comparison of FG-DM with prior works such as Make-a-Scene \cite{makeascene} or SpaText \cite{Avrahami_2023_CVPR} in addition to 4M model compared in Figure \ref{fig:qual_attn_loss}. Note that FG-DM generates both the segmentation and the image while for the other methods, it is manually sketched and fed to them. It is seen that FG-DM generates high quality images that adhere well to the prompts as compared to the prior works.

\subsubsubsection{\textbf{Comparison of generated conditions by FG-DM to conditions recovered by off-the-shelf models} }

\begin{figure}[h]\RawFloats
\centering
    \begin{subfigure}{\columnwidth}
        \centering
        \includegraphics[width=0.8\columnwidth]{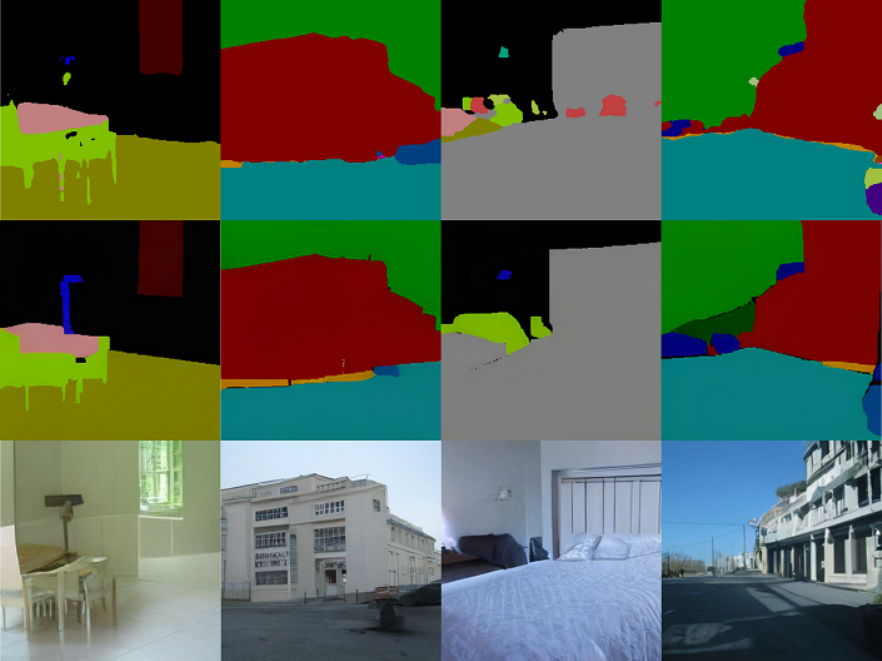}
        \caption{\label{gen_ext_sem} Segmentation maps. (Middle: Generated, Top: Extracted)}
    \end{subfigure}
    \hfill
    \begin{subfigure}{\columnwidth}
        \centering
        \includegraphics[width=0.8\columnwidth]{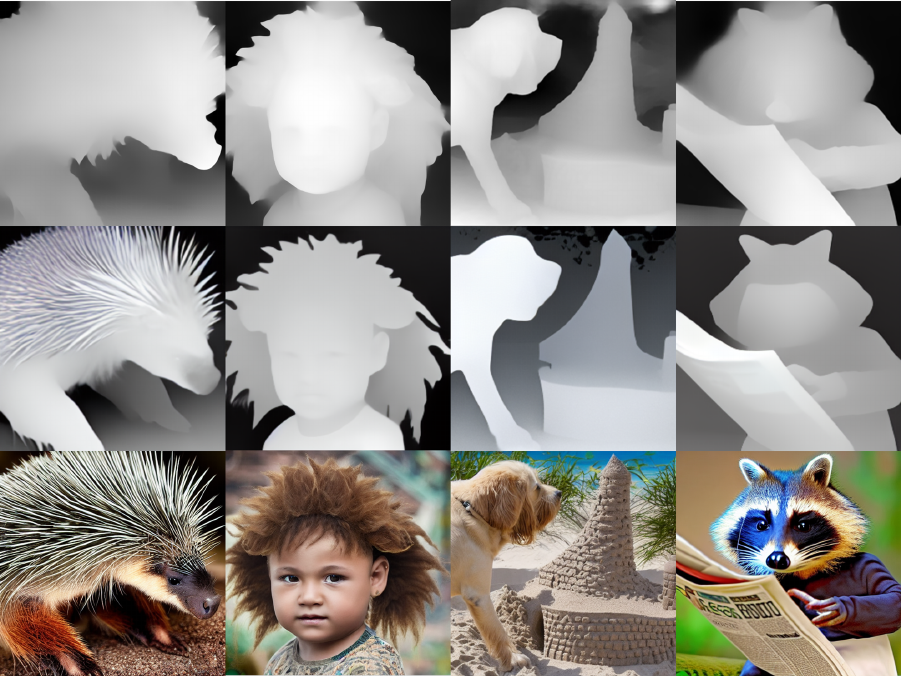}
        \caption{\label{gen_ext_depth} Depth maps. (Middle: Generated, Top: Extracted)}
    \end{subfigure}
    \caption{\label{fig:qual_cond_gen_vs_extract} Qualitative comparison of generated conditions for FG-DM vs extracted conditions using SD (Stable Diffusion) + CEM (Condition Extraction Model) for segmentation and depth maps. The generated conditions for depth maps are superior to the extracted ones.}
\end{figure}

Figure \ref{fig:qual_cond_gen_vs_extract} shows a qualitative comparison of the conditions synthesized by FG-DMs to those recovered from the synthesized image using off-the-shelf pretrained models for segmentation and depth estimation. The qualitative results corroborate with the user study as the generated conditions are better than extracted ones for depth while they are similar for segmentation conditions. 

\subsubsubsection{\textbf{Synthesis of Conditioning Variables with SD Autoencoder}} \label{synthesis-conditions}

\begin{figure}[h]\RawFloats
\centering
    \begin{subfigure}{\columnwidth}
        \centering
        \includegraphics[width=0.48\columnwidth]{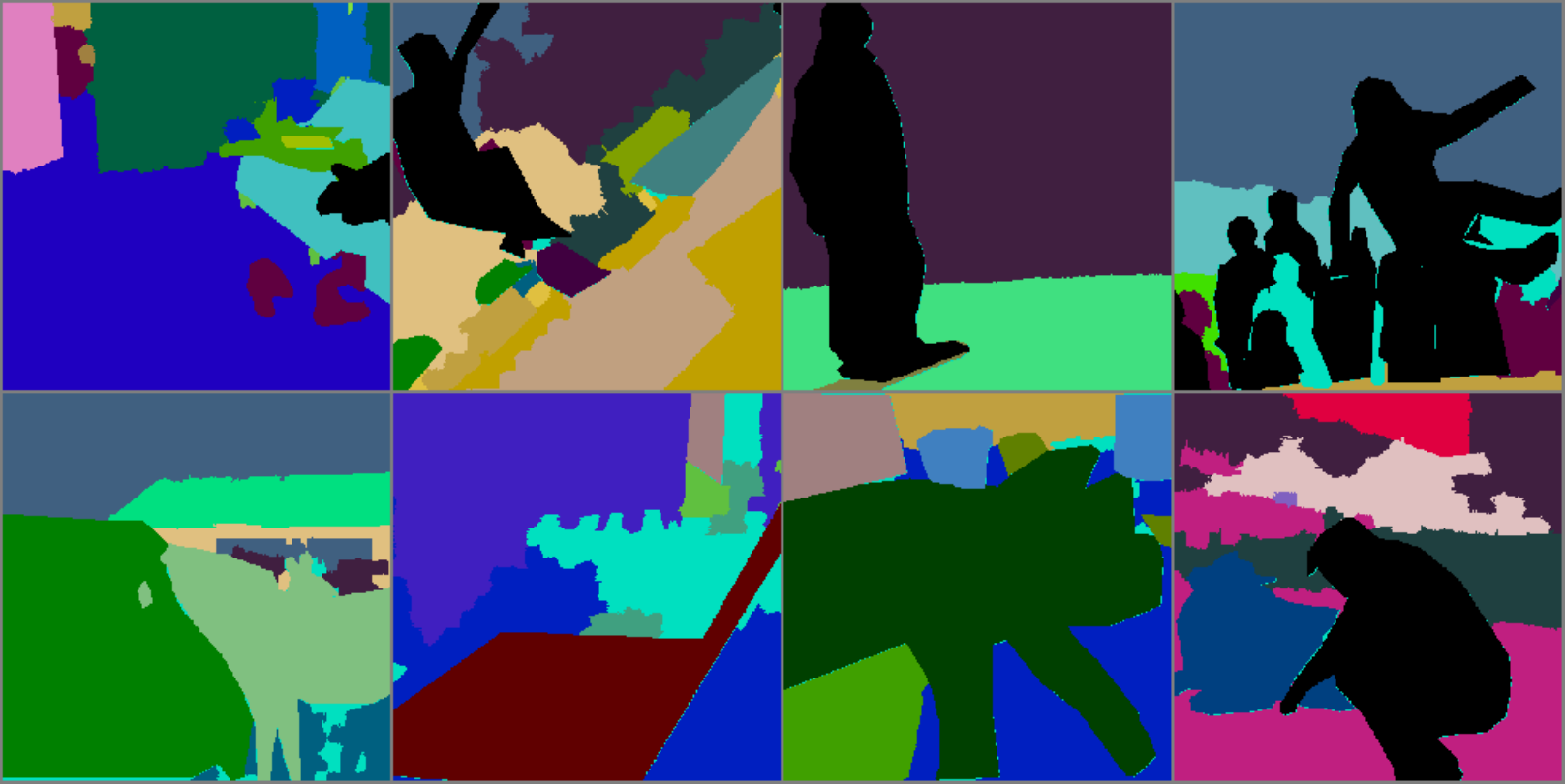}
        \includegraphics[width=0.48\columnwidth]{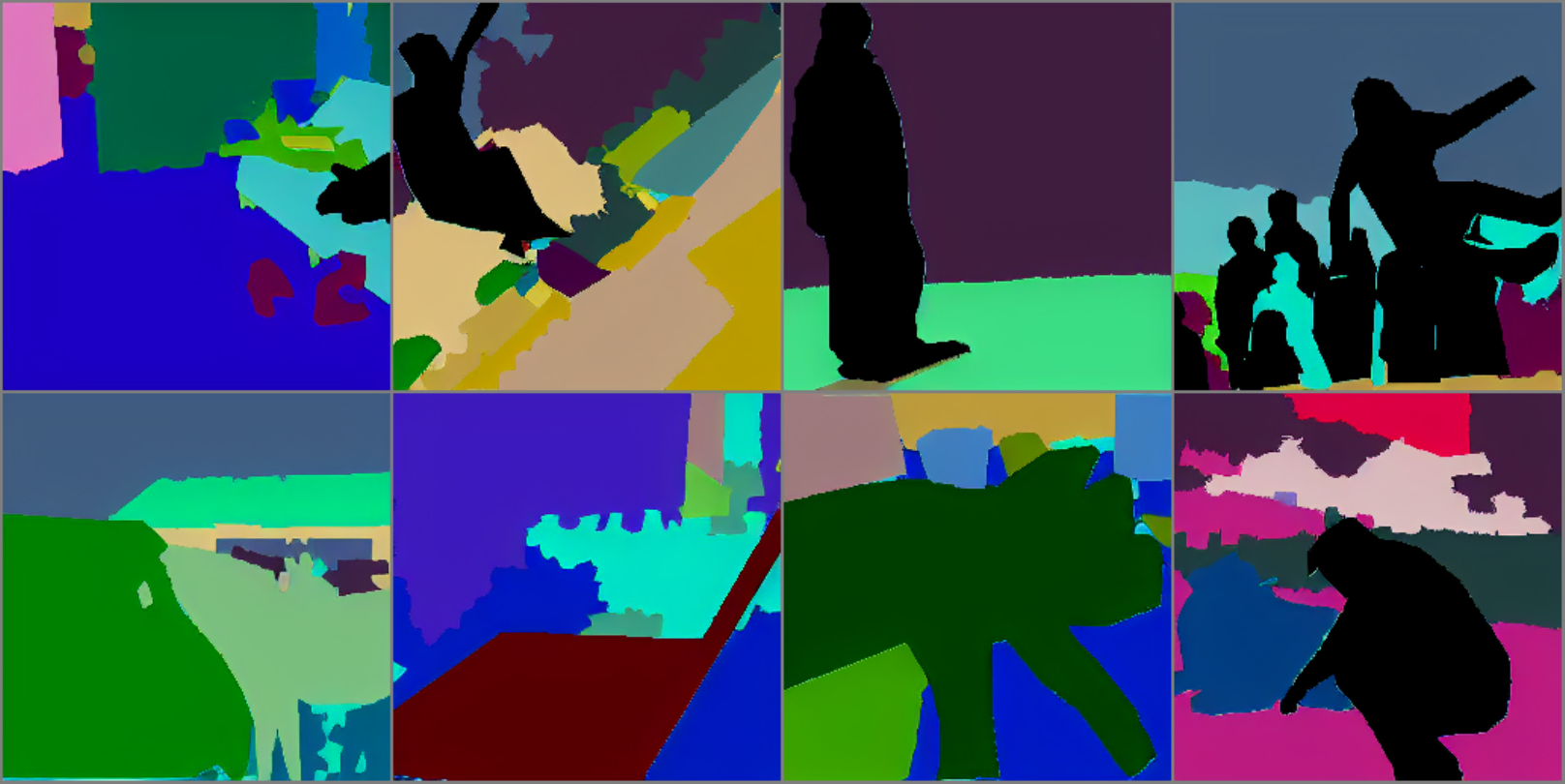}
        \caption{\label{recon_sem} Segmentation maps. (Left: Groundtruth, Right: Reconstructed)}
    \end{subfigure}
    \hfill
    \begin{subfigure}{\columnwidth}
        \centering
        \includegraphics[width=0.48\columnwidth]{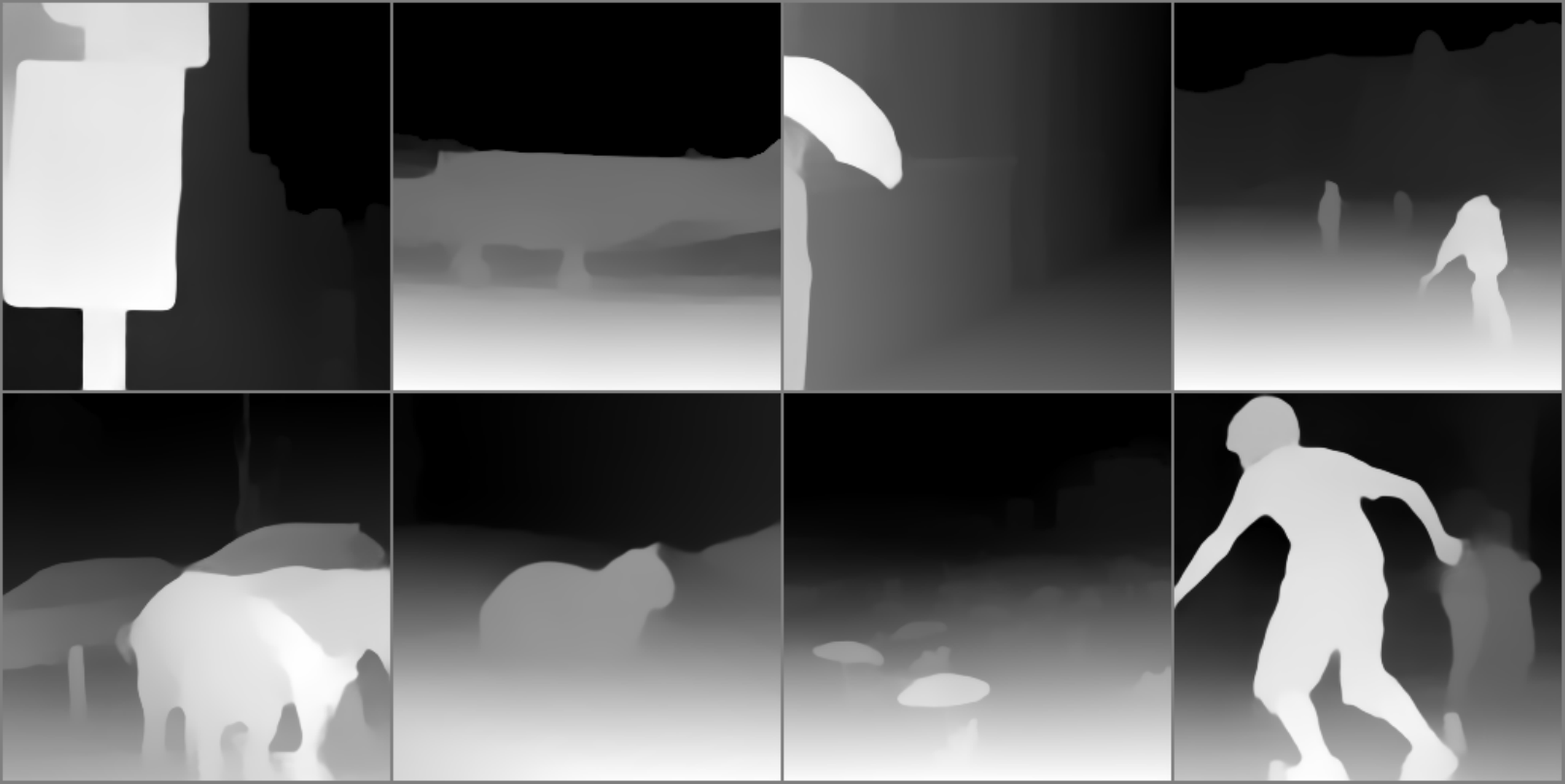}
        \includegraphics[width=0.48\columnwidth]{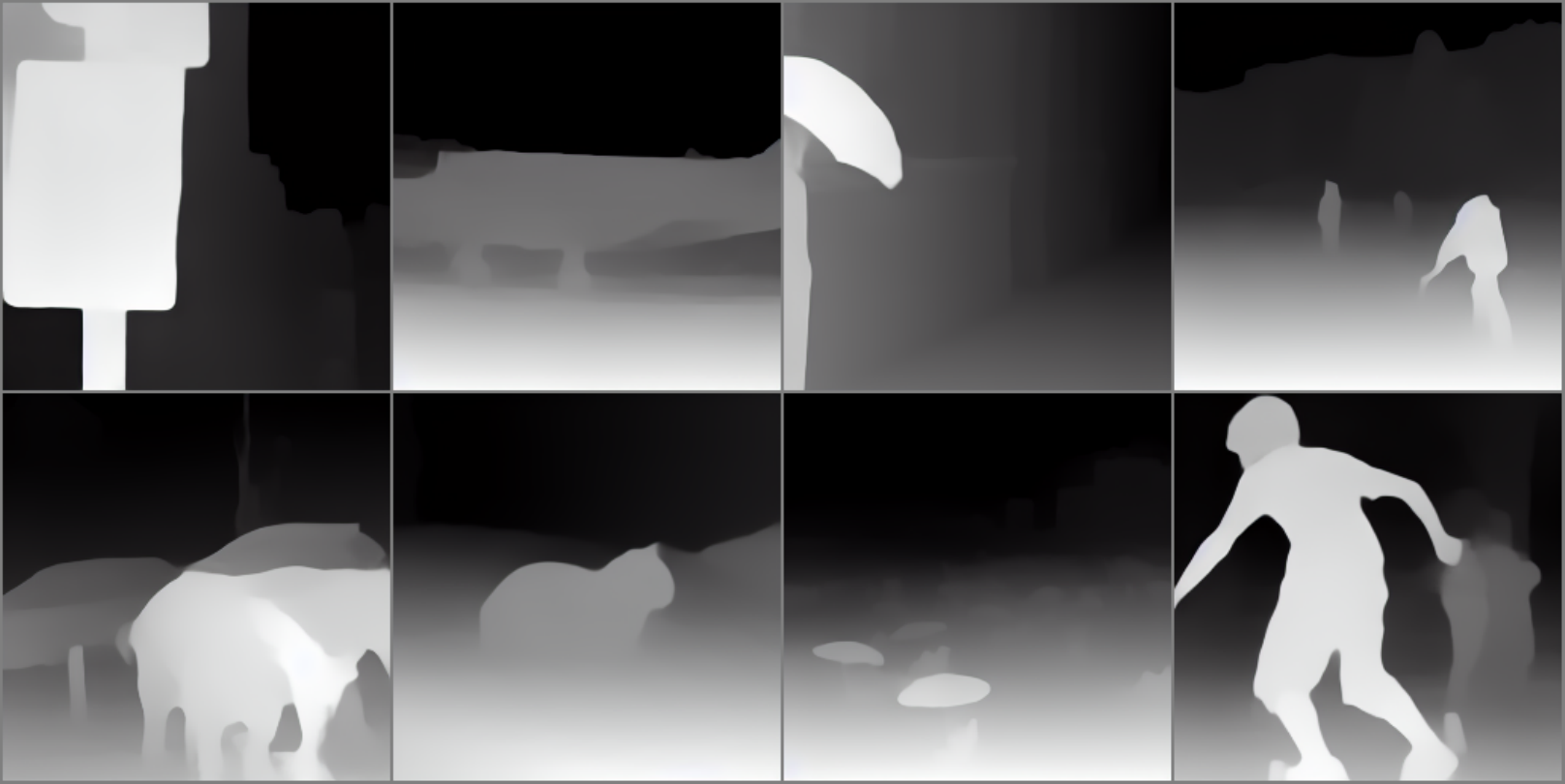}
        \caption{\label{recon_depth} Depth maps. (Left: Groundtruth, Right: Reconstructed)}
    \end{subfigure}
    \hfill
    \begin{subfigure}{\columnwidth}
        \centering
        \includegraphics[width=0.48\columnwidth]{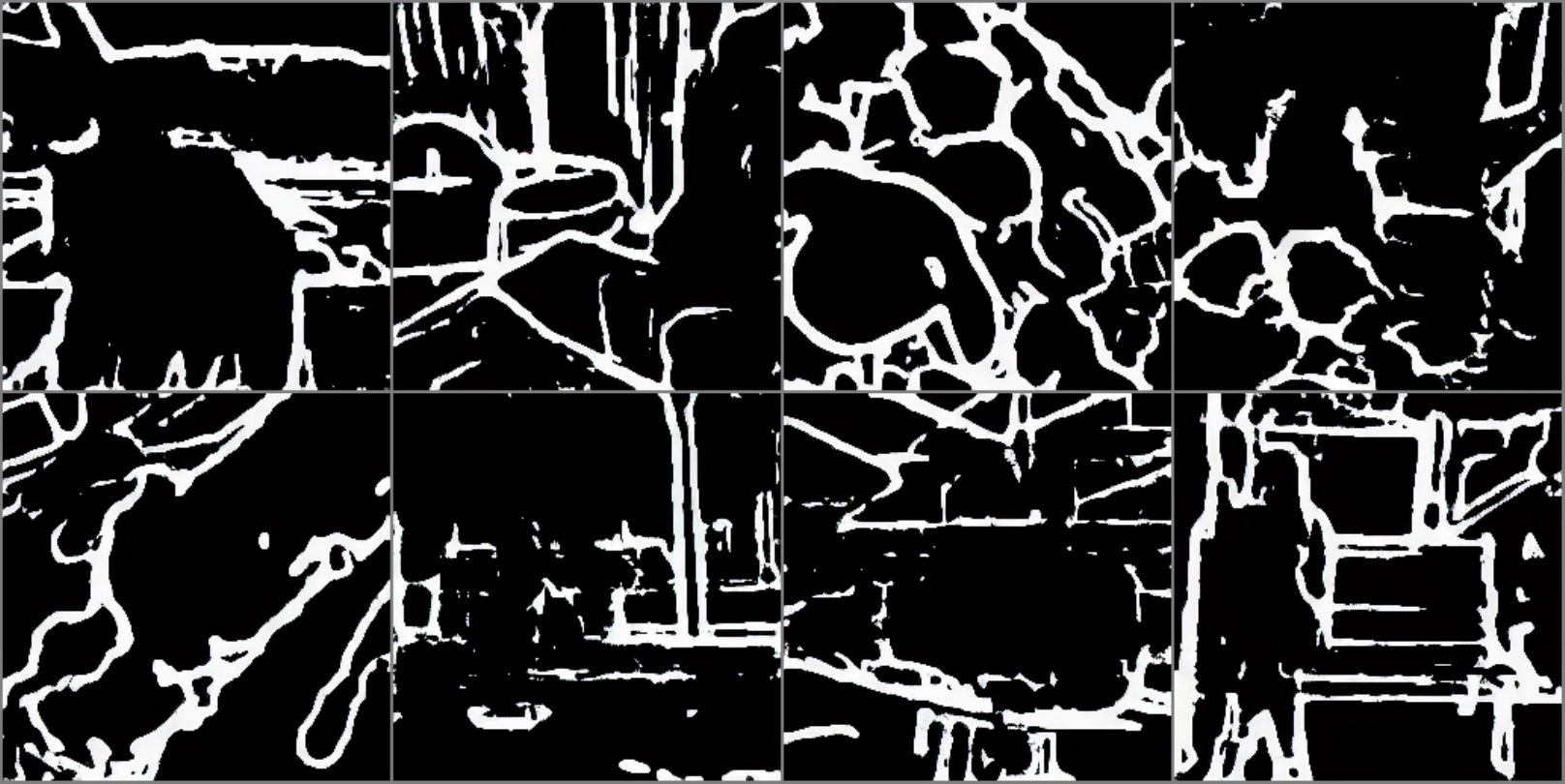}
        \includegraphics[width=0.48\columnwidth]{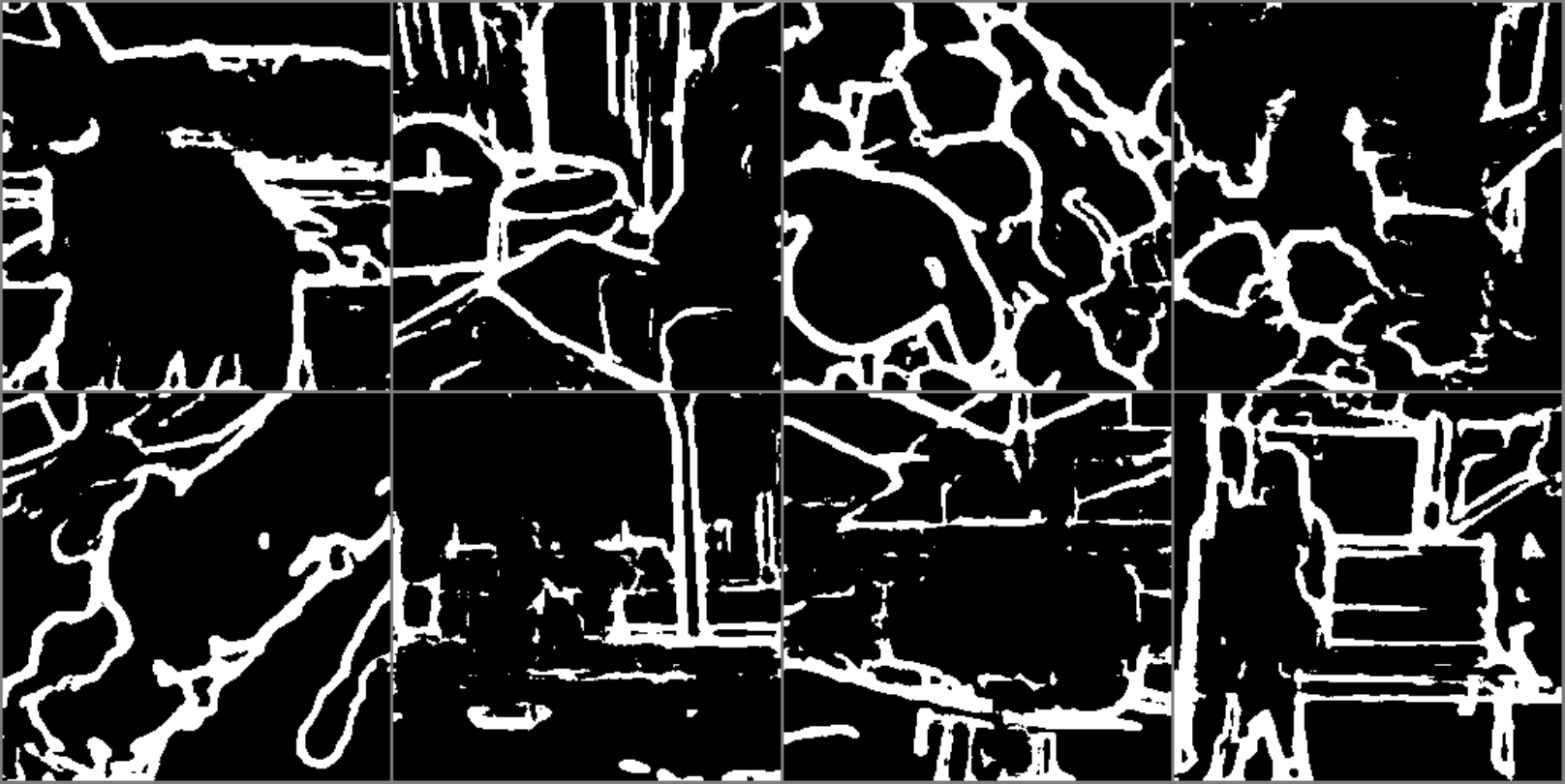}
        \caption{\label{recon_sketch} Sketch maps. (Left: Groundtruth, Right: Reconstructed)}
    \end{subfigure}
    \hfill
    \begin{subfigure}{\columnwidth}
        \centering
        \includegraphics[width=0.48\columnwidth]{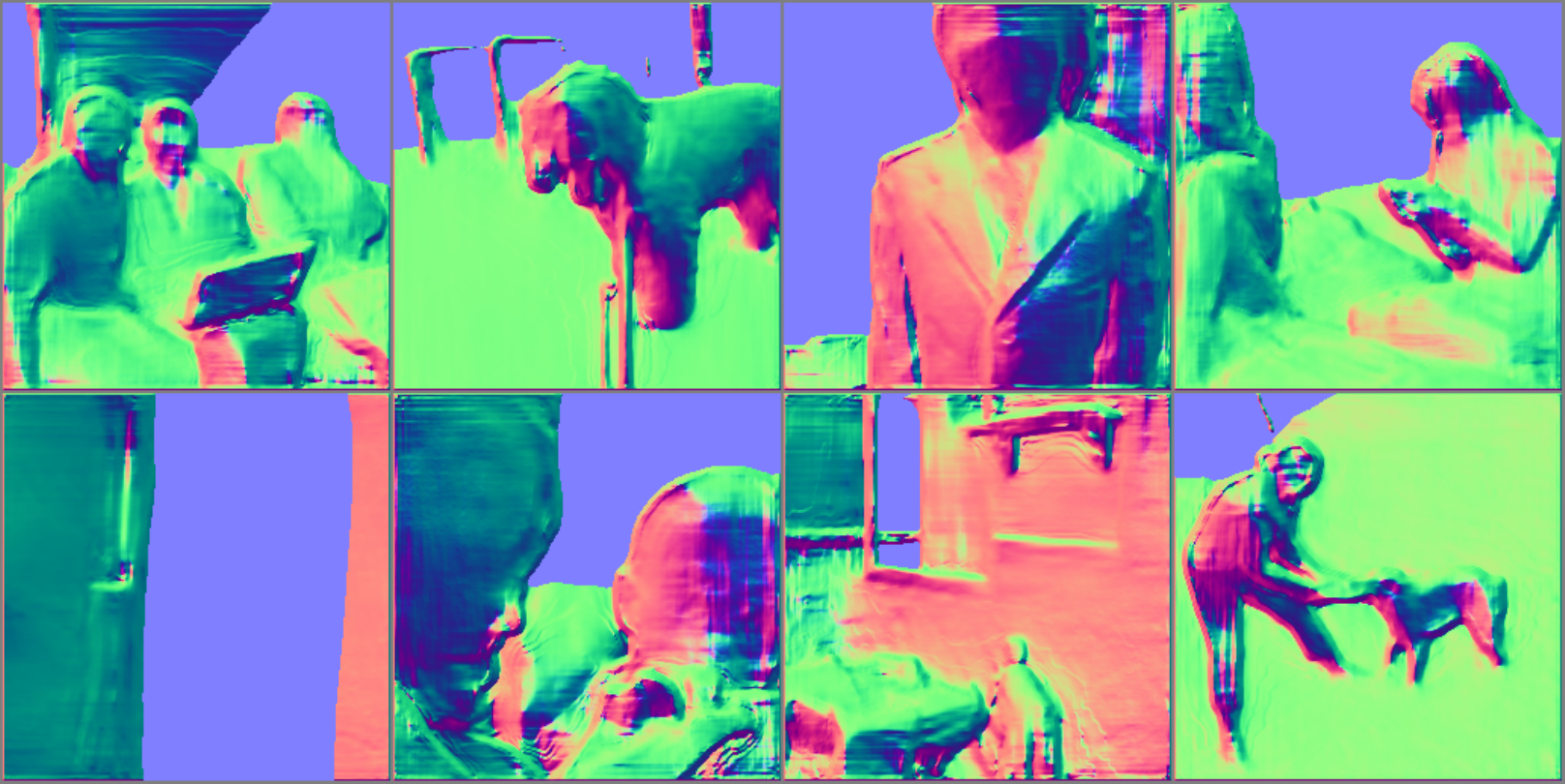}
        \includegraphics[width=0.48\columnwidth]{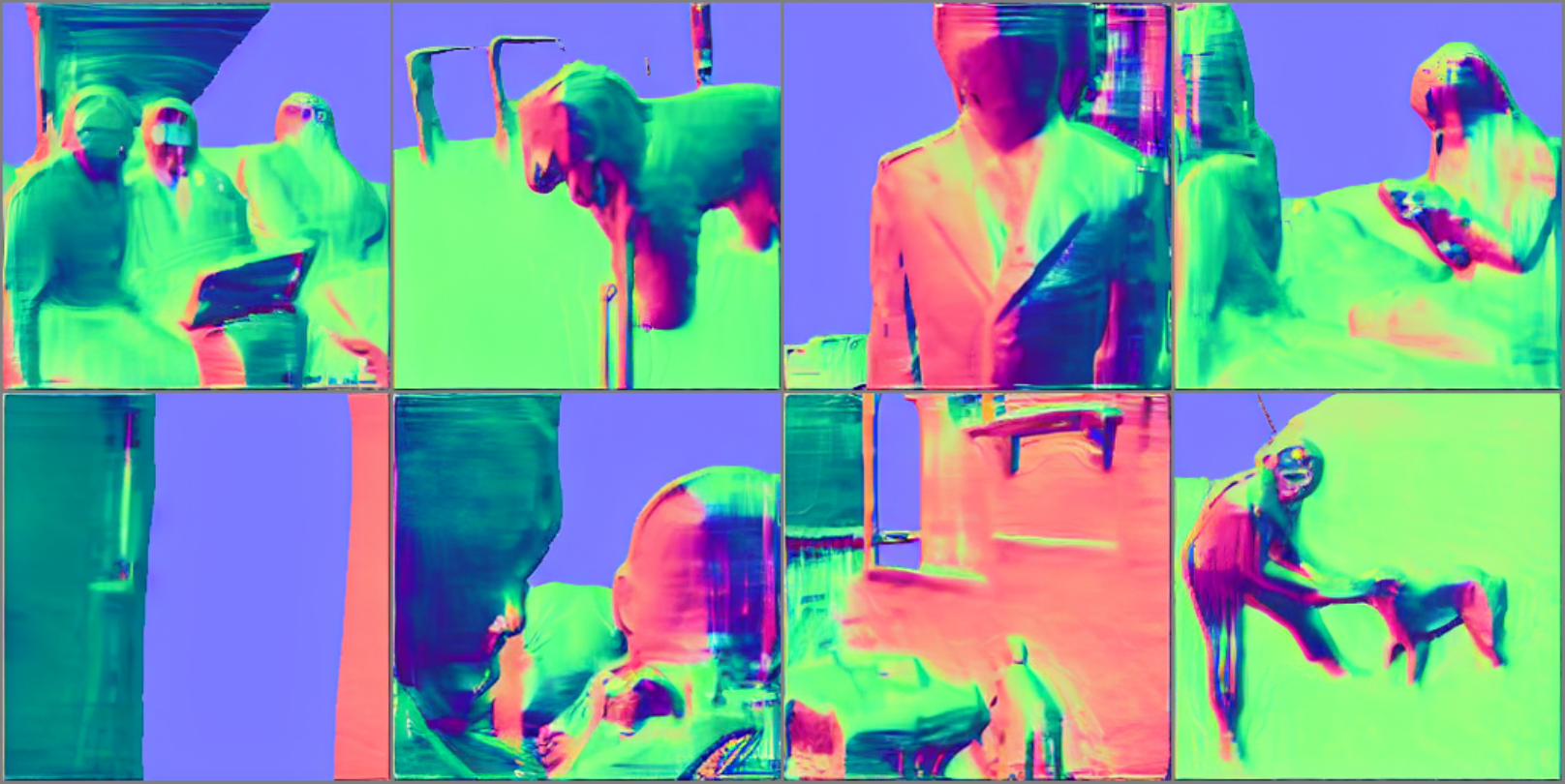}
        \caption{\label{recon_normal} Normal maps. (Left: Groundtruth, Right: Reconstructed)}
    \end{subfigure}
    \caption{\label{fig:qual_map_recon} Visualization of groundtruth (left) and reconstructed (right) maps by applying the pretrained stable diffusion autoencoder to segmentation, depth, sketch and normal maps.}
\end{figure}

High-quality image synthesizes requires DMs trained from large datasets. A common solution is to adopt the LDM~\cite{rombach2022high} architecture, where an encoder ($\cal E$)-decoder ($\cal D$) pair is used to map images into a lower dimensional latent space. Using a pre-trained $\cal E$-$\cal D$ pair, e.g. from the SD model, guarantees that latent codes map into high quality images, making it possible to train high-fidelity DMs with relatively small datasets. However, it is unclear that this approach will work for the synthesis of conditioning variables, such as segmentation maps, which SD is not trained on. For example,~\cite{lemoing2021semanticpalette,park2023learning} explicitly address training DMs to produce the discrete outputs required by many conditioning variables.
Somewhat surprisingly, our experiments show that off-the-shelf foundation DMs are quite effective at synthesizing visual conditioning variables. All our results use the following procedure: 1) visual conditioning variables are converted to 3 channel inputs. For discrete variables, a different color is simply assigned to each variable value (e.g. each semantic label of a segmentation map). 2) All VC-DMs in the FG-DM re-use the pre-trained encoder-decoder pair of SD~\cite{rombach2022high}, as illustrated in Figure~\ref{fig:arch}. 3) At the decoder output, discrete values are recovered with a hashmap that matches image colors to their color prototypes. This is done by simple thresholding, with a margin threshold empirically set to 28 for each pixel. To test this procedure, we measured the mean squared pixel reconstruction error of the segmentation maps from 2,000 validation images of the ADE20K dataset. This was 0.0053 (normalized) or 1.34 pixel error showing that the pretrained SD autoencoder is highly effective in reconstructing discrete maps. 
Figure \ref{fig:qual_map_recon} compares the auto-encoder reconstructed maps against the groundtruth maps for different conditioning variables such as segmentation, depth, sketch and normal. This shows that simply representing the conditioning variables as 3-channel inputs allows the faithful reconstruction of all conditions. 
Further, we also trained a lightweight segmentation model to recover the labels from the RGB semantic maps but observed no improvement over this simple heuristic.

\begin{table*}[h]\RawFloats
\centering
\scriptsize
\caption{Comparison of segmentation mask quality and throughput of FG-DM trained on ADE20K dataset against  state-of-the-art conventional segmentation models of similar size. The FG-DM samples higher quality masks with only 10 DDIM steps, comparable to the throughput of Segformer-B5 but with superior quality as shown in the FID, Precision and Recall metrics.}
\label{tab:seg_models}
\setlength{\tabcolsep}{2pt}
\begin{tabular}{l|c|c|c|c|c|c}
\hline
Model (ADE20K) & Params (M) & FID$\downarrow$ & LPIPS$\uparrow$ & Precision$\uparrow$ & Recall$\uparrow$ & T (imgs/s)$\uparrow$ \\
\hline
% SegNext-L \cite{guo2022segnext} & \textbf{49} & 112.9 & 0.782 & 0.58 & 0.04 & \textbf{10.3} \\
SegFormer-B5 \cite{xie2021segformer} & 85 & 112.6 & 0.781 & 0.61 & 0.04 & \textbf{6.4} \\
% FG-DM-Seg (5 DDIM steps) & 53 & 124.3 & 0.704 & 0.59 & 0.01 & \textbf{11.8} \\
FG-DM-Seg (10 DDIM steps) & 53 & \textbf{86.1} & \textbf{0.788} & \textbf{0.72} & \textbf{0.04} & \textbf{6.4} \\
FG-DM-Seg (20 DDIM steps) & 53 & \textbf{84.0} & \textbf{0.776} & \textbf{0.72} & \textbf{0.05} & 3.7 \\
FG-DM-Seg (200 DDIM steps) & 53 & \textbf{83.6} & \textbf{0.768} & \textbf{0.73} & \textbf{0.06} & 0.4 \\
\hline
\end{tabular}
\end{table*}

\text{\bf Computation-Performance Tradeoff:}

The image generation chain (final factor) of FG-DM has the same computation as existing VC-DMs since it re-uses existing conditional models such as ControlNet. The only additional computation comes from the condition synthesis factor and a faithful comparison of complexity must include the condition synthesis step. For existing DMs, generating/editing conditions for segmentation masks requires the following steps: (1) Generate image with DM. (2) Segment with off-the-shelf segmentation model. (3) Edit the segmentation. (4) Generate the image conditional on the manipulated segmentation.
The FG-DM eliminates step 1. replaces  step 2. with the segmentation synthesis step. This is much more efficient than running an image generation DM and a segmentation model as illustrated in Table \ref{tab:compare_rec_hyperparam} since FG-DM samples segmentations at a lower resolution. For example, a 20-step image generation, FG-DM (with ControlNet as the final factor) takes only 4.5s \textbf{(1.7x speedup)} as compared to 7.5s for the standard pipeline (image generation with SD, Segmentation with SegFormer \cite{xie2021segformer}(CEM), and conditional image generation with ControlNet). 

Next, we evaluate the generated mask quality from FG-DM.
Table \ref{tab:seg_models} compares the segmentation mask quality vs. throughput, for FG-DM and SegFormer (CEM) on the ADE20K dataset. In these experiments, SegFormer are applied to the validation dataset images and FG-DM masks are obtained with the validation prompts. Throughput is calculated on a single NVIDIA-TitanXp GPU with batch size 1 averaged over 2,000 runs. Performance is reported in terms of FID, LPIPS, Precision, and Recall metrics of the masks. For the FG-DM, results are presented at different sampling steps. \\
With only 10 timesteps, the FG-DM produces segmentations of quality superior to those of the CEM segmentation model. This is achieved with a throughput comparable to that of the bigger SegFormer-B5 model. Beyond 10 steps there are negligible improvements in all performance metrics, showing that conditions like segmentation masks can be generated much faster than natural images, due to their lower frequency content. The FG-DM merges the capability to generate segmentation masks of superior quality compared to current segmentation models, \textit{crucial for creative tasks with unseen images}, with a throughput comparable to the latter, which can only segment existing images.

\subsubsection{Models trained from scratch: Ablation Studies}\label{supp-ablation}

We next discuss some results for FG-DMs trained from scratch where the adapter in Fig. \ref{fig:arch} is removed and the intermediate conditions are concatenated to be fed to the subsequent factors.

\begin{figure*}\RawFloats
\begin{minipage}{\columnwidth}
\centering
\scriptsize
\captionof{table}{\textbf{Ablation study on Image Synthesis} by FG-DM trained separately and jointly with segmentation factor. FG-DM results presented as Images/Semantic maps.}
\label{tab:uncond}
\label{fig:pose_and_maps}
\setlength{\tabcolsep}{1pt}
\resizebox{\textwidth}{!}{
\begin{tabular}{ l| l |l |l |l |l| l |l |l |l| l |l |l |l| l |l |l |l}
\toprule
Model &
 {\centering \#P} & 
\multicolumn{4}{c|}{\centering $\textbf{MM-CelebA}$} &
\multicolumn{4}{c|}{\centering $\textbf{CityScapes}$} &
\multicolumn{4}{c|}{\centering $\textbf{ADE-20K}$} 
& \multicolumn{4}{c}{\centering $\textbf{COCO-Stuff}$} 
\\
& (M)& FID $\downarrow$ & LPIPS $\uparrow$& Pr$\uparrow$ & Re$\uparrow$& FID $\downarrow$& LPIPS $\uparrow$& Pr$\uparrow$ & Re$\uparrow$& FID $\downarrow$& LPIPS $\uparrow$& Pr $\uparrow$& Re $\uparrow$ & FID $\downarrow$& LPIPS $\uparrow$& Pr $\uparrow$& Re $\uparrow$  \\
\midrule
% Unconditional LDM \cite{rombach2022high} & 87& 24.3/- & 0.58/- &  0.78&0.34&97.2/-&  0.63/-& &&31.0/- &  0.78/-&0.69&0.32&  48.2/-& 0.79/-& \textbf{0.73}&0.24\\
% \hline
% \rowcolor{Cyan}
\textbf{FG-DM (Separate training)}& 140& 23.2/20.8 & 0.57/0.54 & \textbf{0.83}&0.29&  54.7/61.8&  0.56/0.54& 0.66&0.17&34.0/83.9 &  0.79/0.77&0.71&0.25 & 35.3/\textbf{40.6}& 0.83/0.8& \textbf{0.71}&0.33\\
% \rowcolor{Cyan}
\textbf{FG-DM (Joint training)}& 140& \textbf{21.3}/\textbf{20.3} & \textbf{0.58}/\textbf{0.54} &  0.81&\textbf{0.34}&\textbf{47.6}/\textbf{61.8}&  \textbf{0.59}/\textbf{0.57}& \textbf{0.69}&\textbf{0.31}&\textbf{29.6}/\textbf{83.6} &  \textbf{0.79}/\textbf{0.77}&\textbf{0.72}&\textbf{0.34} & \textbf{33.1}/57.4& \textbf{0.83}/\textbf{0.8}& 0.69&\textbf{0.43}\\
%&  & \\
% \rowcolor{Cyan}
% FG-DM& 312& 21.3/20.3 & 0.578/0.543 & 29.6/83.6 &  0.785/0.768&  55/55&  0.598/\\ %&  & \\
\bottomrule
\end{tabular}
}
\end{minipage}
\end{figure*}

\subsubsubsection{\textbf{Joint Synthesis:}}
Table~\ref{tab:uncond} compares end-to-end training of FG-DM to separate training of VC-DM factors. Jointly training the FG-DM improves the image quality (lower FID) and diversity (higher LPIPS) on all four datasets. 

% \text{\bf Joint Synthesis by concatenation}
Table \ref{tab:ablation} shows the comparison of FG-DM with joint modeling by concatenation, where conditioning variable(s) are concatenated in the latent space and denoised jointly with the image using a single DM. The table clearly shows that FG-DM outperforms the concatenation approach by \textbf{8} points on the FID metric despite being smaller in size. Note that joint denoising by concatenation requires a larger model and forfeits many advantages of the FG-DM, such as higher object recall, image editing, and computational efficiency. 

\text{\bf Order of conditions: }
Table \ref{ablation:order} ablates the order of the conditioning variables for the model with semantic (S) and pose (P) condition on CelebA-HQ and COCO datasets which shows that the order of the chain affects final image synthesis quality (see FID).  This primarily stems from the misalignment of the pose map with semantic map and image, where the generated pose maps are inferior for the alternate order (pose-semantic-image). It shows that the unconditional generation of sparse pose maps is much more difficult than segmentations, due to the incomplete poses (occlusion) and missing context in the pose images. Further, Table~\ref{ablation:order} shows that the extra pose conditioning improves the image synthesis quality, reducing FID by \textbf{3.7}/\textbf{2.33} points on CelebA-HQ/COCO datasets respectively. This shows that adding more conditions is helpful for improving the image quality.

\subsubsubsection{\textbf{Data Augmentation}}

\begin{table}[!t]\RawFloats
\centering
\captionof{table}{Ablation study on data augmentation with synthetic data generated by the FG-DM for facial part segmentation on MM-CelebA-HQ\cite{lee2020maskgan} dataset.} 
\scriptsize
\setlength{\tabcolsep}{4pt}
%\resizebox{\linewidth}{!}{
\begin{tabular}{l|c|c|c|c}
\hline
 Data& \#Samples& mIoU$\uparrow$ & F.W. mIoU$\uparrow$& F1-score$\uparrow$ \\
\hline
Orig & 24993& 51.1 & 85.6& 61.5 \\
Orig+Syn & +1000& \textbf{55.1} & \textbf{87.9}& \textbf{65.5} \\
Orig+Syn & +2000& \textbf{55.0} & \textbf{87.1}& \textbf{65.0}   \\
\hline
% \rowcolor{LightCyan}
% \hline
\end{tabular}
%}
\label{tab:segmentation}
% \end{minipage}
% \begin{minipage}{0.38\columnwidth}
\captionof{table}{Ablation study on data augmentation with synthetic data generated by the FG-DM for face landmark estimation on 300W\cite{300wdataset} dataset.}
\scriptsize
\setlength{\tabcolsep}{4pt}
%\resizebox{\linewidth}{!}{
\begin{tabular}{l|c|c|c|c}
\hline
 NME& \#Samples&  Common$\downarrow$ & Full$\downarrow$ & Challenge$\downarrow$ \\
\hline
% Orig & 3000 &2.91	&3.34	&3.85 \\
Orig & 3000 &	3.21&3.64	& 5.81\\
Orig+Syn& +1000 &  \textbf{3.12}& \textbf{3.54}& \textbf{5.80}\\
Orig+Syn& +2000 &  \textbf{3.18}& \textbf{3.61}& \textbf{5.81}\\
\hline
% \rowcolor{LightCyan}
% \hline
\end{tabular}
%}
\label{tab:keypoints}
\end{table}

\begin{figure*}[!t]\RawFloats
\centering
\begin{minipage}{0.38\columnwidth}
\centering
\scriptsize
\captionof{table}{Comparison of FG-DM conditioning vs concatenation approach for joint synthesis on CelebA-HQ. U-LDM reported for reference.}
\label{tab:ablation}
\setlength{\tabcolsep}{2pt}
\begin{tabular}{ l| l | l| l l} 
\toprule
 Model&\#P (M) & FID $\downarrow$ & LPIPS $\uparrow$ \\
 % & & & Acc (\%)\\
\midrule
U-LDM \cite{rombach2022high} &    87& 24.3& 0.586 \\
\hline
Joint (Concat)&   248& 29.4& \textbf{0.618}\\
FG-DM (Ours) &     \textbf{140}& \textbf{21.3}&0.578\\
\bottomrule
\end{tabular}
%}
\end{minipage}
\hspace{1mm}
\begin{minipage}{0.56\columnwidth}
\centering
\scriptsize
\captionof{table}{Ablation on the order of generated semantic map (S)  and pose (P) conditions on CelebA-HQ (Top) and COCO (Bottom). I - Image.}
\label{ablation:order}
\setlength{\tabcolsep}{2pt}
\begin{tabular}{l|c|c|c|c}
\toprule
Model & FID $\downarrow$& LPIPS $\uparrow$& P $\uparrow$& R $\uparrow$\\
\midrule
% CelebA-HQ (M$\to$I) & 21.31 & 0.578 & \textbf{0.81} & 0.34 \\
(P$\to$S$\to$I) & 23.43 & \textbf{0.616} & 0.616 & \textbf{0.466} \\
(S$\to$P$\to$I) & \textbf{17.61} & 0.594 & \textbf{0.754} & 0.403 \\
\hline
(P$\to$S$\to$I) & 31.43 & 0.855 & 0.547 & \textbf{0.343} \\
(S$\to$P$\to$I) & \textbf{30.77} & \textbf{0.857} & \textbf{0.564} & 0.322 \\
\bottomrule
\end{tabular}
\end{minipage}
\end{figure*}

An additional benefit of the FG-DM is that it can be used as a data augmentation technique, synthesizing  data to train models for segmentation, pose estimation, etc.  
By sampling data from the FG-DM, it is possible to produce labeled datasets of virtually unlimited size. These could be, in principle, useful to train downstream models for various vision applications. To investigate this, we start by training both the FG-DM and the downstream model on a labeled dataset $A$. We then use the FG-DM to synthesize an additional dataset B of images and labels and retrain the downstream model on $A$ $\cup$ $B$. We finally compare the performance of the two downstream models. We performed this experiment for two downstream tasks: {\it part segmentation\/} with the BiSeNet\cite{Yu_2018_ECCV_bisenet} network on MM-CelebAA-HQ, and {\it pose estimation\/} with the HRNetV2-W18  \cite{WangSCJDZLMTWLX19_hrnet} network on the 300W~\cite{300wdataset} dataset. In all cases, the downstream model is initialized with ImageNet pre-trained weights.  Tables \ref{tab:segmentation} and \ref{tab:keypoints} show that the addition of synthetic data always improves downstream model performance. The gains are larger for the more challenging segmentation task, where 1000 samples of synthetic data significantly improve the baseline mIOU and F1-score by \textbf{4\%}. For pose estimation, the addition of synthetic data reduces the already low normalized mean error (NME) of keypoint location by an additional 0.1\%. Performance saturates or decreases slightly when the number of synthetic images is increased to 2000 as compared to using only 1000 images. We conjecture that this occurs from the inclusion of redundant or noisy images in the enlarged dataset, which are less likely to contribute meaningfully to the model's training. Nevertheless, the performance still shows improvement compared to the baseline model that did not use any synthetic data augmentation. These results suggest that the FG-DM can be used for data augmentation, enabling significant performance gains by automated data augmentation.  

\def\ew{.05in}
\begin{figure*}\RawFloats
\begin{minipage}{\columnwidth}
\centering
\captionof{table}{\textbf{Ablation study on Image Synthesis} by FG-DM for sequential and joint inference. For the same segmentation maps, joint inference is superior to sequential inference showing the benefit of joint modeling.}
\label{tab:joint-vs-sequential}
\scriptsize
\setlength{\tabcolsep}{2pt}
\resizebox{\textwidth}{!}{
\begin{tabular}{ l| l |l |l |l |l| l |l |l |l| l |l |l |l| l |l |l |l}
\toprule
Model &
 {\centering \#P} & 
\multicolumn{4}{c|}{\centering $\textbf{MM-CelebA}$} &
\multicolumn{4}{c|}{\centering $\textbf{Cityscapes}$} &
\multicolumn{4}{c|}{\centering $\textbf{ADE-20K}$} 
& \multicolumn{4}{c}{\centering $\textbf{COCO}$} 
\\
& (M)& FID $\downarrow$ & LPIPS $\uparrow$& Pr$\uparrow$ & Re$\uparrow$& FID $\downarrow$& LPIPS $\uparrow$& Pr$\uparrow$ & Re$\uparrow$& FID $\downarrow$& LPIPS $\uparrow$& Pr $\uparrow$& Re $\uparrow$ & FID $\downarrow$& LPIPS $\uparrow$& Pr $\uparrow$& Re $\uparrow$  \\
\midrule
\textbf{FG-DM (Sequential Inference)}& 140& 34.5/20.3 & 0.56/0.54 & 0.79&0.20&  57.2/61.8&  0.59/0.57& 0.44&0.11&31.1/83.9 &  0.78/0.77&0.71&0.32 & 35.3/57.4& 0.82/0.8& 0.67&0.41\\
\textbf{FG-DM (Joint Inference)}& 140& \textbf{21.3}/\textbf{20.3} & \textbf{0.58}/\textbf{0.54} &  \textbf{0.81}&\textbf{0.34}&\textbf{47.6}/\textbf{61.8}&  \textbf{0.59}/\textbf{0.57}& \textbf{0.69}&\textbf{0.31}&\textbf{29.6}/\textbf{83.6} &  \textbf{0.79}/\textbf{0.77}&\textbf{0.72}&\textbf{0.34} & \textbf{33.1}/\textbf{57.4}& \textbf{0.83}/\textbf{0.8}& \textbf{0.69}&\textbf{0.43}\\
\bottomrule
\end{tabular}
}
\end{minipage}
\end{figure*}

\begin{figure}[h]\RawFloats
\centering
\includegraphics[width=0.5\columnwidth]{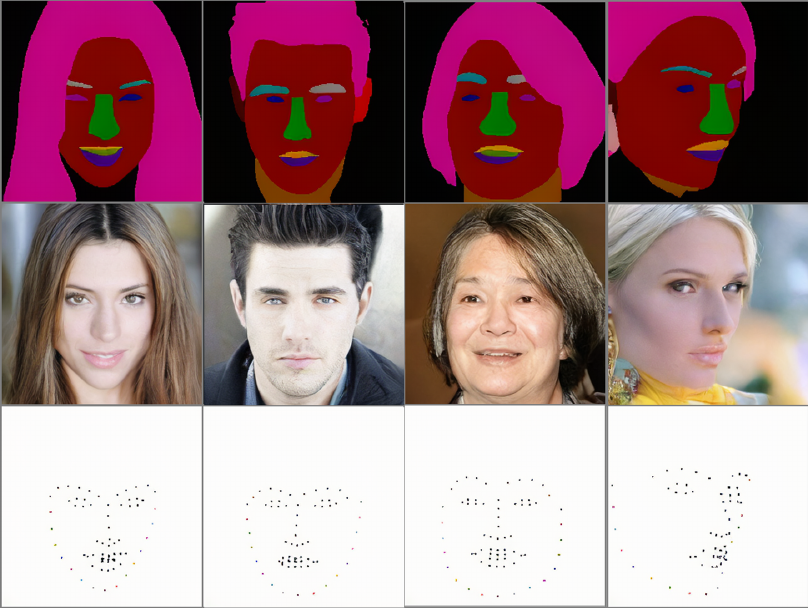}
\caption{\label{fig:qualitative-celeba} \textbf{Unconditional generated semantic, pose masks, and corresponding images using an FG-DM trained from scratch on MM-CelebA-HQ (256 x 256).}}
\end{figure} 

\subsubsubsection{\textbf{Sequential vs Joint Inference} } Table \ref{tab:joint-vs-sequential} shows the comparison of two modes of operation of FG-DM during inference. Joint inference is the standard mode of operation that samples the condition(s) and image jointly at each timestep (equations (2)-(4) in the paper). Sequential inference is the alternative mode of operation where the condition synthesis chain is run fully before feeding it to the conditional image generation factor. It is observed that the standard mode of joint inference sampling produces images of higher quality (lower FID) than the alternative mode.

\begin{figure*}[h]\RawFloats
\centering
    \begin{subfigure}{0.48\columnwidth}
        \centering
        \includegraphics[width=\columnwidth]{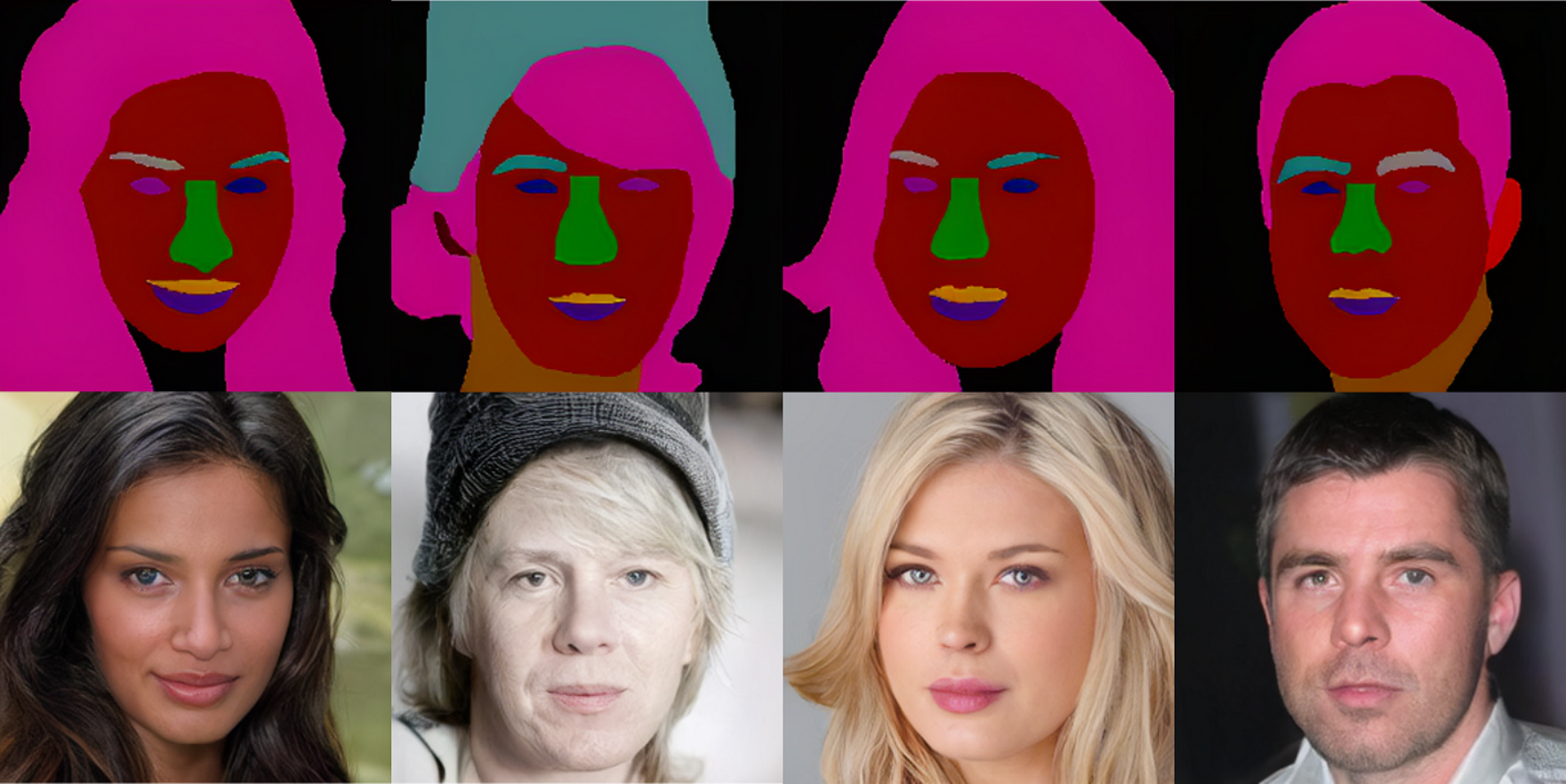}
        \caption{\label{teaser_celeba}MM-CelebA-HQ 256 x 256 samples}
    \end{subfigure}
    \hfill
    \begin{subfigure}{0.48\columnwidth}
        \centering
        \includegraphics[width=\columnwidth]{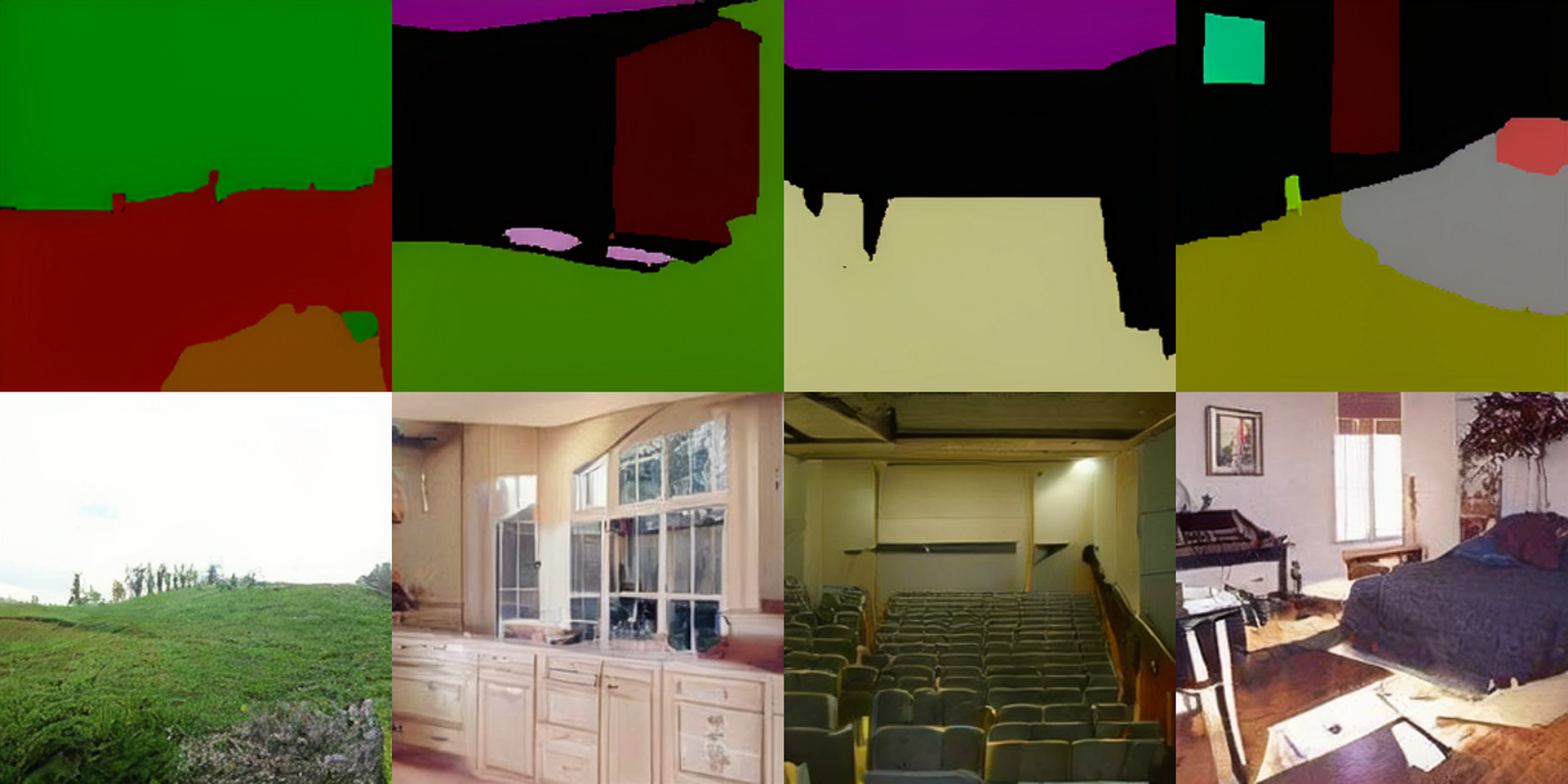}
        \caption{\label{teaser_ade}ADE-20K  256 x 256 samples}
    \end{subfigure}
    \hfill
    \begin{subfigure}{0.48\columnwidth}
        \centering
        \includegraphics[width=\columnwidth]{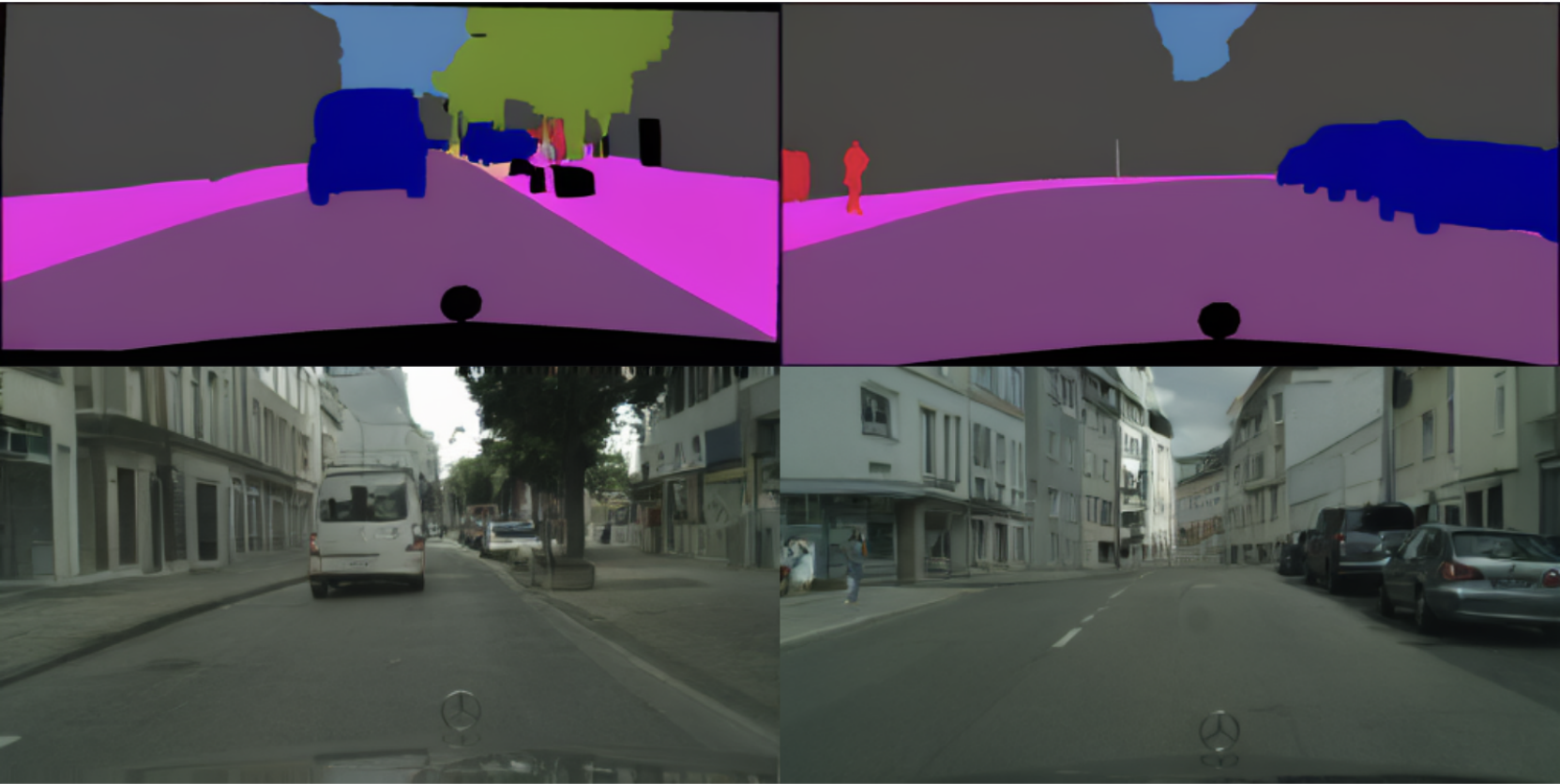}
        \caption{\label{teaser_city}Cityscapes 256 x 512 samples}
    \end{subfigure}
    \hfill
    \begin{subfigure}{0.48\columnwidth}
        \centering
        \includegraphics[width=\columnwidth]{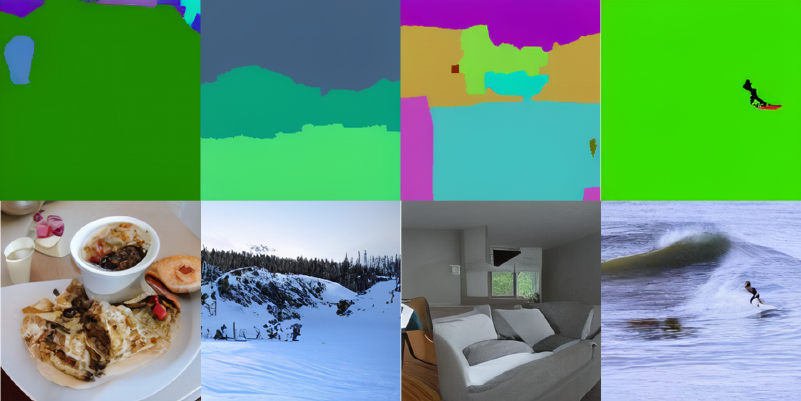}
        \caption{\label{teaser_coco}COCO 256 x 256 samples}
    \end{subfigure}
\caption{\label{fig:uncond} \textbf{Segmentation mask/image pairs synthesized by FG-DMs trained from scratch (53 M parameters) on the MM-CelebA-HQ, ADE-20K, Cityscapes and COCO datasets.}}
\end{figure*}

\begin{figure*}[h]\RawFloats
\centering
    \begin{subfigure}{0.48\columnwidth}
        \centering
        \includegraphics[width=\columnwidth]{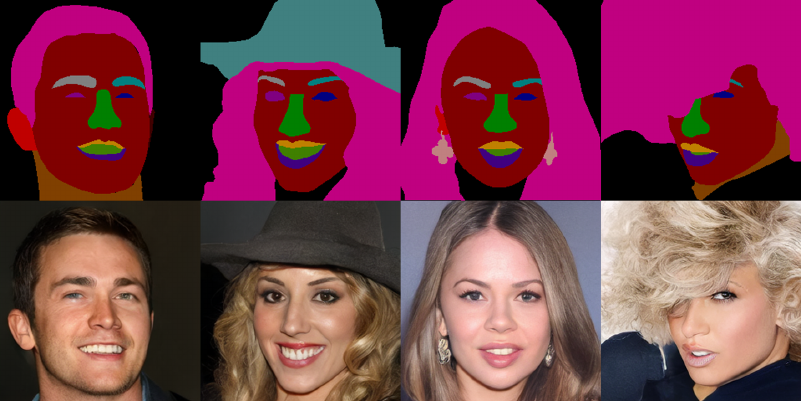}
        % \caption{\label{samples_sem_map_celeb}CelebA-HQ}
        % \includegraphics[width=\columnwidth]{images/ade/semantic-map/ade_semantic_map_sample1.pdf}
        % \caption{\label{samples_sem_map_ade}ADE-20K}
        % \includegraphics[width=\columnwidth]{images/celeba/samples1_gs-214000_e-000567_b-000322.pdf}
        \caption{\label{samples_sem_map_celeba}MM-CelebA-HQ}
    \end{subfigure}
    \hfill
    \begin{subfigure}{0.48\columnwidth}
        \centering
        % \includegraphics[width=\columnwidth]{images/celeba/semantic-map/sem-map-celeba-sample1.pdf}
        % \caption{\label{samples_sem_map_celeb}CelebA-HQ}
        \includegraphics[width=\columnwidth]{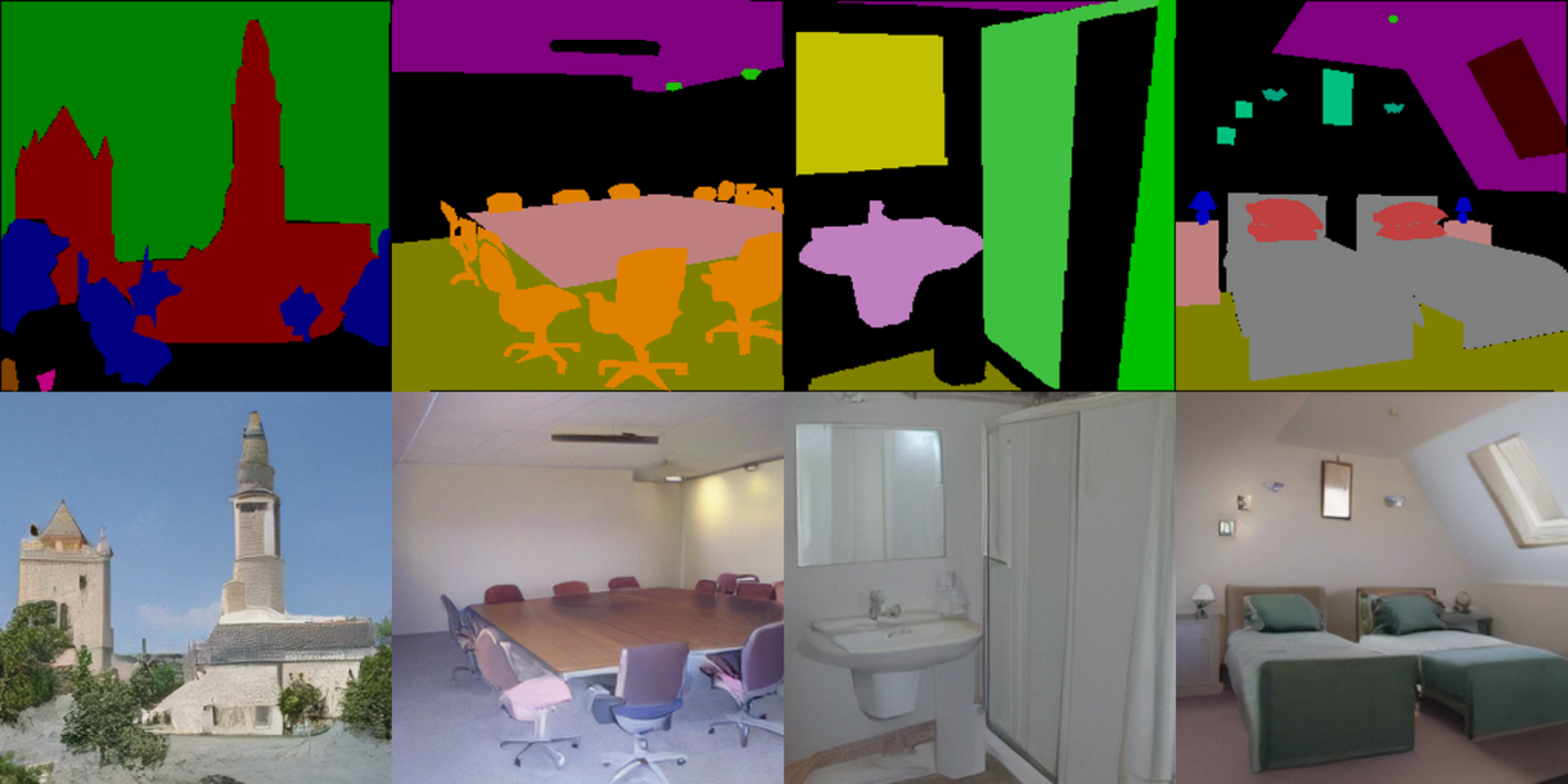}
        % \caption{\label{samples_sem_map_ade}ADE-20K}
        % \includegraphics[width=\columnwidth]{images/celeba/samples1_gs-214000_e-000567_b-000322.pdf}
        \caption{\label{samples_sem_map_ade_}ADE-20K}
    \end{subfigure}
        \hfill
    \begin{subfigure}{0.48\columnwidth}
        \centering
        % \includegraphics[width=\columnwidth]{images/celeba/semantic-map/sem-map-celeba-sample1.pdf}
        % \caption{\label{samples_sem_map_celeb}CelebA-HQ}
        % \includegraphics[width=\columnwidth]{images/ade/semantic-map/ade_semantic_map_sample1.pdf}
        % \caption{\label{samples_sem_map_ade}ADE-20K}
        \includegraphics[width=\columnwidth]{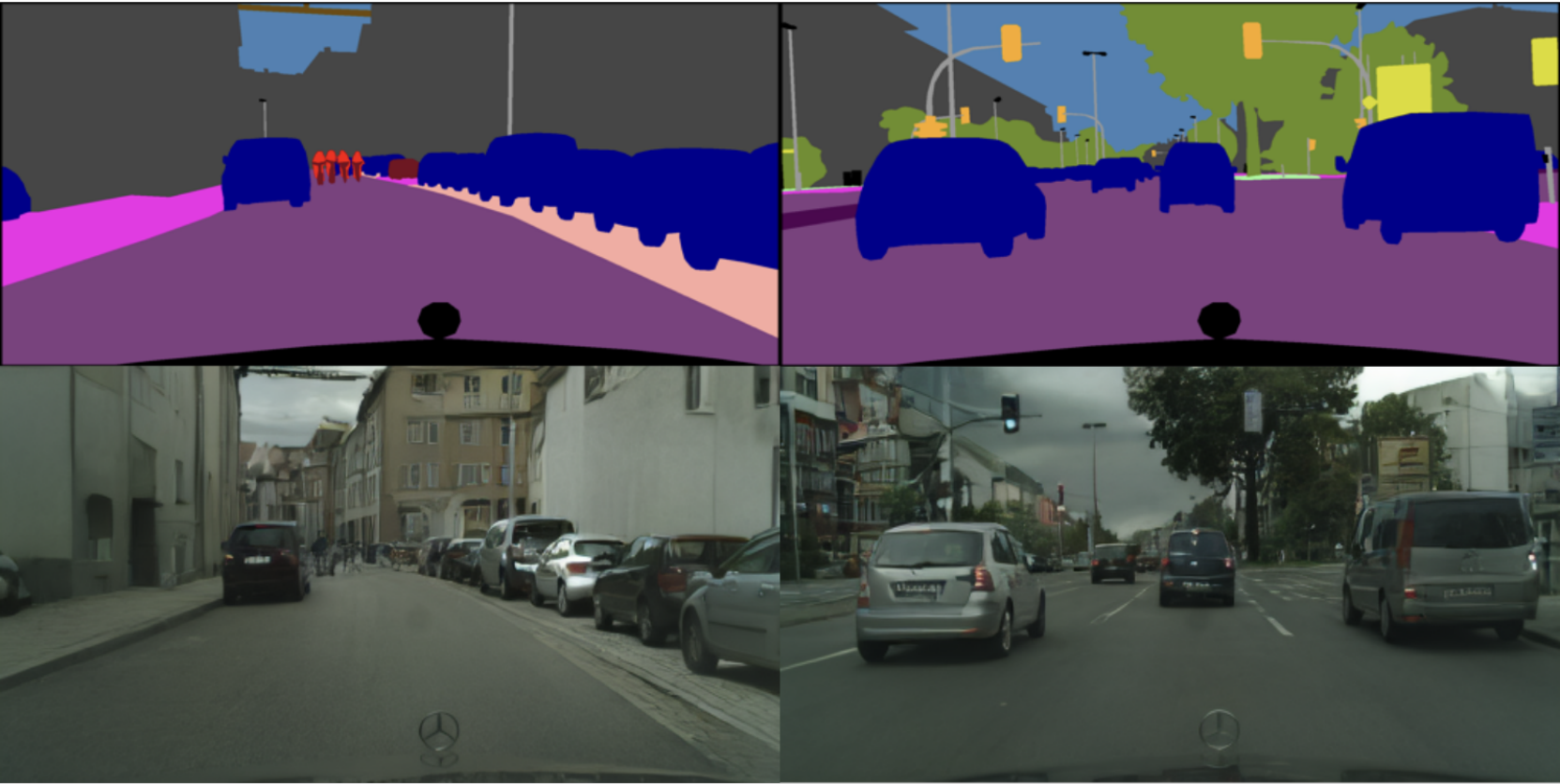}
        \caption{\label{samples_sem_map_city_}Cityscapes}
    \end{subfigure}
    \hfill
    \begin{subfigure}{0.48\columnwidth}
        \centering
        \includegraphics[width=\columnwidth]{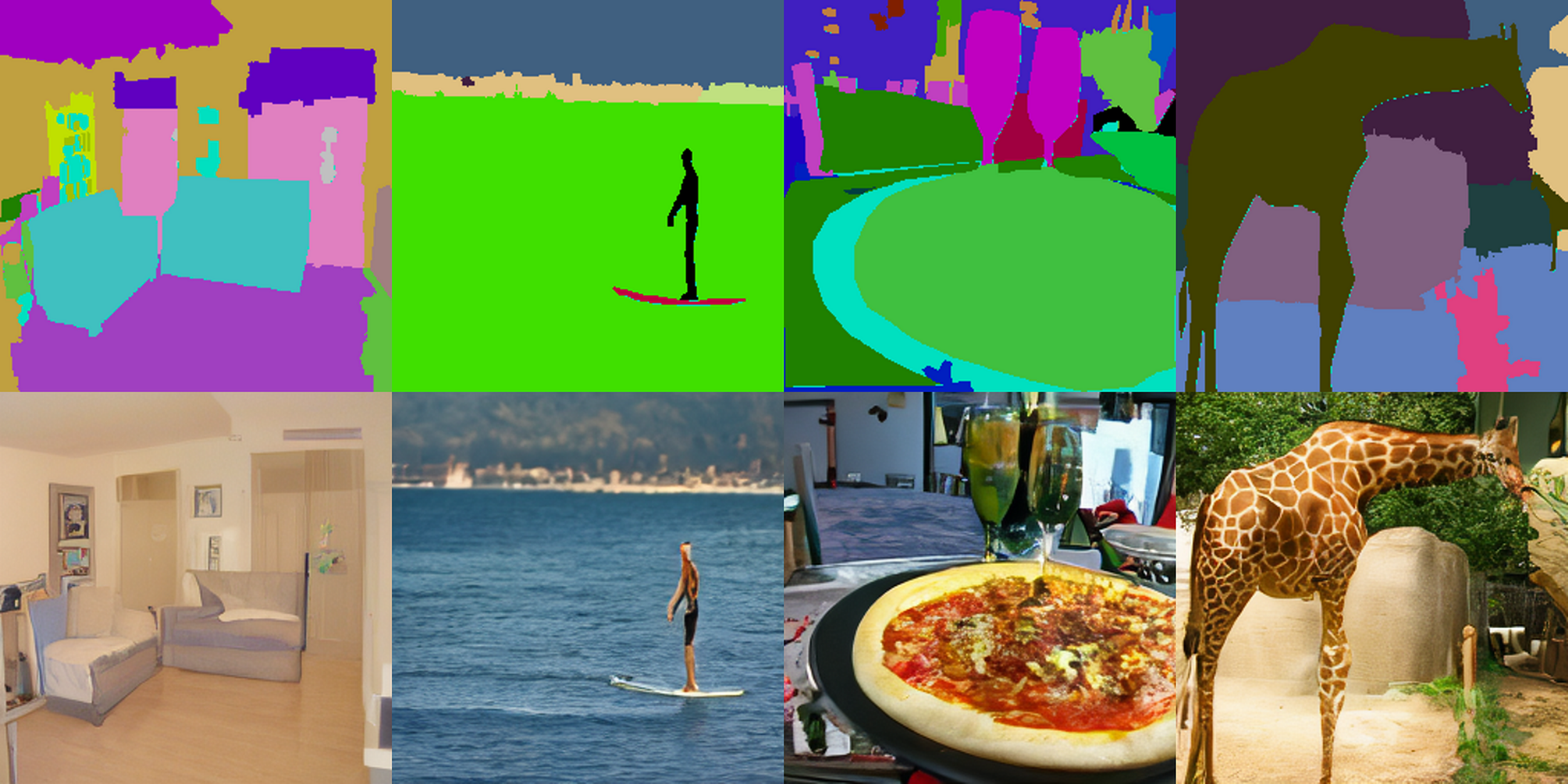}
        \caption{\label{samples_sem_coco}COCO}
    \end{subfigure}
    \hfill
\caption{\label{fig:sem_guide_gt} \textbf{Semantic Guided conditional image synthesis by FG-DM models trained from scratch (53 M parameters)}. For conditional synthesis, only the image-synthesis VC-DM factor is used. When conditioned by the validation dataset segmentation maps shown at the top, this factor synthesizes the images shown at the bottom.}
\end{figure*}

\subsubsubsection{\textbf{Qualitative results: Unconditional and Conditional Image Generation}}

Fig. \ref{fig:qualitative-celeba} shows qualitative results of unconditional synthesis on MM-CelebA-HQ for a model with factors for semantic map, pose and image synthesis. This improves the overall image quality as shown in Table \ref{ablation:order}. It is observed that the generated pose maps are accurate even for hard examples such as side-view faces.

Figure \ref{fig:uncond} shows qualitative results of image/segmentation mask pairs generated by FG-DM models trained from scratch on four popular semantic segmentation datasets. The FG-DM can generate good quality samples even on complex datasets such as ADE-20K and COCO using only a small model (53M parameters for each factor).

Figure \ref{fig:sem_guide_gt} shows qualitative results of conditional image synthesis with groundtruth maps of validation set on the four datasets using FG-DMs trained from scratch. For conditional synthesis, only the image-synthesis factor of the FG-DM is used. In these experiments, the model is conditioned by the ground-truth validation segmentation masks of the datasets. It is seen that FG-DM can generate high quality samples that align with the latter.

\def\ew{.05in}
\begin{figure*}\RawFloats
\begin{minipage}{\columnwidth}
\centering
\captionof{table}{\textbf{Ablation study on Image Alignment} with the segmentation masks by FG-DM trained separately and jointly.}
\label{fig:alignment}
\scriptsize
\setlength{\tabcolsep}{1pt}
\resizebox{\textwidth}{!}{
\begin{tabular}{ l| l |l |l |l |l| l |l |l |l}
\toprule
{\centering Model} &
 {\centering \#P} & 
\multicolumn{2}{c|}{\centering $\textbf{MM-CelebA}$} &
\multicolumn{2}{c|}{\centering $\textbf{Cityscapes}$} &
\multicolumn{2}{c|}{\centering $\textbf{ADE-20K}$} 
& \multicolumn{2}{c}{\centering $\textbf{COCO}$} 
\\
& (M)& mIoU $\uparrow$ & f.w. IoU $\uparrow$&  mIoU $\uparrow$ & f.w. IoU $\uparrow$& mIoU $\uparrow$ & f.w. IoU $\uparrow$& mIoU $\uparrow$ & f.w. IoU $\uparrow$  \\
\midrule
\textbf{FG-DM (Separate training)}& 140& 70.26& 85.77& 45.38& 81.84&18.07& 48.65 & 23.29& 37.33\\
\textbf{FG-DM (Joint training)}& 140& \textbf{70.57}& \textbf{87.79}& \textbf{52.73}& \textbf{86.72}&\textbf{22.22}& \textbf{52.76} & \textbf{23.76}& \textbf{38.64}\\
\bottomrule
\end{tabular}
}
\end{minipage}
\end{figure*}

\subsubsubsection{\textbf{Segmentation-Image Alignment}}

We quantitatively evaluate the alignment of the generated images with the corresponding segmentation mask using off-the-shelf pretrained models as described in Sec. \ref{implementation_details}. We compare the image alignment of FG-DM when training separately and jointly. Table \ref{fig:alignment} shows the results on four datasets where the mIoU score is computed using the groundtruth validation samples for each of the dataset. Once again, FG-DM trained jointly outperforms the FG-DM trained separately indicating the advantage of joint modeling in following the segmentation conditions accurately.

\subsection{Implementation details}
\subsubsection{Extracting COCO Object classes from the prompt using a LLM}\label{chatgpt-3.5-extract-objects}
We proposed FG-DM for faster sampling of images with high object recall and validated it by by using the groundtruth segmentation maps from the ADE20K validation dataset. For practical use, the object classes can either be manually specified or automatically extracted from the captions which is useful for images involving cluttered scenes with multiple objects. Here, we show that the object classes can be extracted from the caption using a LLM (e.g., chatGPT-3.5). 

\begin{figure*}[h]
\centering
\caption{\label{fig:chatgpt-class-extract} Snapshots of using chatGPT-3.5 to extract the object classes from the input prompt.}
    \begin{subfigure}{0.48\columnwidth}
        \centering
        \includegraphics[height=0.6\columnwidth]{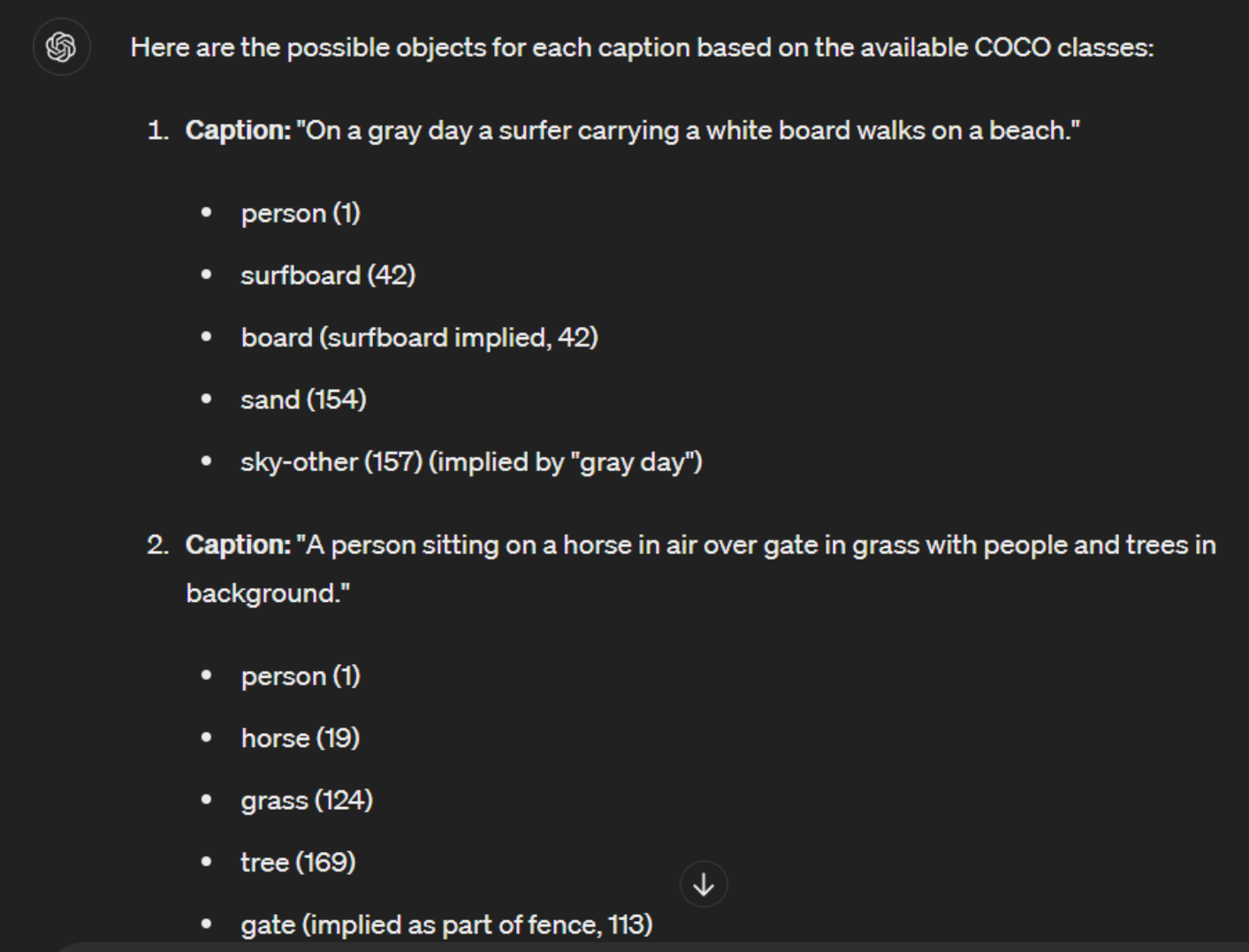}
        \caption{\label{chatgpt-object-class}Extracting COCO object classes from the prompt using chatGPT-3.5 in zero-shot manner.}
    \end{subfigure}
    \hfill
    \begin{subfigure}{0.48\columnwidth}
        \centering
        \includegraphics[height=0.6\columnwidth]{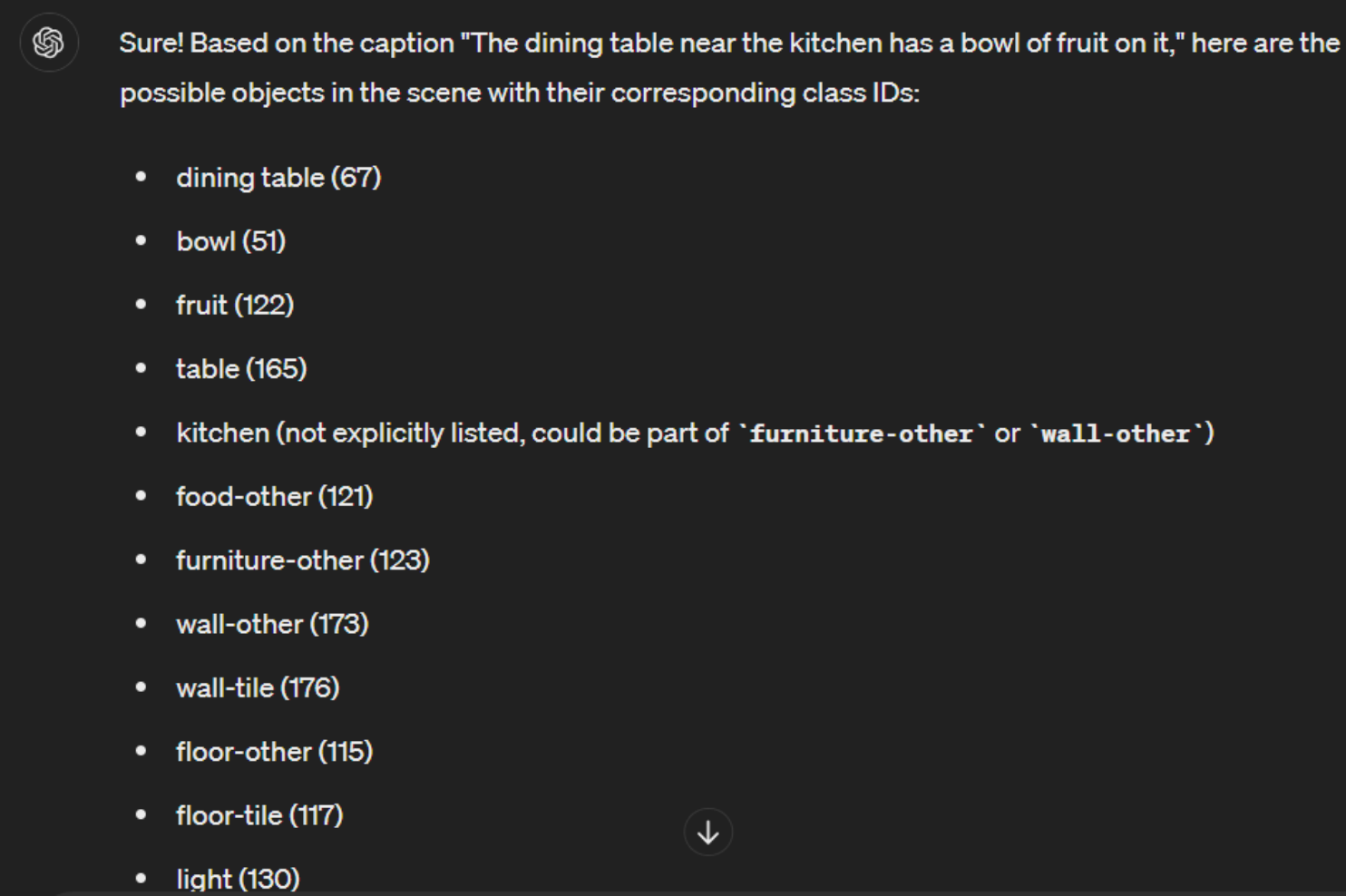}
        \caption{\label{chatgpt-1-shot-icl}Extracting COCO object classes from the prompt using 1-shot ICL with chatGPT-3.5}
    \end{subfigure}
\end{figure*}

Specifically, we use the following prompt to elicit responses from chatGPT-3.5. 
\begin{tcolorbox}[colback=yellow!10!white, colframe=yellow!50!black, title=Prompt]
You are ObjectGPT. You will list all possible objects in a scene from the caption description using the set of available classes. The available classes are as follows. 

\{COCO Classes inserted here as a dictionary mapping the class id to class name.\}
\end{tcolorbox}
Figure \ref{chatgpt-object-class} shows example outputs for the two prompts shown in the image. Note that this is zero-shot output where the model is not provided with any example pairs of prompt and corresponding object classes. The accuracy of the task can be further improved by using few-shot in-context examples as shown in the in-context learning (ICL) literature. We show an example of one-shot ICL with one prompt and its corresponding object classes from the COCO validation dataset. 
\begin{tcolorbox}[colback=blue!10!white, colframe=blue!50!black, title=1-shot ICL]
 "A man is in a kitchen making pizzas"

\{Object classes = [
        "person",
        "bottle",
        "cup",
        "knife",
        "spoon",
        "bowl",
        "broccoli",
        "carrot",
        "dining table",
        "oven",
        "sink",
        "branch",
        "cabinet",
        "floor-other",
        "floor-stone",
        "food-other",
        "furniture-other",
        "leaves",
        "light",
        "metal",
        "table",
        "textile-other",
        "wall-other",
        "wall-stone"]\}
\end{tcolorbox}
 Fig. \ref{chatgpt-1-shot-icl} shows the result for the prompt "The dining table near the kitchen has a bowl of fruit on it." using one-shot ICL shown above.

\subsubsection{Segmented Image Editor}
\begin{figure*}[h]
\centering
\caption{\label{fig:segmented-image-editor} Snapshots of editing capabilities using our segmented image editing tool.}
    \begin{subfigure}{0.32\columnwidth}
        \centering
        \includegraphics[width=\columnwidth]{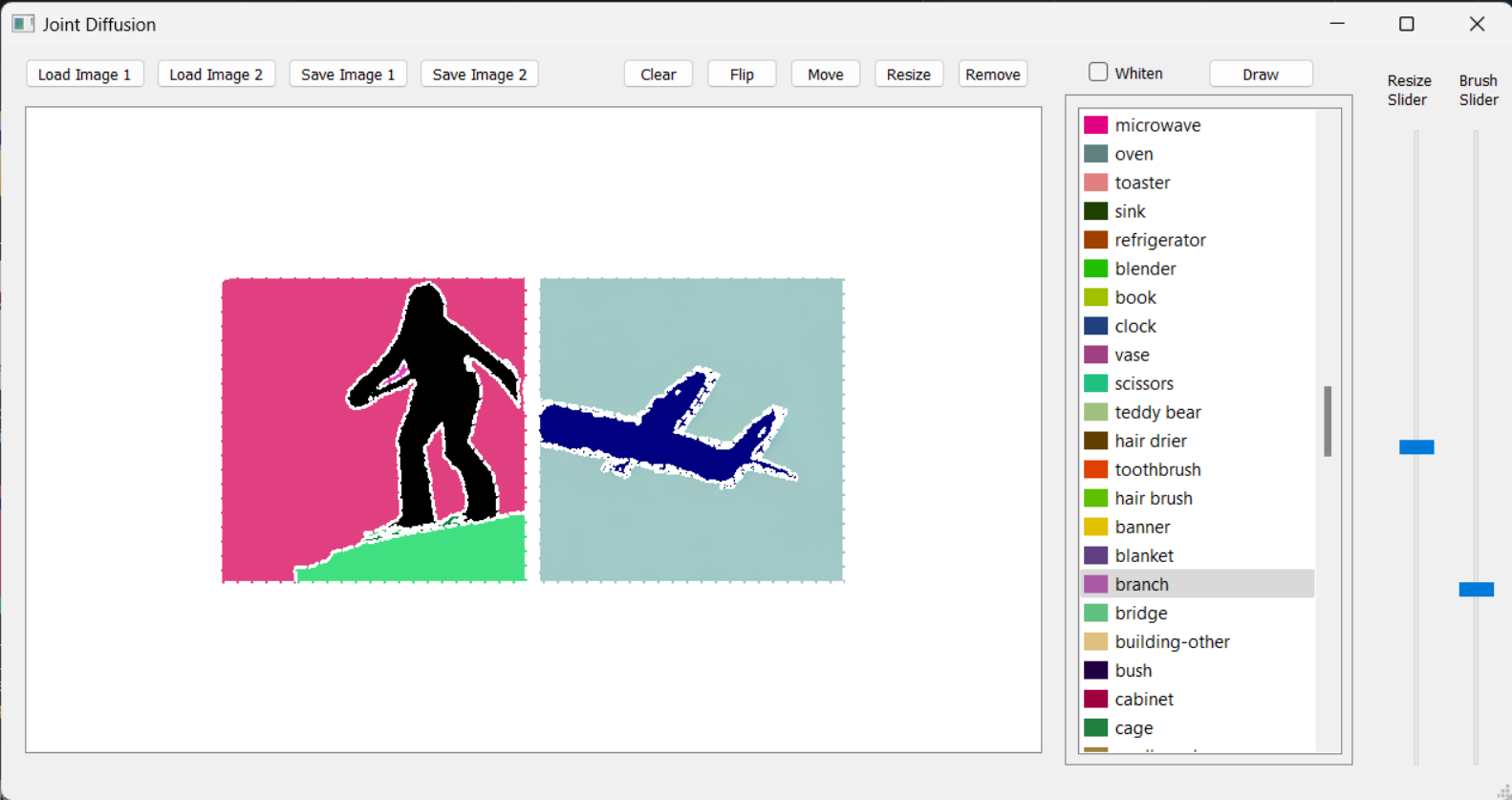}
        \caption{\label{load}Loading the generated segmentation masks into the tool}
    \end{subfigure}
    \hfill
    \begin{subfigure}{0.32\columnwidth}
        \centering
        \includegraphics[width=\columnwidth]{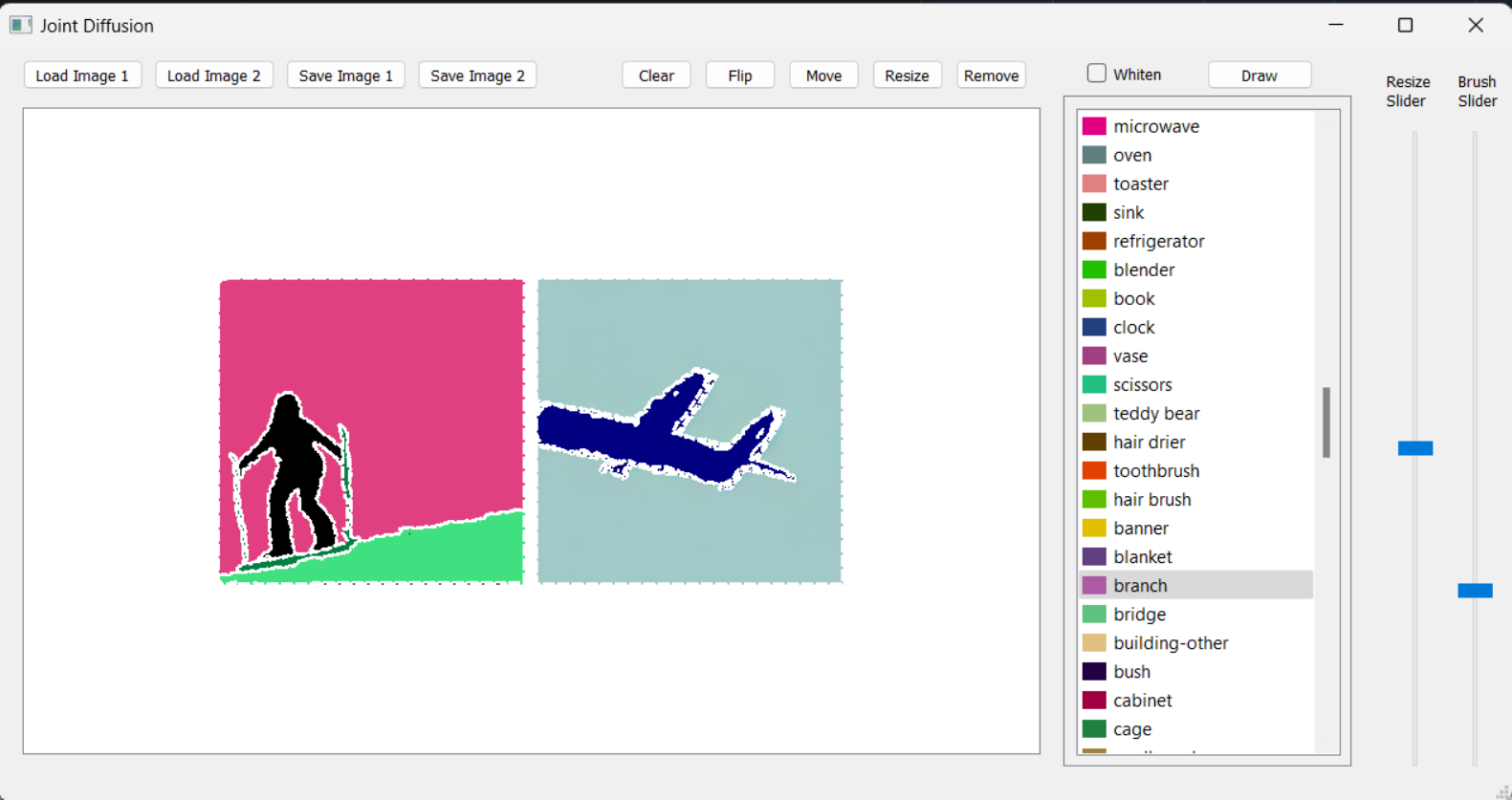}
        \caption{\label{move}Resizing the person and moving them to the left of the image}
    \end{subfigure}
    \hfill
    \begin{subfigure}{0.32\columnwidth}
        \centering
        \includegraphics[width=\columnwidth]{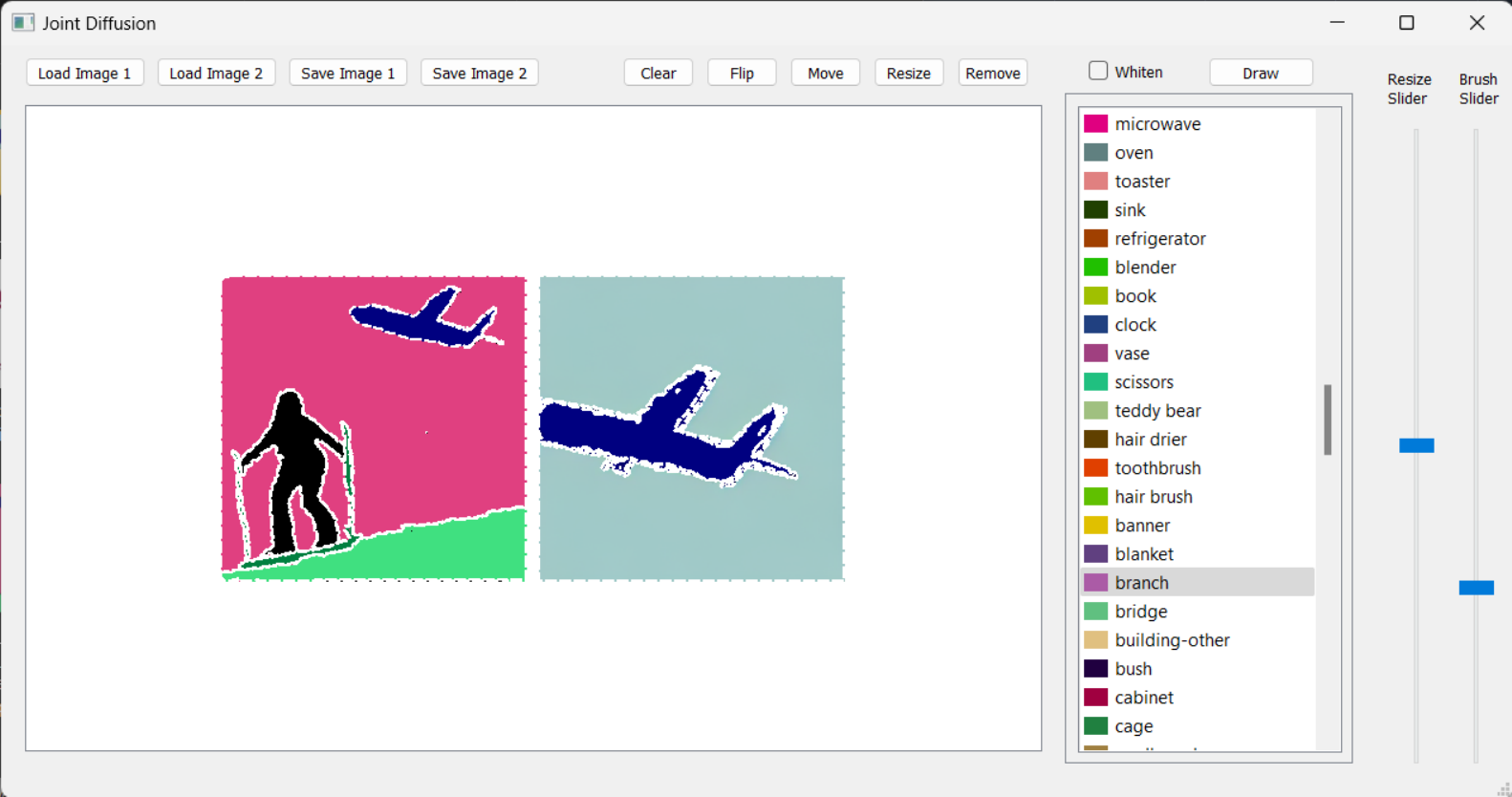}
        \caption{\label{edit_final}Final edited mask after resizing and moving the airplane}
    \end{subfigure}
\end{figure*}

We introduce a tool, developed using PyQT to edit the segmentation masks synthesized by the FG-DM, enabling an array of options. As shown in Figure~\ref{fig:segmented-image-editor}, this app has a user-friendly interface for loading and manipulating objects across two distinct segmented maps, which enables very flexible image synthesis. Key features include the ability to add, move, resize or remove objects, flip them, or replace backgrounds with ease. A unique drawing tool, augmented by a color palette representing 183 objects from the COCO dataset, allows for precise and detailed customization. Furthermore, the app's pointer size adjustment slider for drawing and resizing ensures users can achieve the exact level of size, detail and boldness needed for image editing. Currently, the complexity of these edits is limited by the simplicity of the image editing tool we developed. More complex images will likely be possible with further editing tool development. 

\begin{figure}[h]\RawFloats
\begin{center}
\captionof{table}{\label{tab:hyperparams}Hyperparameter Settings for the FG-DMs trained from scratch on MM-CelebA-HQ, ADE-20K, Cityscapes and COCO datasets. We use an image resolution of 256 × 512 for  Cityscapes and 256 × 256 for the others.}
% \resizebox{\linewidth}{!}{
\scriptsize
\setlength{\tabcolsep}{4pt}
\begin{tabular}{ l| l | l | l | l}
\toprule
 &
{\centering MM-CelebA-HQ} &
{\centering ADE-20K} &
{\centering Cityscapes} &
{\centering COCO} 
\\
\midrule
$f$ & 4 & 4 & 4 & 4 \\
$z$-shape & 64 × 64 × 3 & 64 × 64 × 3 & 64 × 128 × 3 & 64 × 64 × 3 \\
$|Z|$ & 8192 & 8192 & 8192 & 8192 \\
Diffusion steps & 1000 & 1000 & 1000 & 1000 \\
Optimizer & AdamW & AdamW & AdamW & AdamW\\
Noise Schedule & linear & linear & linear & linear\\
Nparams & 86M & 86M & 86M & 86M\\
Channels & 128 & 128 & 128 & 128\\
Depth & 2 & 2 & 2 & 2\\
Channel Multiplier & 1,4,8 & 1,4,8 & 1,4,8  & 1,4,8\\
Attention resolutions & 32, 16, 8 & 32, 16, 8 & 32, 16, 8 & 32, 16, 8\\
Head Channels & 32 & 32 & 32 & 32\\
Batch Size & 12 & 12 & 12 & 12\\
Iterations & 632k & 632k & 93k & 632k \\
Learning Rate & 1e-6 & 1e-6 & 1.0e-4 & 1e-6 \\ 
\hline
\end{tabular}
% }
\end{center}
\end{figure}

\subsubsection{Hyperparameter Settings}
Table \ref{tab:hyperparams} summarizes the detailed hyperparameter settings of the FG-DMs trained from scratch reported in the main paper. 
For FG-DMs adapted from Stable Diffusion, we use the same settings as Stable Diffusion~\cite{rombach2022high} and train only the adapters for 100 epochs with a learning rate of 1e-6 using AdamW optimizer.

\subsubsection{Training Details}
We train all models using 2-4 NVIDIA-A40 GPUs or 2 NVIDIA-A100 GPUs based on the availability. For adapting Stable Diffusion, since we reuse existing conditional model such as ControlNet, we first pretrain the model for 100 epochs to synthesize the conditions (e.g., segmentation, depth, normal or sketch). We then jointly finetune the condition (e.g., segmentation) factor with the conditional image synthesis factor (e.g., ControlNet) for an additional 100 epochs by only updating a subset of parameters of the ControlNet adapter denoted by the $\texttt{input}$-$\texttt{hint}$-$\texttt{block}$ in the model while keeping the rest frozen. Note that the pretrained ControlNet can still be used in FG-DM without finetuning which results in a slightly lower image quality (also validated in Table \ref{tab:joint-vs-sequential}). For models trained from scratch, we train all the parameters of the U-Net model from scratch.

\subsubsection{Evaluation Details}\label{implementation_details}
This section provides additional details on evaluation for the experiments of Sec. 4.
% Quantitative Results in Unconditional and Class-Conditional Image Synthesis
We follow common practice and estimate the statistics for calculating the FID values \cite{fid_gan_17} shown in
Table \ref{tab:uncond} are based on 10k samples between FG-DM generated samples and the entire training set of each of the datasets. For calculating
FID scores we use the torch-fidelity package \cite{obukhov2020torchfidelity}. Following standard practice, we pre-process all the images by resizing to $256\times 256$ for MM-CelebA-HQ, ADE-20K and COCO datasets and $256\times 512$ for Cityscapes dataset for calculating the metrics. Samples are generated with 200 DDIM \cite{song2021denoising} steps and $\eta=1.0$. 
For the measuring the semantic alignment, we use off-the-shelf networks to evaluate the alignment of generated results. We use DRN-D-105 \cite{yu2017dilateddrn} for
Cityscapes, ResNet50Dilated-PPM \cite{zhou2017scene} for ADE-20K, Unet \cite{lee2020maskgan, ronneberger2015u}
for MM-CelebA-HQ and DeepLabV2 \cite{deeplabv2} for COCO dataset. The generated images are fed to these segmentation models to obtain a pseudo-mask which is compared against the mask which was used to generate the image. We use mean Intersection-overUnion (mIoU) and frequency weighted mIoU to measure the overlap between the generated images and the semantic masks. We calculate the mIoU by upsampling the generated images to the same resolution as default input resolution of the off-the-shelf segmentation models. For computing the metrics, we use the validation segmentation maps for each dataset (3000 images for MM-CelebA-HQ, 2000 images for ADE-20K, 500 images for Cityscapes and 5000 images for COCO) is used. However, it should be noted that this pseudo metric highly depends on the capability of the off-the-shelf network. A strong segmentation network measures the semantic alignment of the generated images more accurately. 
For all text-to-image synthesis results reported in the paper, we use classifier free guidance with scale 7.5, $\eta=0.0$ and 20 DDIM steps unless otherwise stated.

\subsection{Broader Impact}\label{broader-impact}
We introduce a new framework for controlling diffusion models that offers creative image synthesis with higher recall, greater flexibility, modularity and explainability. While it offers the benefits of revealing hidden harmful biases in existing image generative models and offers better interpretability, it can also be potentially misused to propagate harmful, unlawful or unethical information with harmful edits. Since, the framework is modular, any harmful edits can be identified before the image generation step where the segmentation or pose map factor can be filtered (automatically or manually) before proceeding to the image generation factor. Additionally, recent advancements in image watermarking \cite{luo2022leca} can help to identify generated image contents to protect against these risks.

\subsection{Future Work} For future work, FG-DM can be extended for Novel View Synthesis by adding a factor for Novel Views. Further, the Novel View FG-DM with depth/normal factors can be used as a strong prior for controllable Text-to-3D generation with SDS technique \cite{poole2022dreamfusion}. The modular nature of FG-DM also allows a potential extension for audio/video generation making FG-DM framework to be a strong candidate for multi-modal content generation.

%%%%%%%%%%%%%%%%%%%%%%%%%%%%%%%%%%%%%%%%%%%%%%%%%%%%%%%%%%%%

\end{document}